\documentclass[journal]{IEEEtran}

\usepackage{epsfig,graphicx,amsfonts,multirow,amsmath,hyperref}
\usepackage{times}
\usepackage{epsfig}
\usepackage{graphicx}
\usepackage{amsmath}
\usepackage{amssymb}
\usepackage{bbding}
\usepackage{pifont}
\usepackage{wasysym}
\usepackage{amssymb}
\usepackage{subfigure}
\usepackage{multirow}
\usepackage{pifont}
\usepackage{booktabs}
\usepackage{diagbox}
\usepackage[numbers,sort&compress]{natbib}
\usepackage[table]{xcolor}
\usepackage{stfloats}
\usepackage{threeparttable}

\begin{document}

\title{Dense Hybrid Proposal Modulation for Lane Detection}

\author{
    Yuejian Wu,
    Linqing Zhao,
    Jiwen Lu,~\IEEEmembership{Senior Member,~IEEE},
    and Haibin Yan,~\IEEEmembership{Member,~IEEE}
    \IEEEcompsocitemizethanks{
         \IEEEcompsocthanksitem This work was supported in part by the National Natural Science Foundation of China under Grant 61976023, Grant U22B2050, and Grant 62125603.
        \textit{(Corresponding author: Haibin Yan).}
        \IEEEcompsocthanksitem Yuejian Wu and Haibin Yan are with the School of Automation, Beijing University of Posts and Telecommunications, Beijing, 100876, China (e-mail: wuyuejian@bupt.edu.cn; eyanhaibin@bupt.edu.cn).
        \IEEEcompsocthanksitem Linqing Zhao and Jiwen Lu are with the Department
        of Automation, Tsinghua University, Beijing 100084, China, and also
        with the Beijing National Research Center for Information Science and
        Technology (BNRist), Beijing 100084, China (e-mail: linqingzhao@tju.edu.cn; lujiwen@tsinghua.edu.cn).
    }
}

\maketitle

\begin{abstract}
In this paper, we present a dense hybrid proposal modulation (DHPM) method for lane detection. 
Most existing methods perform sparse supervision on a subset of high-scoring proposals, while other proposals fail to obtain effective shape and location guidance, resulting in poor overall quality.
To address this, we densely modulate all proposals to generate topologically and spatially high-quality lane predictions with discriminative representations.
Specifically, we first ensure that lane proposals are physically meaningful by applying single-lane shape and location constraints. 
Benefitting from the proposed proposal-to-label matching algorithm, we assign each proposal a target ground truth lane to efficiently learn from spatial layout priors.
To enhance the generalization and model the inter-proposal relations, we diversify the shape difference of proposals matching the same ground-truth lane.
In addition to the shape and location constraints, we design a quality-aware classification loss to adaptively supervise each positive proposal so that the discriminative power can be further boosted. 
Our DHPM achieves {very competitive} performances on four popular benchmark datasets. Moreover, we consistently outperform the baseline model on most metrics without introducing new parameters and reducing inference speed. The codes of our method are available at \url{https://github.com/wuyuej/DHPM}.

\end{abstract}

\begin{IEEEkeywords}
    Lane Detection, High-quality Proposal, Proposal Modulation, Hybrid Constraints
\end{IEEEkeywords}

\section{Introduction}
\label{sec:intro}
Given a front-viewed image taken by a camera mounted on the vehicle, lane detection aims to distinguish and locate the lane markings in an image. As a traditional yet fast-growing computer vision task, lane detection is one of the most essential and safety-critical components in autonomous driving and advanced driver-assistance systems (ADAS). It is developed with the desired purpose of avoiding traffic accidents and improving traffic efficiency. Thus it has drawn the increasing attention of researchers from academia and industry. 
However, a variety of challenging issues that may interfere with the detection have not been fully addressed yet. For example, many complex-shaped lane lines are difficult to fit well with existing methods. Moreover, lane lines may be inapparent or even invisible due to illumination conditions or due to occlusion by nearby vehicles. There has been a number of recent attempts to overcome these challenges.

To overcome these challenges, a variety of approaches have been proposed to detect lane lines with complex topologies, which can be roughly categorized into segmentation-based methods~\cite{culane,Lanenet,RESA,KeyPINet}, anchor-based methods~\cite{LaneATT,Lanenet,CurveLane,UFAST,UFLDv2,Philion_2019_CVPR,E2ELMD,CondLaneNet}, and curve-based methods~\cite{PolyLaneNet,LSTR,bezier}. Unlike general semantic segmentation methods, \textbf{segmentation-based} lane detectors are expected to be able to discriminate instance-level lanes. The disadvantage of these detectors is that they cannot be directly transferred to predict more lanes without modifying the model.
In addition, they usually require heavy post-processing or post-clustering strategies~\cite{cluster}.
To avoid trading performance for low latency, \textbf{anchor-based} methods~\cite{LaneATT,Lanenet,CurveLane} estimate the line shape by regressing the relative key coordinates. For example, some methods~\cite{Lanenet,CurveLane} construct line-shape anchors with vertical lines, while recent works~\cite{LaneATT} build a one-stage detection approach by designing straight-line anchors with arbitrary direction. However, these predefined anchors cannot describe complex line shapes, which results in relatively inferior performance. To make good use of the shape priors, recent methods adopt row-wise anchors~\cite{UFAST,UFLDv2,Philion_2019_CVPR,E2ELMD,CondLaneNet} and predict the line location for each row.
Different from points regression, \textbf{curve-based} methods~\cite{PolyLaneNet,LSTR,bezier} formulate the lane curve with parameters and detect lanes by regressing these parameters. LSTR~\cite{LSTR} and BézierLaneNet~\cite{bezier} predict the fitting parameters of each lane line, which can provide more proposals to improve the migration ability. Curve-based methods achieve high efficiency and superior performance benefitting from the end-to-end frameworks. However, curve-based methods are sensitive to the predicted parameters because the high-order coefficient may directly cause shape changes of lanes. As a brief summary, we attribute the unsatisfactory performance of existing methods to two fundamental but intractable challenges.

\begin{figure}[t]
    \centering
    \subfigure[Input image]{
    	\includegraphics[width=1.66in]{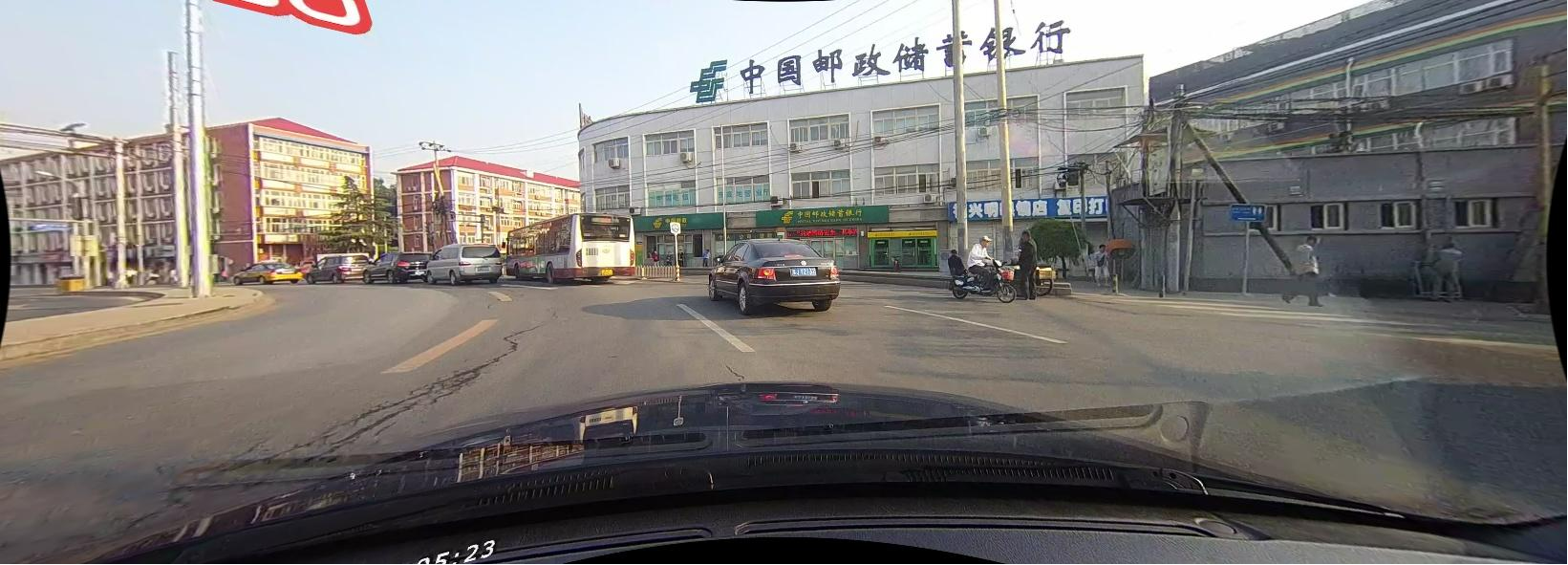}
    }
    \hspace{-3mm}
    \subfigure[Ground truth]{
    	\includegraphics[width=1.66in]{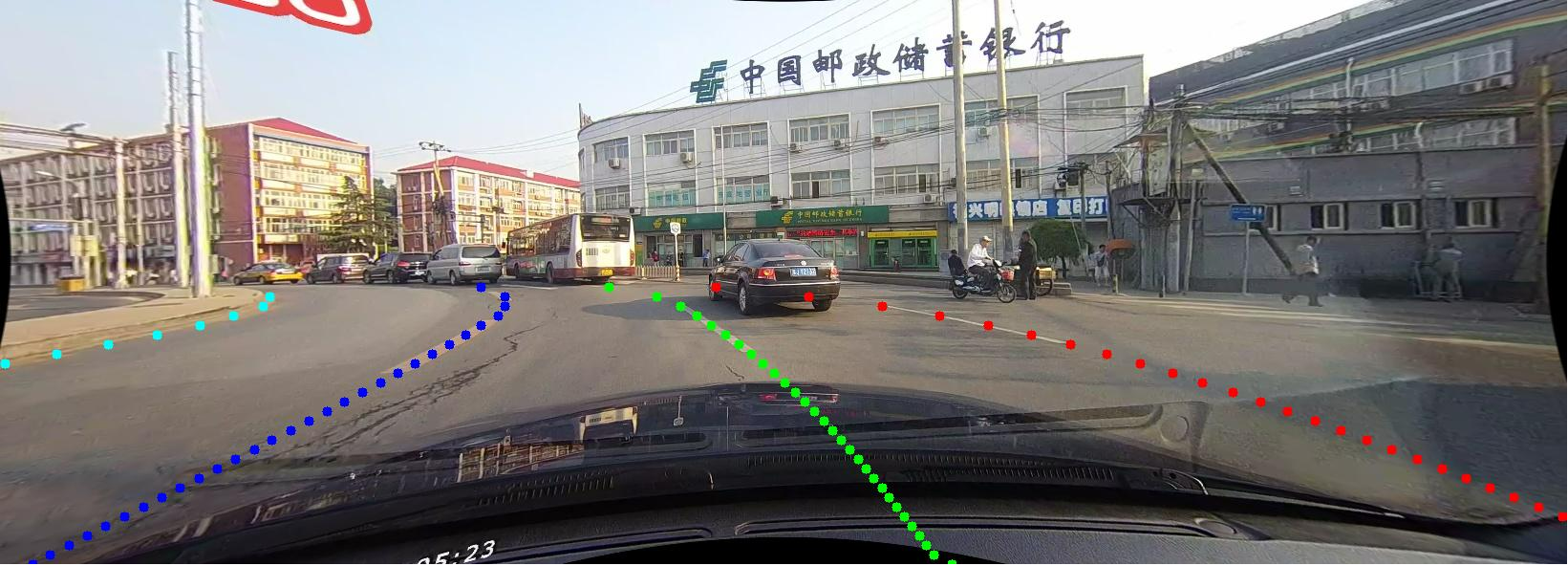}
    }
    \subfigure[Proposals of BézierLaneNet~\cite{bezier}]{
    	\includegraphics[width=1.66in]{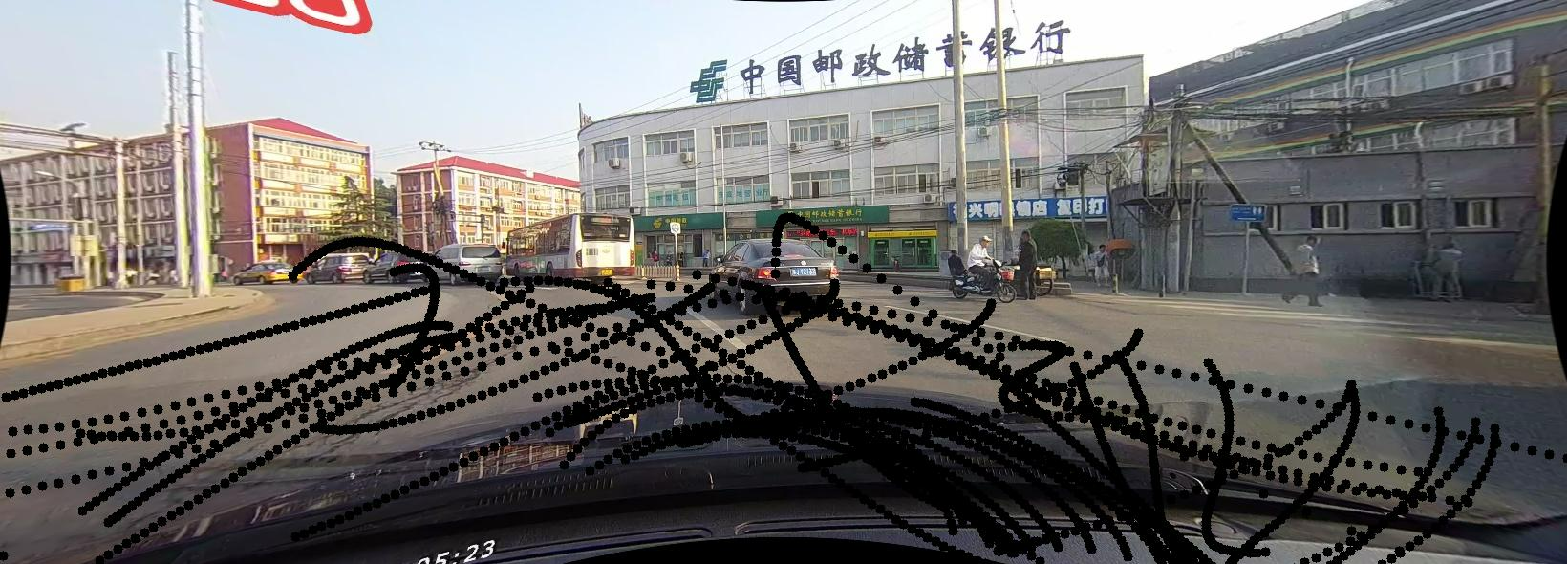}
    }
    \hspace{-3mm}
    \subfigure[Proposals of our method]{
    	\includegraphics[width=1.66in]{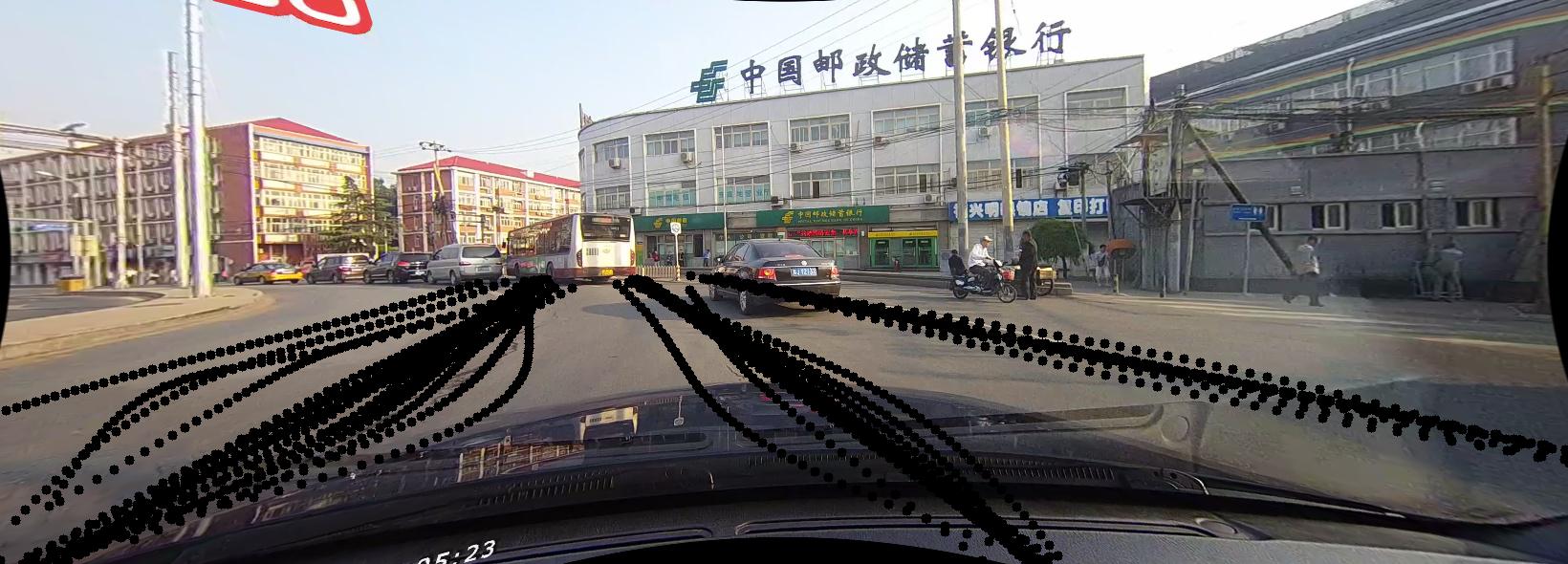}
    }
    \subfigure[Top 4 proposals of BézierLaneNet~\cite{bezier}]{
    	\includegraphics[width=1.66in]{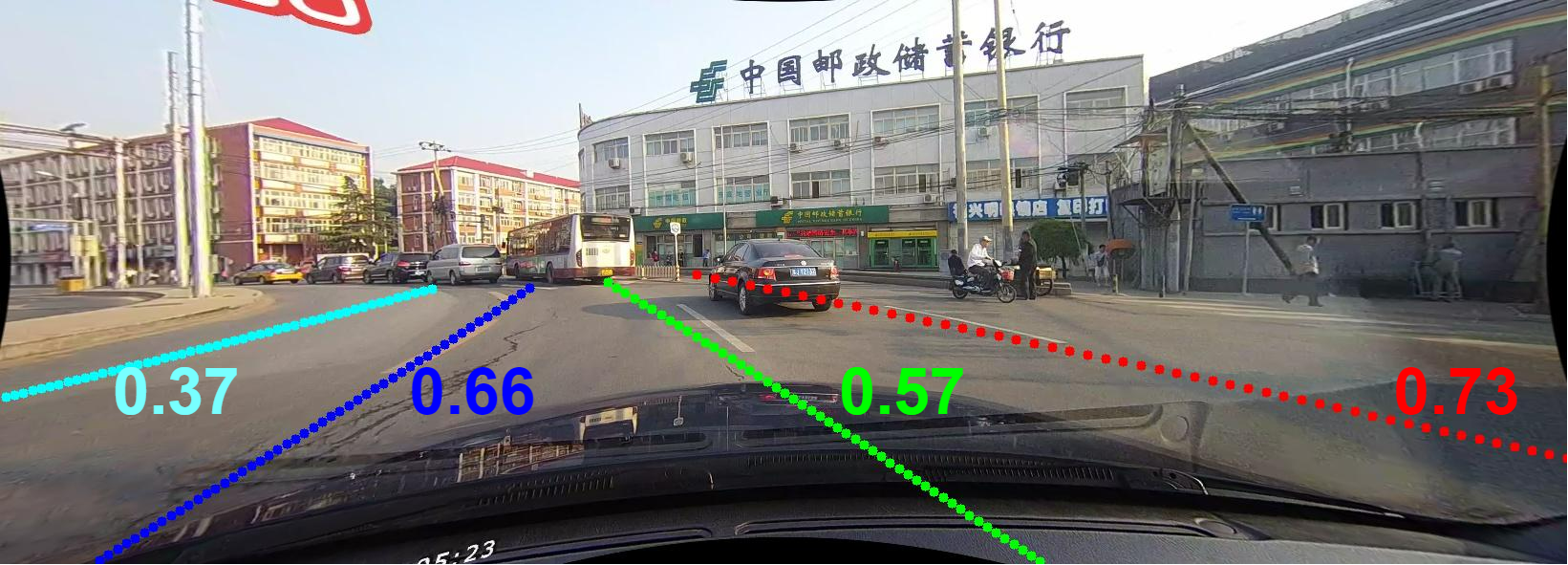}
    }
    \hspace{-3mm}
    \subfigure[Top 4 proposals of our method]{
    	\includegraphics[width=1.66in]{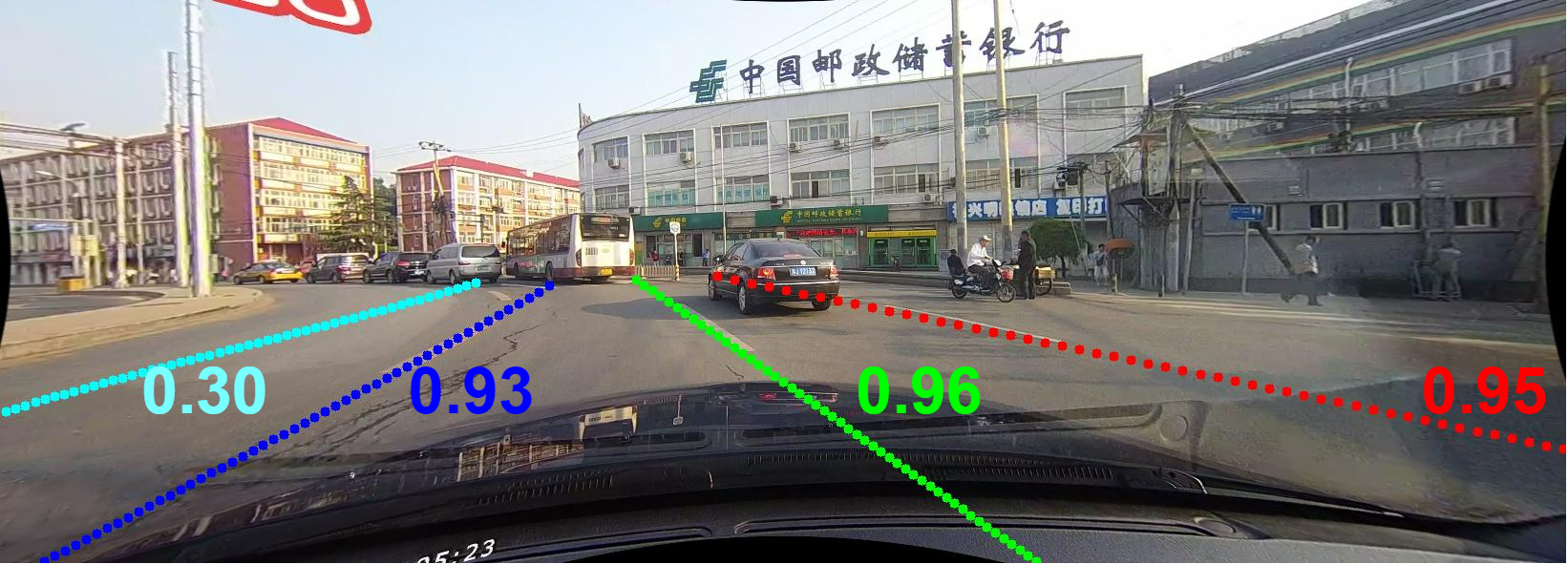}
    }
    \caption{The comparison between our DHPM and BézierLaneNet~\cite{bezier}. Compared with BézierLaneNet~\cite{bezier}, our DHPM can produce topologically and spatially high-quality lane proposals. In addition, the confidence gap between high-quality and low-quality proposals is enlarged by our proposed quality-aware constraints, where a threshold can be easily found to distinguish between true and false predictions with less effort.}
    \label{fig:fusion_strategy}
    \vspace{-6mm}
\end{figure}

\emph{First, how to increase the number of TP (i.e., true positive) predictions?}
Many methods~\cite{culane,UFAST,RESA,ENet-SAD,CondLaneNet,xr18} choose to predict a small number of proposals, which equals the maximum number of annotations of a single image. This indicates that the representations of these proposals must have enough generalization performance to adapt to various situations. However, there exist many complex-shaped lane lines in real scenes, even including some invisible lanes, which limit the overall detection performance. 
Therefore, we believe it is necessary to produce more proposals to deal with varied situations, which have been widely adopted by many object detection models~\cite{DETR,hu2018relation}. Nevertheless, existing methods~\cite{bezier,LSTR} only supervise a subset of proposals with high classification scores. In other words, these methods typically give no clear location or shape supervision to the low-scoring proposals, let alone the underlying relations between proposals. For example, we show the 50 proposals predicted by the top-performing BézierLaneNet~\cite{bezier} in Fig.~\ref{fig:fusion_strategy}~(c), where each proposal is represented as $100$ key points. We observe that a vast majority of proposals are with unreasonable shapes or improper locations, which indicates that these proposals have a very low probability of being optimized during training, and these unreasonable proposals have no potentials to fit lane lines at the test phase. That means a small number of proposals will be supervised to fit all lanes, which is very challenging for the learned proposal representation.
As a consequence, even the well-trained proposals cannot handle challenging samples and produce very confident lane proposals at the test phase, as shown in Fig.~\ref{fig:fusion_strategy}~(e). This observation motivates us to densely modulate all lane proposals w.r.t. ground truth labels since the increasing number of high-quality proposals are beneficial to raise the number of TP predictions. 

\emph{Second, how to decrease the amount of FP (i.e., false positive) and FN (i.e., false negative) predictions?} Most lane detection methods~\cite{bezier,LSTR,UFAST,RESA,culane} restrain the proposal quality (shape and location accuracy) and instance-level discrimination individually. To be specific, they typically use a standard binary cross-entropy or multi-class cross-entropy loss as the discrimination constraint. However, these methods will inevitably produce some proposals with high scores but inaccurate shapes or 
locations, resulting in harmful effects on the discriminative power. To enhance the instance-level discrimination and decrease false classification, we propose to adaptively assign quality-aware classification labels to all proposals, which aims to give relatively lower classification labels to a low-quality proposal and vice versa.

To address these, we propose a dense hybrid proposal modulation method to generate topologically and locational high-quality lane proposals with discriminative representations. Specifically, our hybrid modulation can be divided into three aspects: 1) availability constraint, 2) diversity constraint, and 3) discrimination constraint. As mentioned above, only a tiny fraction of lane proposals are sufficiently supervised in shape and location, which leads to inaccurate predictions and poor generalization ability of most existing methods. Therefore, our prime objective is to make each proposal appear around the ground truth lanes with a reasonable shape. To avoid over-curved proposals in Fig.~\ref{fig:fusion_strategy}~(c), we define a simplified curvature metric to measure and control the degree of bending of curves. Then we force each proposal to approach its specific locating target assigned by our proposal-to-label matching algorithm. In addition to availability constraints, we encourage topological diversification of the proposals targeting the same ground-truth lanes, aiming to improve the fitting ability in the training phase and the generalization ability in the testing phase, respectively. To achieve this, we efficiently measure and regulate intra-cluster shape differences by extending our single-curve metric into a pairwise one (see Fig.~\ref{fig:fusion_strategy}~(d)). 
Last but not least, we propose to enhance the instance-level discrimination of proposals to reduce the number of false predictions. Instead of a standard cross-entropy loss, we build a quality-aware classification objective by explicitly taking shape and location accuracy into consideration. With this quality-aware design, our method can redistribute the probability distribution of existence classification, where a threshold can be easily found to classify all proposals into high- and low-quality groups. Fig.~\ref{fig:fusion_strategy}~(f) shows that our method tends to give relatively higher scores to good proposals while the score gap between good and bad proposals is enlarged. Since our method modulates the proposals with three constraints without modifying the network structure, it only increases the time of back-propagation, while the forward propagation remains unchanged. Our DHPM achieves very competitive performance without introducing new parameters and reducing inference speed on four widely-used datasets.

The main contributions can be summarized as follows:

\begin{itemize}
    \item[1)] Unlike existing lane detection methods, which either predict a few proposals or sparsely supervise a small fraction of proposals, we propose a dense hybrid proposal modulation method to improve the overall quality of proposals.
    \item[2)] We first present the availability constraint to locate each proposal around the ground truth lanes with a reasonable shape. Then we enhance the inter-proposal topological diversification to exploit the underlying relations between proposals with our diversity constraint.
    \item[3)] We propose a quality-aware discrimination constraint to build a bridge between the proposal quality (including topological and locational accuracy) and instance-level discrimination. 
    \item[4)] Experimental results on four popular datasets demonstrate that our DHPM achieves very competitive performance without introducing new parameters, and cross-dataset experiments show that DHPM significantly improves the generalization ability of proposals compared to the baseline.
\end{itemize}

The remainder of this paper is organized as follows: Section~\ref{sec:related} describes recent studies related to our work. Section~\ref{sec:model} introduces the crucial components of our DHPM. Extensive experimental results on various datasets are reported in Section~\ref{sec:experiment}, and the conclusion of this paper is in Section~\ref{sec:conclusion}.

\section{Related Work}
\label{sec:related}

In this section, we briefly review three related topics, including lane detection, high-quality proposals learning, and proposal relations learning.

\subsection{Lane Detection Strategies}
    \subsubsection{Segmentation-based Methods} 
        Segmentation-based methods~\cite{xr18,ENet-SAD,Lanenet,CasCNN,culane,RONELDv2,YOLOP,ContinuityLearner,LaneAF,SEFG,SANNet,E2ELMD,SUPER,Inter-Region,ONCE-3DLanes} view lane detection as a per-pixel classification task. To handle the classification problem of lane line points, some prior arts~\cite{Inter-Region} treat each lane line as a category for segmentation. In some methods, pixels are classified as either on lane or background to generate a binary segmentation mask. LaneNet~\cite{Lanenet} and LaneAF~\cite{LaneAF} predict the instance segmentation mask, and use post-clustering strategy~\cite{cluster} to determine the final lane instance. However, segmentation-based methods are limited by a predefined and fixed number of lanes, which is not robust to real driving scenarios where the number of lanes is unknown.
        
    \subsubsection{Anchor-based Methods}
        Anchor-based Methods~\cite{UFAST,PointLaneNet,CurveLane,CondLaneNet,LaneATT,Eigenlanes,CLRNet,2021Structure,SegSelfAttent,persformer} use the predefined anchor to help describe the lane line. Instead of predicting lane line points pixel by pixel, UFAST~\cite{UFAST} manages to split the images into row-wise anchors, where the number of anchors is much smaller than the pixels, so it can achieve a good trade-off between performance and efficiency. To minimize the gap between anchor-based prediction and ground truth, some works~\cite{PointLaneNet,CurveLane,CondLaneNet,LaneATT} also predict an offset map to refine the initial prediction. LaneATT~\cite{LaneATT} designs a line-shape anchor according to the slender and monotonous characteristics of the lane line, which is followed by many recent anchor-based methods~\cite{persformer}. CLRNet~\cite{CLRNet} leverages the high-level semantic features and low-level texture features to refine the proposal representations.
        However, the fixed anchor shape results in a low degree of freedom in describing the complex topology.
         
    \subsubsection{Curve-based Methods}     
        Curve-based methods~\cite{PolyLaneNet,LSTR,bezier,E2ECurfit,UltraLow}  directly output parametric lines expressed by curve equation.
        PolyLaneNet~\cite{PolyLaneNet} fits all lane lines into polynomial functions, and directly predicts the fitting curve parameters through a deep network. LSTR~\cite{LSTR} introduces the transformer with a similar structure to DETR~\cite{DETR} into the curve-based method, and redesigns the fitting curve to take the camera information into account. 
        Different from segmentation-based methods and anchor-based methods, curve-based methods can avoid the separation between the predicted adjacent lane line points. Although the accuracy of curve-based methods still lags behind that of other methods, they have been proved~\cite{Robustness} to be more robust due to the smoothness of the fitting curve.

\subsection{High-Quality Proposals Learning}
        In the field of object detection, many methods have made progress in generating high-quality bounding box proposals. Cai~\emph{et al.}~\cite{Cascade-R-CNN} prove that high-quality proposals can bring high-quality detector. Some works improve the quality of proposal through ingeniously region proposal network design, for example, FPN~\cite{FPN} uses both high-resolution low-level features and high semantic information of high-level features to improve the quality of multi-scale proposals. Besides, LocNet~\cite{LocNet} iteratively recognizes and locates the proposal boxes to generate higher-quality proposal boxes. Cascade RCNN~\cite{Cascade-R-CNN} gradually improves the quality of proposal boxes through a designed cascade regressor, in which each regressor will optimize the output proposals according to the previous regressor. In weakly supervised object detection tasks, PG-PS~\cite{High-Quality-WSOD} combines selective search and Grad-Cam~\cite{selvaraju2017grad} to perform coarse classification and fine classification in turn to obtain high-quality proposal boxes. 
        Recently, many lane detection works~\cite{LSTR,bezier} predict proposals that are significantly larger than the number of lane lines. However, they only supervised a subset of high-scoring proposals, ignoring the overall quality. In this paper, we propose a dense hybrid modulation mechanism to improve the overall quality of proposals.

\subsection{Proposal Relations Modeling} 
    Modeling proposal relations has emerged in many fields, including 2D~\cite{P-GCN,RDN,SmallObject} and 3D object detection~\cite{Feng}. In the field of action localization, P-GCN~\cite{P-GCN} constructs a graph of proposals by establishing the edges and then applying GCNs to do message aggregation among proposals. For 3D object detection, Feng \emph{et al.}~\cite{Feng} extract uniform appearance features for each 3D object proposal and construct a relation graph that exploits the 3D object-object relationships. In order to explore and utilize object-to-object relations, RDN~\cite{RDN} assembles and propagates object relations to enhance object features for video object detection. Despite the rapid development of modeling the proposal relations of 2D or 3D bounding boxes, the relations between lane proposals have not been well exploited yet. In our work, we aim to mine the location and score relations between proposals to generate diverse and discriminative proposals.

  \begin{figure*}[t]
      \centering
      \includegraphics[scale=0.09]{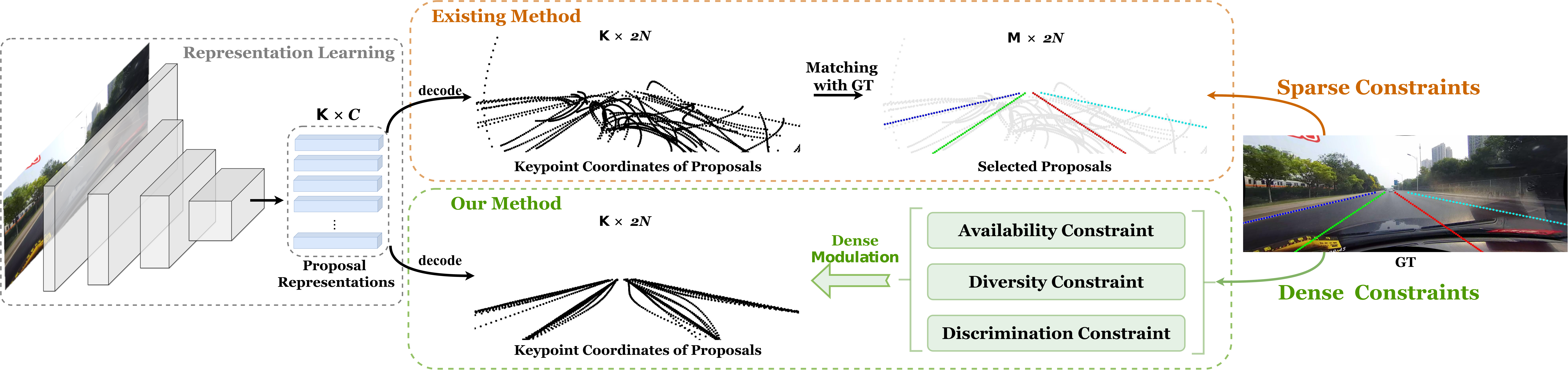}
      \caption{Comparison of our overall framework with existing methods. Unlike existing methods that sparsely supervise a subset of proposals with high classification scores, we densely modulate all proposals to generate topologically and spatially high-quality lane proposals with three constraints, i.e., availability, diversity, and discrimination.}
      \label{fig:full_pipeline}
      \vspace{-4mm}
      \end{figure*}

\section{The Proposed Approach}
\label{sec:model}
In this section, we first formulate the problem of lane detection and then describe our main idea. At last, we detail the overall objective function for lane detection.

\subsection{Overview}
Given an input image \(I\in \mathbb{R}^{H\times W \times 3}\) with the corresponding ground truth $G = \left \{ g_1, g_2, ..., g_M \right \}$, our goal is to predict a collection of lane proposals \(P=\left \{ p_1, p_2, ..., p_K \right \}\), and predict a confidence value for each lane lines \(L=\left \{ l_1, l_2, ..., l_K \right \}\), where \(K\) is the total number of lane lines. Generally, a lane proposal \(p_k\) is represented by an ordered set of coordinates $p_k=\left\{(x_{1}^{(k)}, y_{1}^{(k)}), (x_{2}^{(k)}, y_{2}^{(k)}), \cdots, (x_N^{(k)}, y_N^{(k)})\right\}$, where \(k\) is the index of proposal lane line and \(N\) is the max number of sampled points. Similarly, each ground truth lane line \(g_j\) is also represented as \(N\) points $g_j=\left\{(u_{1}^{(j)}, v_{1}^{(j)}), (u_{2}^{(j)}, v_{2}^{(j)}), \cdots, (u_N^{(j)}, v_N^{(j)})\right\}$, where \(j\) is the index of ground truth lane line. 
The objective of a general curve-based method can be divided into three terms: 1) a coordinate regression term $J_{reg}$; 2) an existence classification term $J_{cls}$; and 3) an optional binary mask segmentation term $J_{seg}$. Therefore, the overall loss can be written as a weighted sum of all terms:
\begin{align}
    J_{sparse} = \lambda_{1}J_{reg} + \lambda_{2}J_{cls} + \lambda_{3}J_{seg},
    \label{eq:losssparse}
\end{align}
where $J_{reg}$ can be $\ell1$, $\ell2$, smooth $\ell1$ and other regression-based loss. Note that existing methods~\cite{bezier,LSTR} only apply $J_{reg}$ to the positive proposals, which depend on the number of ground truth lanes $M$. To achieve this, they perform a one-to-one assignment between $M$ labels and $K$ predictions ($M$ \textless $K$) using optimal bipartite matching~\cite{DETR}.
Assuming that $p_k$ is a positive proposal and its corresponding ground truth lane is $g_j$, we can represent the $\ell1$-based regression loss $J_{reg}$ as:
    \begin{align}
        J_{reg} = \frac{1}{2N}\sum\limits_{i=1}^{N}||y_i^{(k)}-v_i^{(j)}||+||x_i^{(k)}-u_i^{(j)}||.
    \label{eq:lossreg}
\end{align}
Different from $J_{reg}$, $J_{cls}$ and $J_{seg}$ in a general curve-based method are calculated for each proposal, so we omit the proposal index for clarity. For example, $J_{cls}$ can be written as follows:
    \begin{align}
        J_{cls} = -l\log(p) - w(1 - l)\log(1 - p),
    \label{eq:lossclass}
\end{align}
where $l$ and $p$ represent the existing label and predicted logits of a proposal, respectively, and $w$ is a hyper-parameter for weighting negative samples. In addition, the binary segmentation loss $J_{seg}$ shares the same format with $J_{cls}$.

While existing curve-based methods have made significant progress by extending the general model introduced above, their performance is suppressed by the following limitations:
\begin{itemize}
    \item[1)]\textbf{Overall poor rationality of lane proposals.} Optimal bipartite matching algorithms such as the Hungarian algorithm~\cite{wang2021end} only select a subset of the high-scoring lane proposals during training. Due to insufficient topological and locational supervision, most proposals become very curly and have a very chaotic layout.
    \item[2)]\textbf{Lack of relations mining between proposals.} Existing methods independently supervise each matching proposal without considering the shape and location of other proposals. Therefore, there may exist some highly similar lane proposals, which have a detrimental effect on the generalization ability.
    \item[3)]\textbf{Weak discrimination of classification scores.} During training, traditional detection methods~\cite{AdaptiveRegion,BBC,HighQualityRCNN,Bayesian,Visual-Attention-Based} directly assign labels $1$ to matching proposals and $0$ to others. Recent lane detection methods continue to use such assignments, regardless of the quality (topological and locational accuracy) of matching proposals. This will inevitably generate some proposals with high confidence but inaccurate shapes or locations, increasing the difficulty of finding a suitable threshold.
\end{itemize}

\begin{figure}[t]
    \centering
    \includegraphics[width=3.5in,totalheight=1.2in]{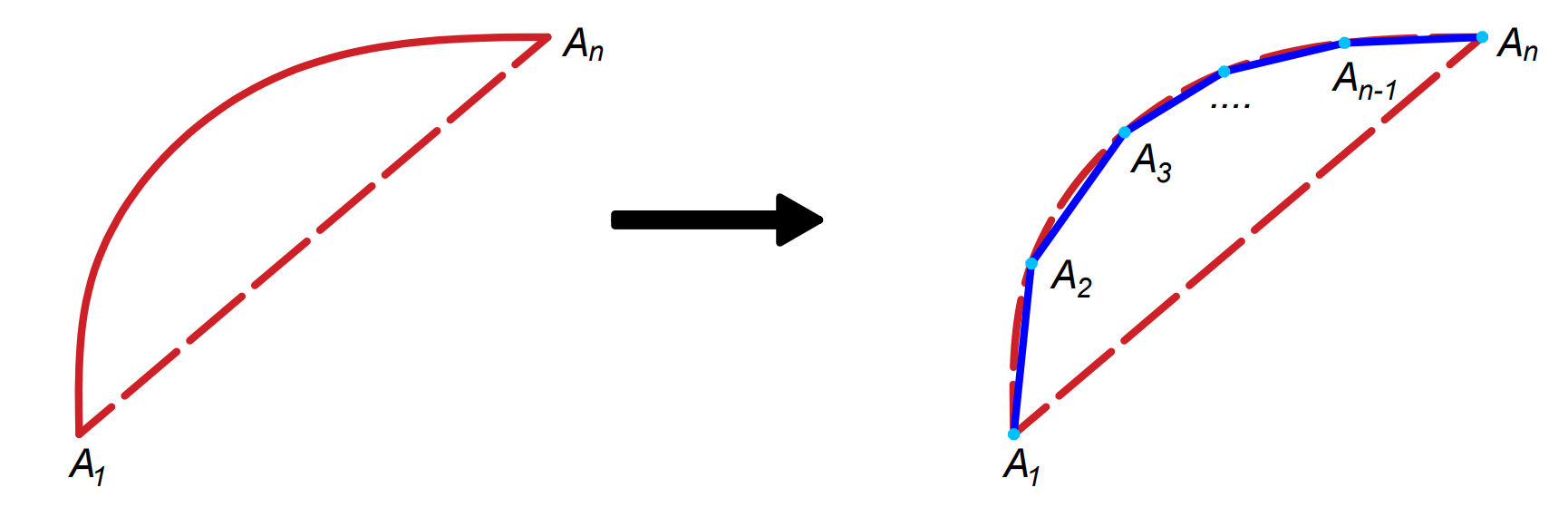}
    \vspace{-3mm}
    \caption{Illustration of the shape loss $S_k$.
    The length of the red dotted line represents the straight-line distance between the starting point and the ending point, and the solid red line represents the curve. Referring to the calculus idea, we use the total length of the blue dashed line to approximate the length of the curve with dense sampling.}
    \label{fig:shape_loss}
    \vspace{-4mm}
\end{figure}

To make lane proposals topologically and locational accurate and discriminative, we enforce three important criteria to modulate these proposal descriptors:
\begin{itemize}
    \item[1)]\textbf{The modulated lane proposals are topologically and spatially sound.} Realistic lane lines usually have strong prior knowledge, e.g., the ending points of different lane lines are usually close to each other, while the starting points are far apart. Therefore, forcing them to be reasonable can increase the number of true positive predictions.
    \item[2)]\textbf{The modulated proposals are with good generalization ability.} There are a few lane lines with complex topology, including bumps and grooves, and some lane lines are even invisible. 
    We need to ensure that there are sufficient and diverse proposals near ground-truth lanes as alternative predictions.
    \item[3)]\textbf{The modulated proposal confidences are closely linked to the quality of proposals.} Since binary cross-entropy loss does not guarantee higher classification labels for higher-quality proposals, it is important to build a bridge between the quality of proposals and the classification labels. By doing this, we can determine a threshold in the testing phase to distinguish between true and false predictions with less human effort.
\end{itemize}

To achieve these objectives, we present DHPM method to generate topologically and spatially high-quality lane proposals with discriminative representations, which is illustrated in Fig.~\ref{fig:full_pipeline}. Without introducing new parameters, our DHPM achieves very competitive performance with three simple yet effective constraints. A light weight version of our method could achieve 200+ FPS with a higher performance over the baseline~\cite{bezier}.

\subsection{Availability Constraint}

The goal of availability constraint is to make all proposal lane lines reasonable. Specifically, we separate the objective into two aspects. On the one hand, the topology of a lane proposal should not be too much complicated than a straight line, which is based on the structure priors in the driving scenarios. Therefore, we hope to find a way to measure and control the curvature of a lane curve efficiently.
On the other hand, the lane proposals are expected to locate near the ground truth lanes evenly. At the same time, the endpoints and starting points should follow the layout priors in real scenarios. Thus, we propose to assign a label lane for each proposal lane to give a specific regression target. With such two constraints, we can make sure that the modulated proposals are topologically and spatially sound. Note that we densely apply these two constraints to each lane proposal, which is fundamentally different from existing sparse constraint methods.
To sum up, we can formulate the full availability constraint as follows:
{
\begin{equation}
    \begin{aligned}
    J_{ava} =& \lambda_{shape} J_{shape} + \lambda_{loc} J_{loc}\\
            =& \lambda_{shape} \frac{1}{K}\sum\limits_{k=1}^{K}S_{k} + \lambda_{loc} \frac{1}{K}\sum\limits_{k=1}^{K}D_{k}, \\
    \end{aligned}
\end{equation}}
where $\lambda_{shape}$ and $\lambda_{loc}$ are weighting coefficients. $S_k$ is the shape loss of $k$-th proposal, while $D_k$ is the location loss.

In general, existing methods represent a lane proposal as an ordered set of coordinates. Therefore, the shape and location representations of a proposal are deeply coupled. We cannot directly apply our constraints on the coordinates because
a coordinate perturbation will change the shape and location simultaneously. To disentangle shape attributes from coordinates, our first step is to densely sample $n$ key points from a lane proposal, and we can re-write a lane representation as $\overset{\frown}{A_1A_n}=\left\{{A_1, A_2, \cdots, A_{n-1}, A_{n}}\right\}$, where we denote $A_1$ and $A_n$ as the start and end points, respectively. Based on these dense points, we can approximate the curve length with the length of multiple line segments when $n \xrightarrow{} \infty$. We present a straightforward and parallel-friendly way to efficiently measure the curvature of lane proposals. That is, the ratio of the length of the curve $|\overset{\frown}{A_1A_n}|$ to the length of the straight line $|A_1A_n|$, as shown in the Fig.~\ref{fig:shape_loss}. Mathematically, we can represent the simplified curvature term as:
{\small
\begin{equation} 
    \begin{aligned}
    S_{k} = \frac{|\overset{\frown}{A_1A_n}|}{|A_1A_n|} \approx \frac{|A_1A_2|+|A_2A_3|+...+|A_{n-1}A_n|}{|A_1A_n|}\\
            = \frac{\sum\limits_{i=1}^{n-1} \sqrt{(x_{i+1}^{(k)} - x_{i}^{(k)})^2+(y_{i+1}^{(k)} - y_{i}^{(k)})^2}}{\sqrt{(x_{n}^{(k)} - x_{1}^{(k)})^2+(y_{n}^{(k)} - y_{1}^{(k)})^2}},
    \label{eq:shapeloss}
    \end{aligned}
\end{equation}}
where the minimum value of $S_k$ is 1, which is only satisfied when the lane proposal is a strictly straight line.

\begin{figure}[t]
    \centering
    \subfigure[Proposals without $J_{ava}$]{
    	\includegraphics[width=3.1in,totalheight=1.1in]{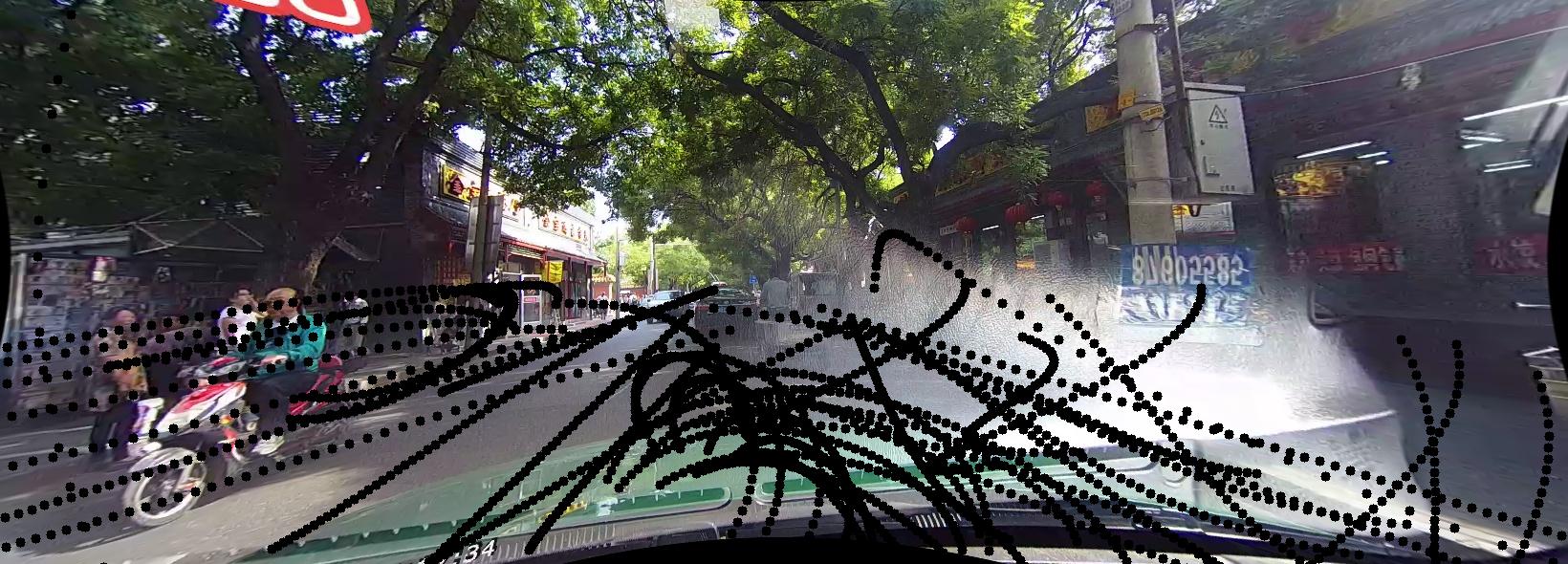}
    }
    \vspace{-3mm}
    \hspace{-3mm}
    \subfigure[Proposals with only $J_{loc}$]{
    	\includegraphics[width=3.1in,totalheight=1.1in]{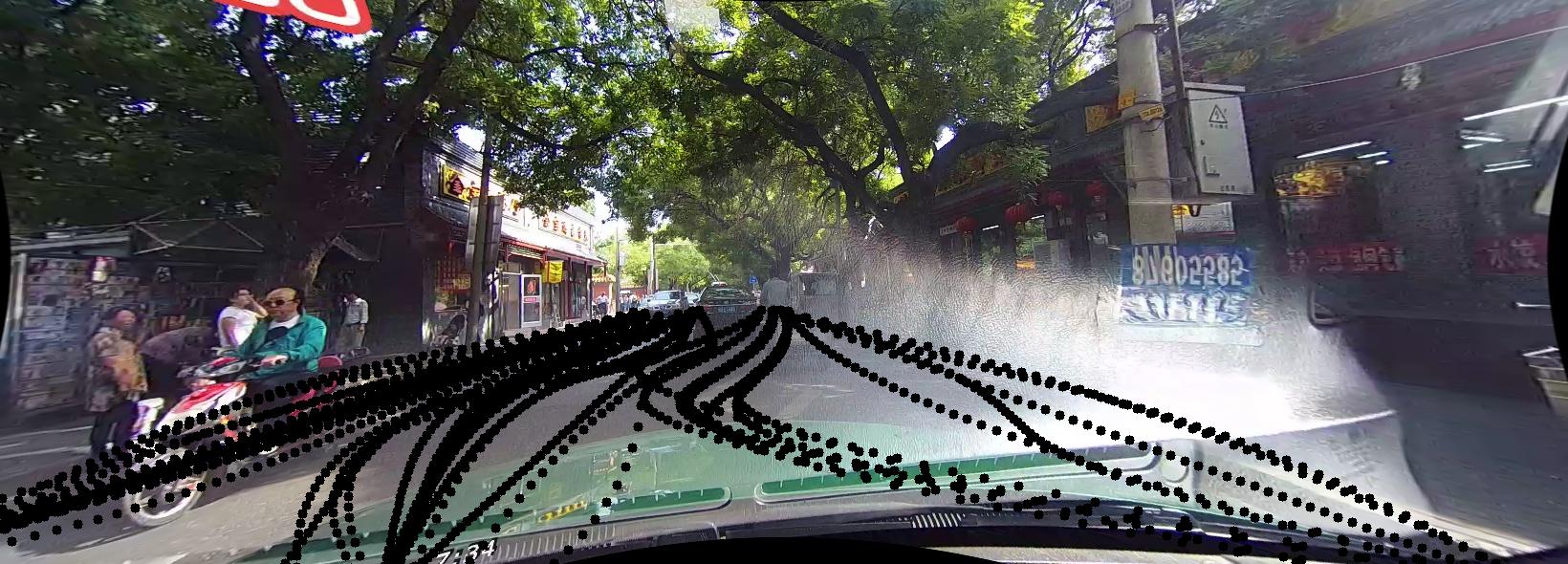}
    }
    \vspace{-3mm}
    \subfigure[Proposals with only $J_{shape}$]{
    	\includegraphics[width=3.1in,totalheight=1.1in]{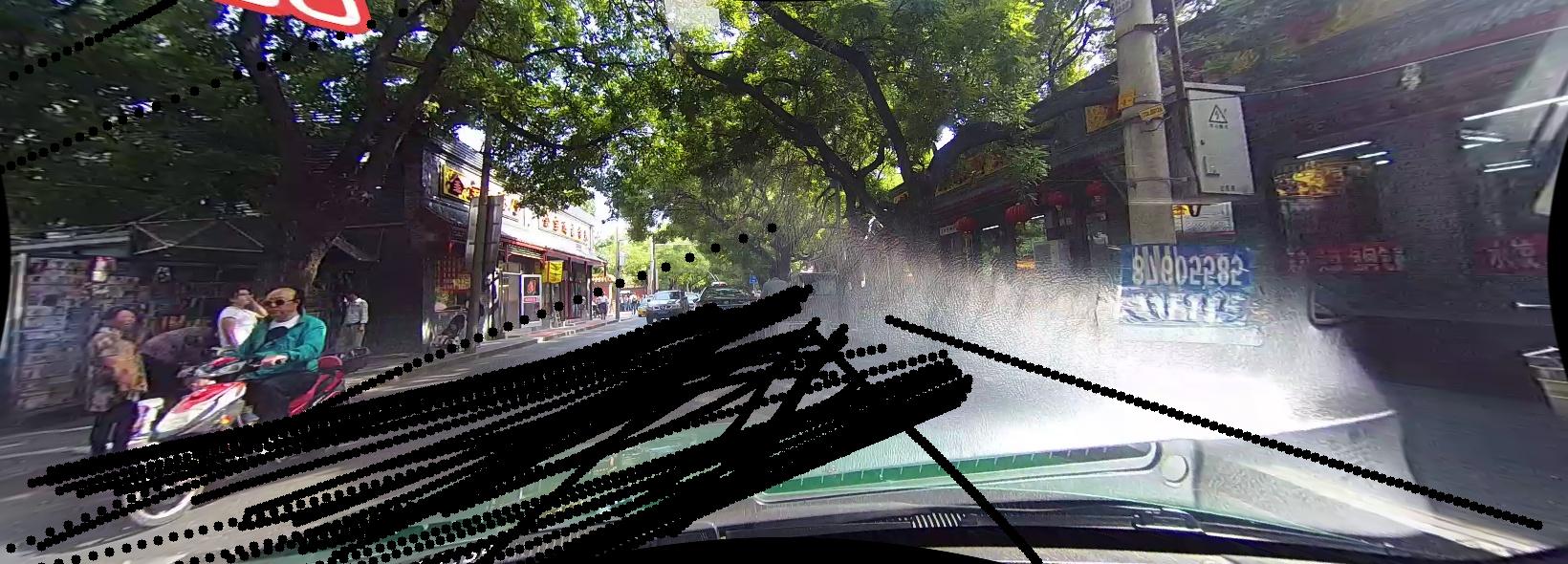}
    }
    \hspace{-3mm}
    \subfigure[Proposals with full $J_{ava}$]{
    	\includegraphics[width=3.1in,totalheight=1.1in]{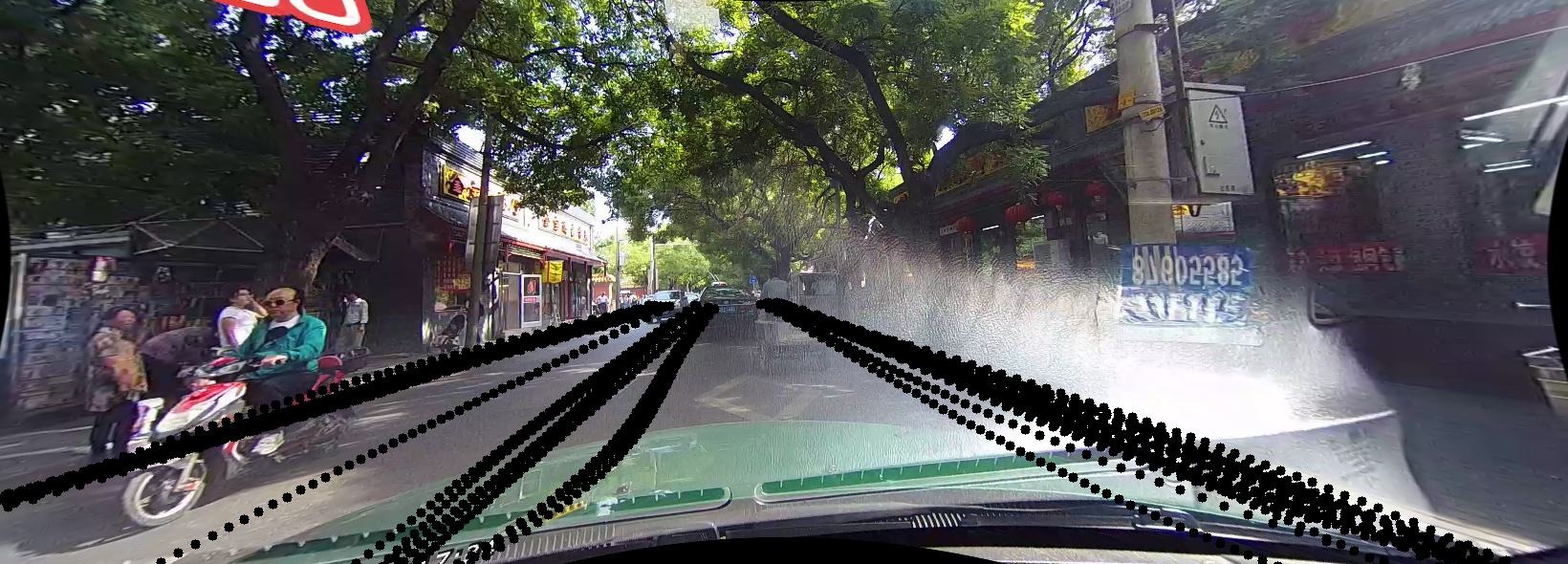}
    }
    \caption{Visual comparisons between different terms in the availability constraint. Compared to proposals without our $J_{ava}$ (a), applying $J_{loc}$ (b) or $J_{shape}$ (c) alone can slightly improve the overall quality, still leaving some unreasonable proposals. Differently, the full availability constraint that includes both shape and location supervision significantly improves the rationality of proposals. The image is from the test split of CULane~\cite{culane}.}
    \label{fig:availability}
    \vspace{-4mm}
\end{figure}

In addition to the shape properties, the spatial arrangement of the proposal should be explicitly guided by the layout prior. We propose to assign a target from ground truth lanes to each proposal so that the layout prior can be propagated to the lane proposals. To this end, we design a proposal-to-label matching algorithm to assign $M$ ground truth lanes to $K$ proposals. Instead of a slow clustering-based method, we only need to compute $M \times K$ $\ell1$ distances, from which each proposal will choose the closest ground truth lane as its target. Therefore, the ground truth lane that matches the $k$-th proposal can be computed as follows:
{\small
\begin{equation}
    \begin{aligned}
    g^{(k)} = \mathop{\arg\min}\limits_{g_i \in G}Q_{(i,k)},~~~ Q_{(i,k)} = 1-L_{1}(g_i,p_k),
    \label{eq:rematch}
    \end{aligned}
\end{equation}}

where $Q_{(i,k)}$ represents the distance between the $i$-th ground truth lane line with the $k$-th proposal. Note that all coordinates are normalized according to the image resolution. Having obtained the matched ground-truth lanes, we formulate the dense location constraints to all proposals as follows:
\begin{equation}
    \begin{aligned}
        D_k =& D_{s}^{(k)} +D_{e}^{(k)} \\
          =& ~~~~(|x_{1}^{(k)} - u_{1}^{(g^{(k)})}|+|y_{1}^{(k)} - v_{1}^{(g^{(k)})}|) \\
           & + (|x_{n}^{(k)} - u_{n}^{(g^{(k)})}|+|y_{n}^{(k)} - v_{n}^{(g^{(k)})}|),
    \end{aligned}
    \label{equ:location_loss}
\end{equation}
where $D_s$ and $D_e$ represent the $\ell1$ distances of starting points and ending points, respectively. For some samples that have no ground truth lanes, we simply omit the location loss while the shape loss is reserved.
It's worth noting that the proposed availability constraint is not the only constraint that supervises the location of proposals. 
The curve loss $J_{reg}$ in basic constraints also provides sparse locational supervision to a subset of proposals. Differently, our location constraints supervise all proposals with clear locational targets.

\begin{figure*}[t]
     \centering
        \setlength{\tabcolsep}{0.8pt}
        \begin{tabular}{ccccc} 
        \includegraphics[width=1.4in,totalheight=0.6in]{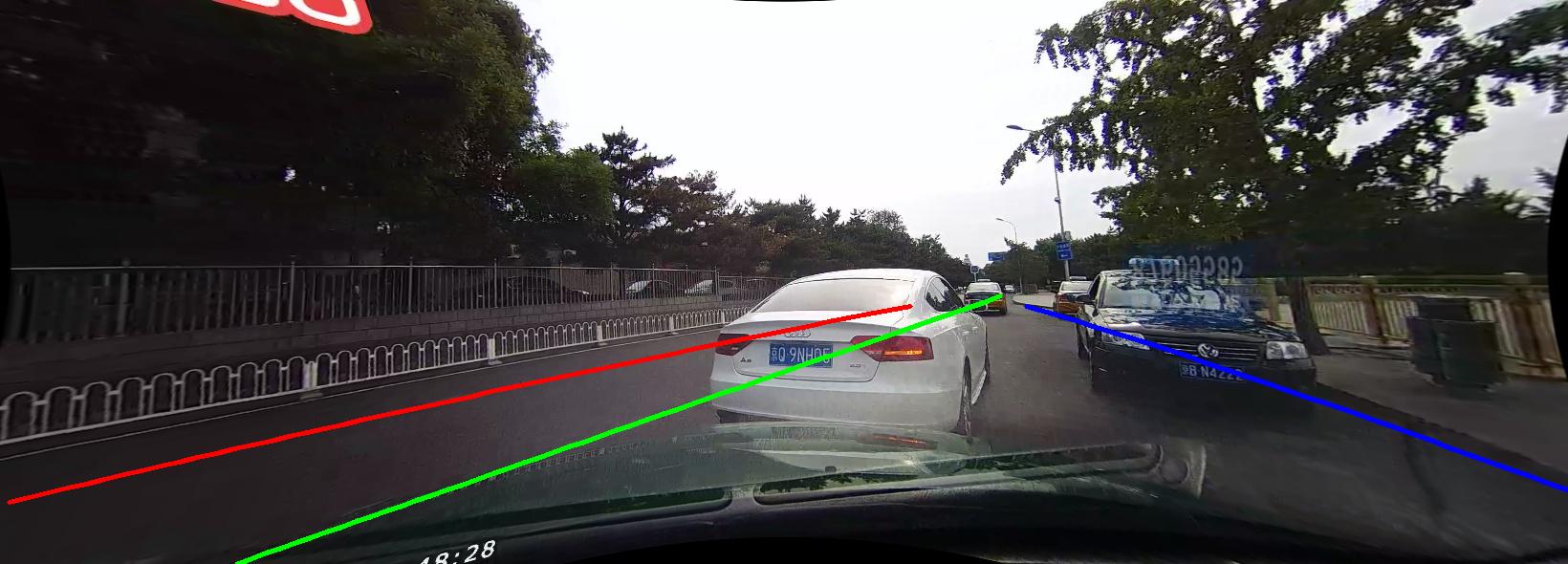}&
        \includegraphics[width=1.4in,totalheight=0.6in]{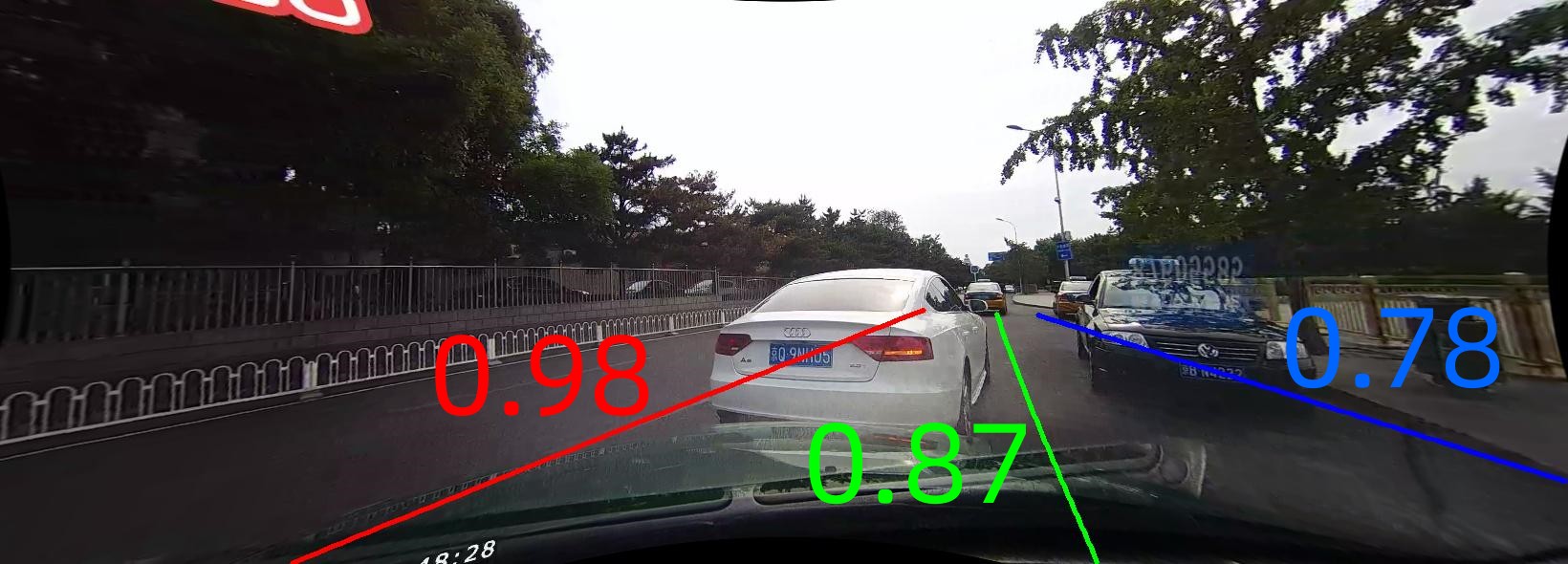}&
        \includegraphics[width=1.4in,totalheight=0.6in]{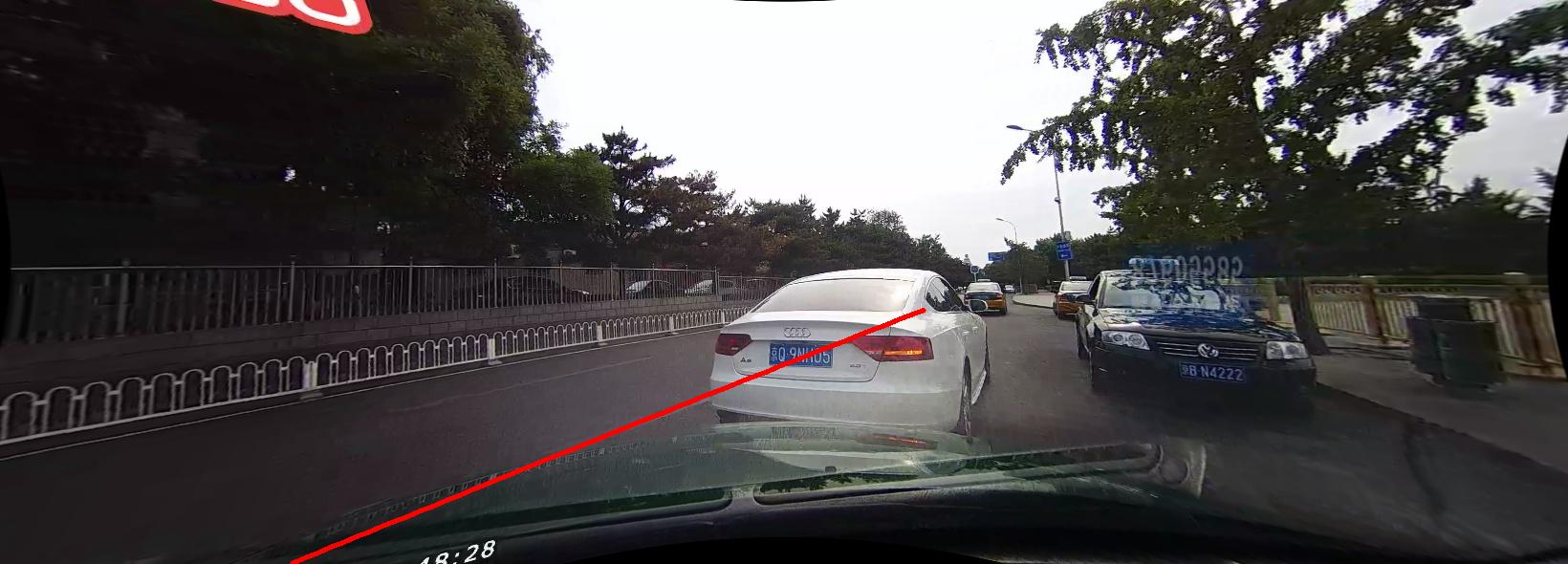}&
        \includegraphics[width=1.4in,totalheight=0.6in]{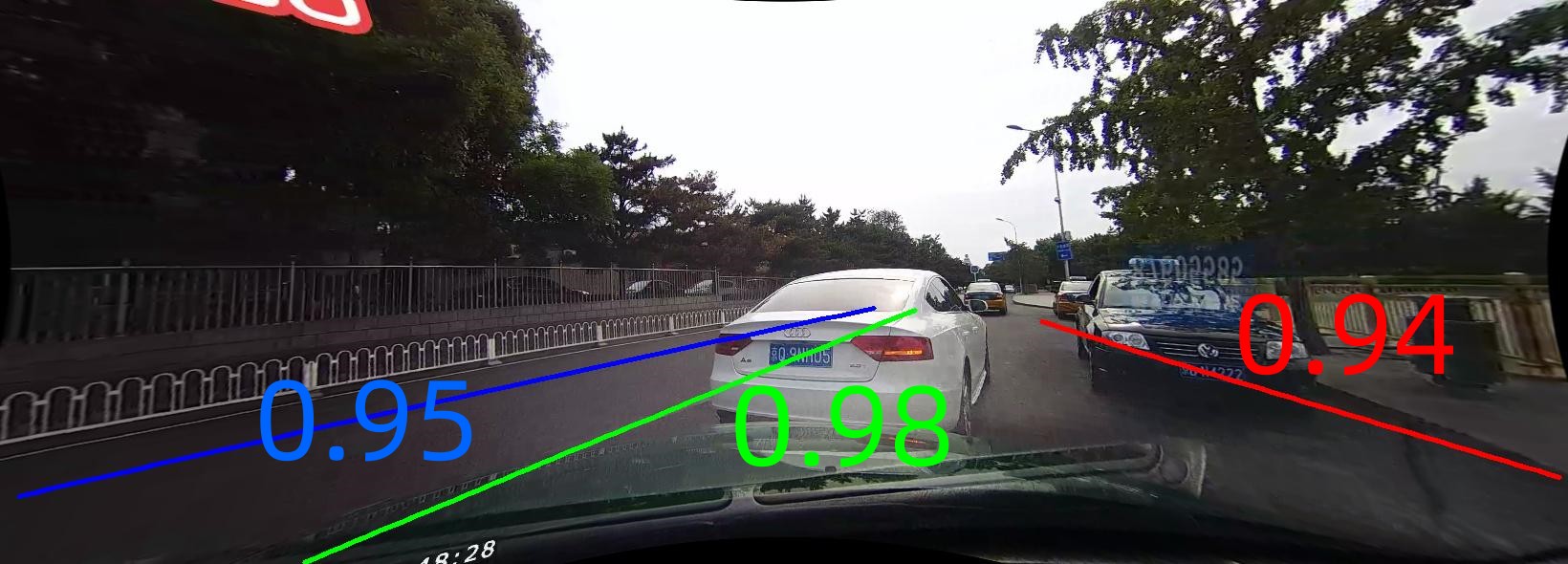}&
        \includegraphics[width=1.4in,totalheight=0.6in]{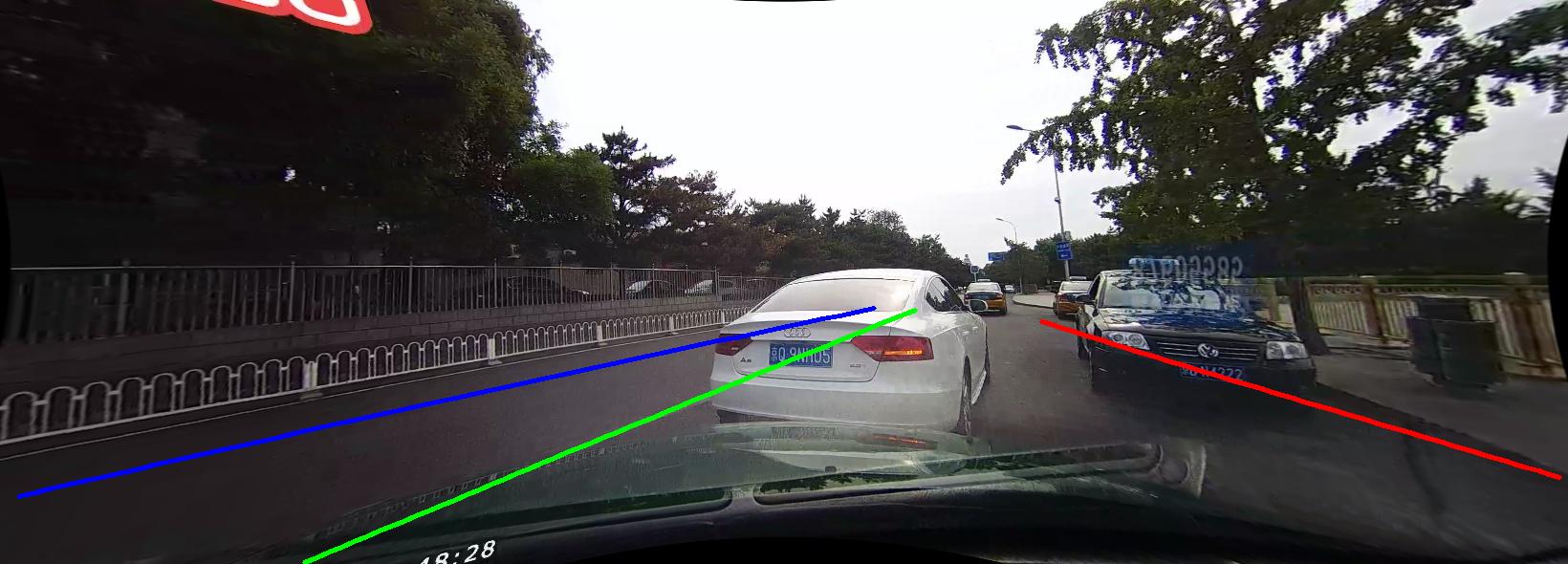}\\
        \includegraphics[width=1.4in,totalheight=0.6in]{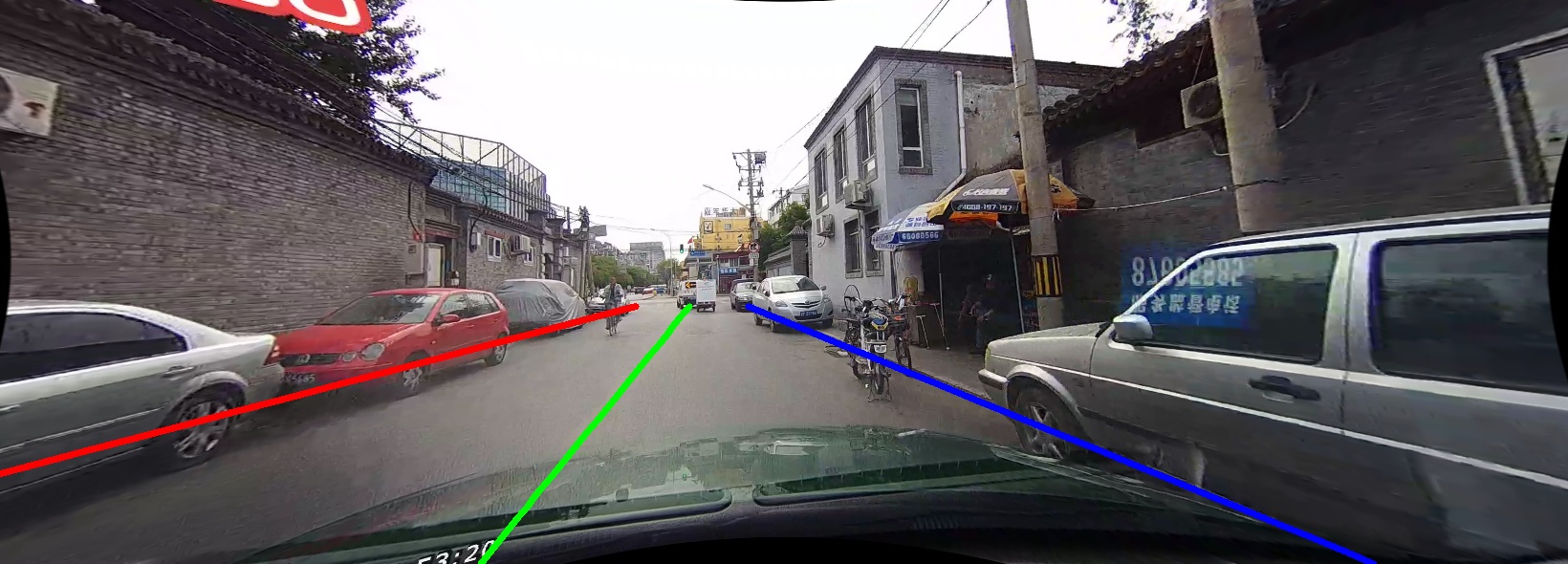}&
        \includegraphics[width=1.4in,totalheight=0.6in]{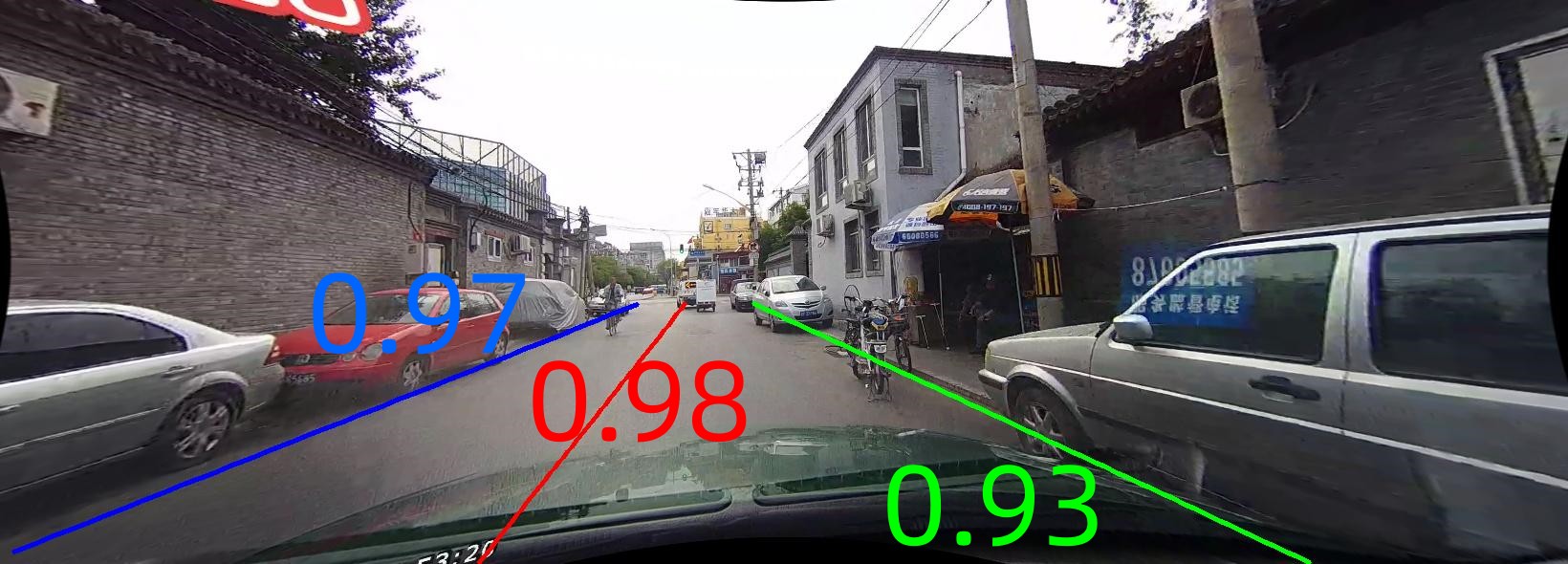}&
        \includegraphics[width=1.4in,totalheight=0.6in]{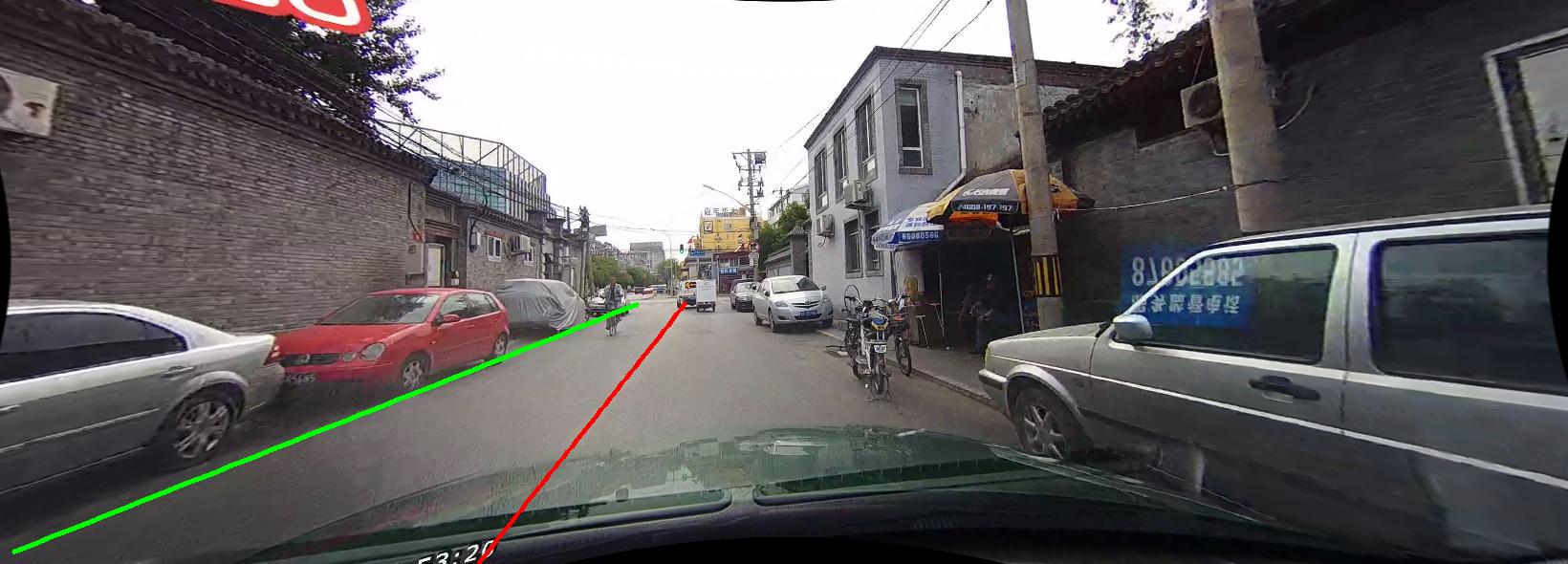}&
        \includegraphics[width=1.4in,totalheight=0.6in]{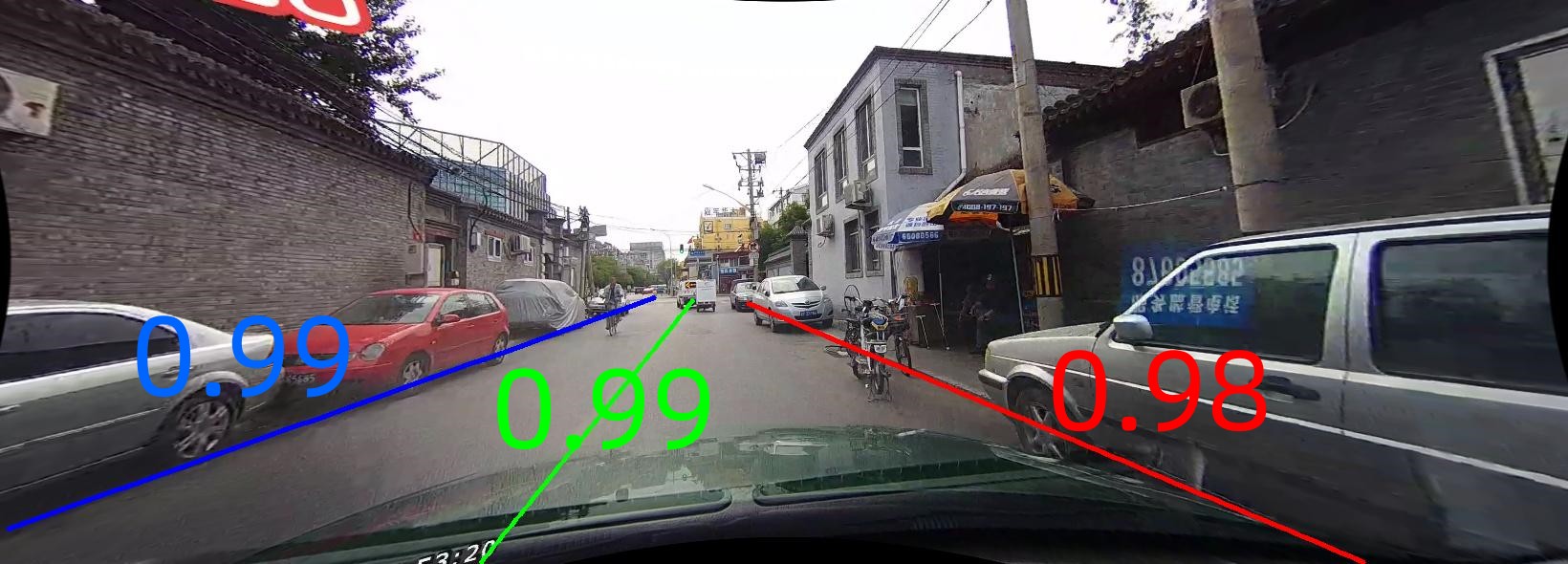}&
        \includegraphics[width=1.4in,totalheight=0.6in]{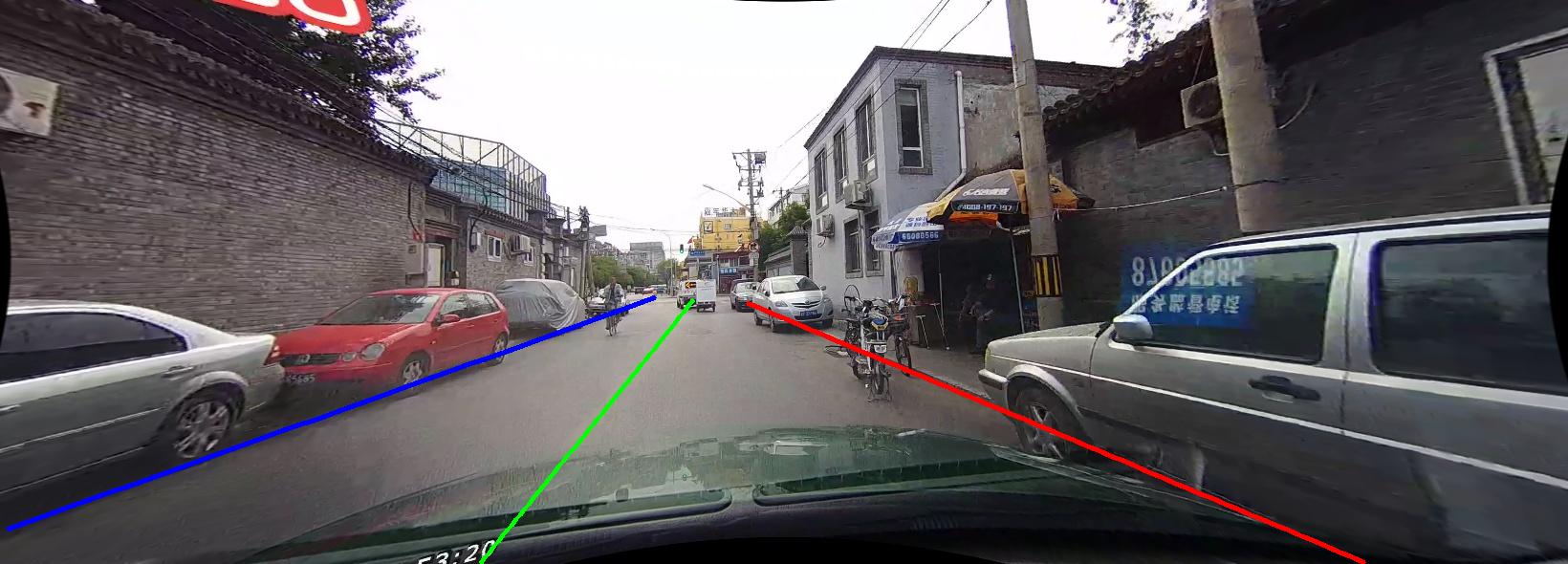}\\
        GT  &   baseline confidence  &  baseline result & confidence with $J_{ava}$   &  result with $J_{ava}$  \\
    \end{tabular}
    \caption{Proposal comparisons between the baseline and the model with only $J_{ava}$. Our $J_{ava}$ is used to increase the high-quality proposals (potential True Positive results), and reduce the bad proposals (useless predictions) at the same time.}
    \label{fig:confidence}
    \vspace{-3mm}
\end{figure*}

To illustrate the effect of each individual component of $J_{ava}$, we provide a detailed visual comparison in Fig.~\ref{fig:availability}. $J_{loc}$ or $J_{shape}$ cannot solely improve the proposals since they are completely decoupled components. On the contrary, the full $J_{ava}$ can effectively modulate most low-quality proposals to be topologically and spatially plausible. To sum up, our location constraint supervises the approximate location of the lane head and tail, while the shape constraint determines the lane neck. Therefore, it can be concluded that it is the combination of them that makes the lane have a reasonable location. Besides, Fig.~\ref{fig:confidence} shows the confidence comparisons between the baseline and the model with only $J_{ava}$, which proves that $J_{ava}$ successfully increases the high-quality proposals (potential True Positive results), and reduce the bad proposals (useless predictions) at the same time. Therefore, the proposals that exceed the threshold will have a greater probability of becoming a TP rather than an FP prediction, which proves that $J_{ava}$ can directly improve the detection performance.

\begin{figure}[t]
    \centering
    \subfigure[Great shape difference]{
    	\includegraphics[width=3in]{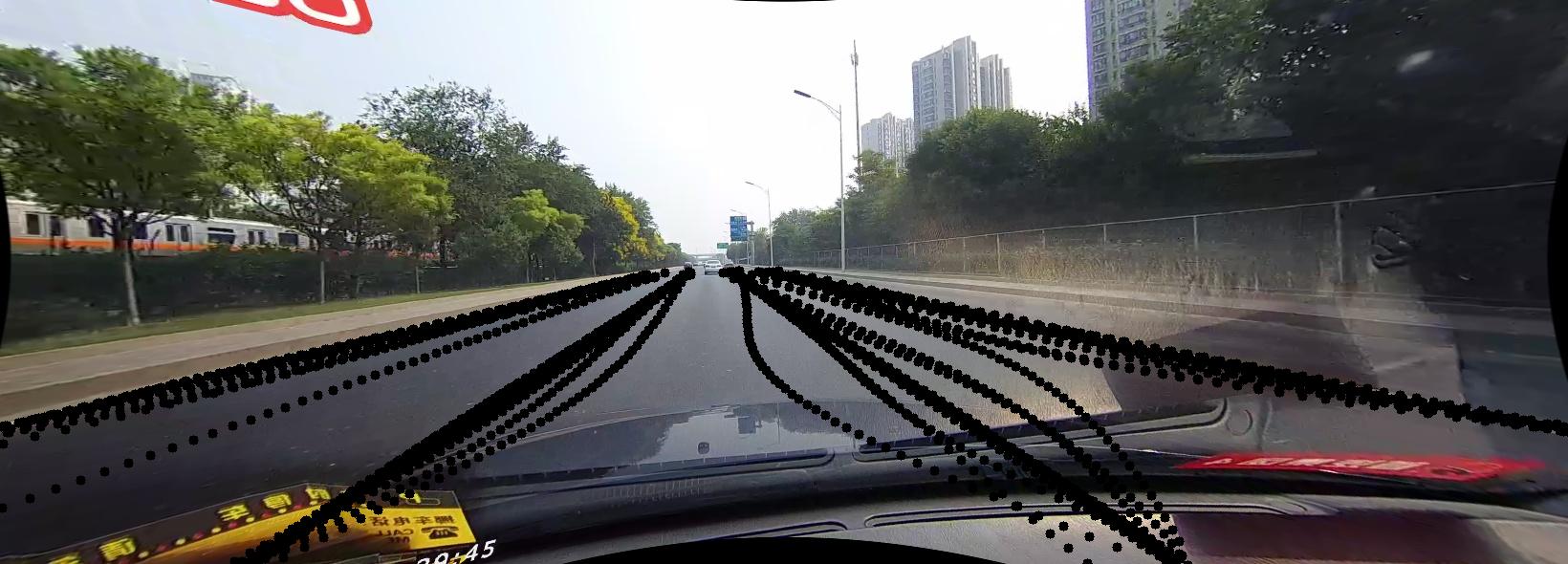}
    }
    \hspace{-3mm}
    \subfigure[Small shape difference]{
    	\includegraphics[width=3in]{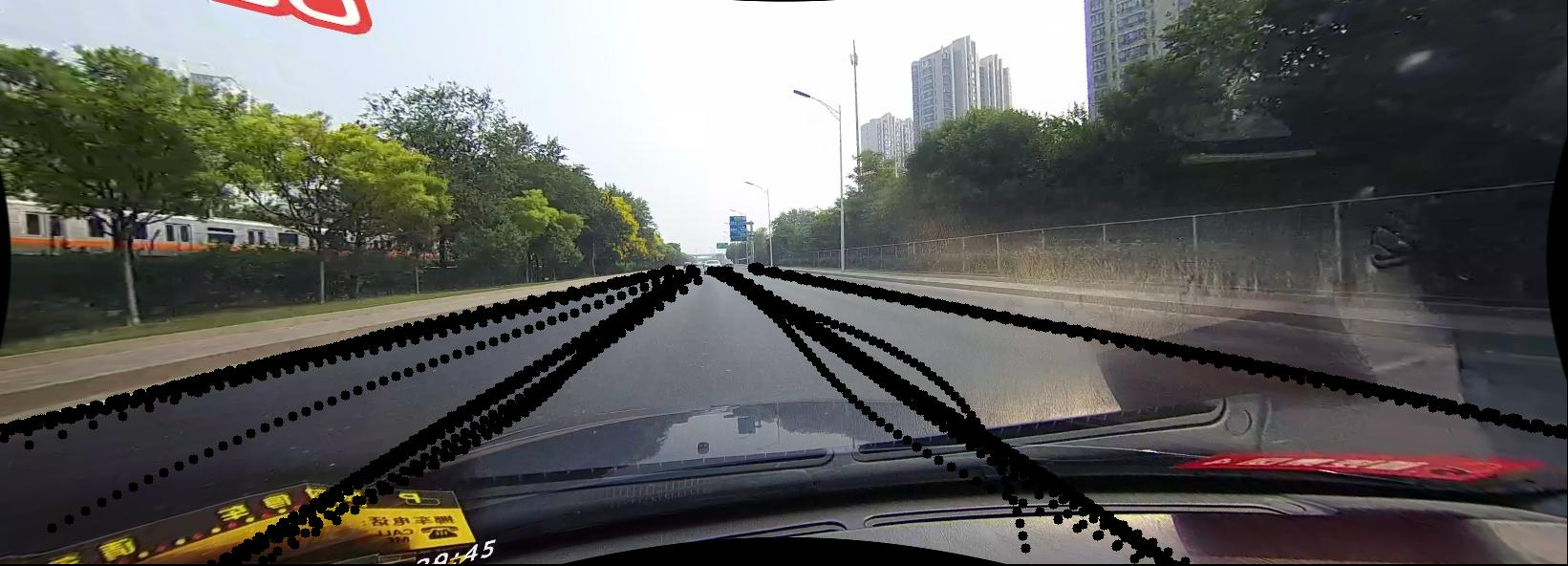}
    }
    \caption{The comparison between great and small differences between proposals. Compared to (b), greater shape difference (a) can enhance the generalization ability of proposals by reducing the mutual information between proposals.}
    \label{fig:difference}
    \vspace{-4mm}
\end{figure}

\subsection{Diversity Constraint}
Having obtained reasonable proposals by applying the availability constraint, we find that the proposals matching the same ground truth have a similar shape. The potential disadvantage can be summarized in the following two aspects. 1) The high similarity will reduce the generalization ability of the model since the unique information of each proposal is limited. 2) The discrimination of classification confidences will be suppressed because similar proposals may be given different binary labels.

To enhance the generalization ability of proposals and further model the underlying relations between proposals, we propose to reduce the mutual information between proposal lane lines. The mutual information between $p_i$ and $p_j$ can be written as:
\begin{equation}
    \begin{aligned}
    I(p_i,p_j) = I(p_i) - I(p_i|p_j),
    \end{aligned}
    \label{equ:mutual_infomation}
\end{equation}
where $I(p_i)$ means the information uncertainty of $p_i$, $I(p_i|p_j)$ means the uncertainty of $p_i$, given $p_j$ has already appeared, and they are calculated as follows:
\begin{equation}
    \begin{aligned}
    I(p_i) = log\frac{1}{P(p_i)},
    I(p_i|p_j) = log\frac{1}{P(p_i|p_j)},
    \end{aligned}
    \label{equ:probability}
\end{equation}
where we know the mutual information has a strong link with the probability between $p_i$ and $p_j$.

As shown in Fig.~\ref{fig:difference} (b), availability constraints lead to the high similarity between proposals and large mutual information values $I$. There are some overlapping proposals, which means that their representations are very similar to each other. Therefore, it is important to make each proposal unique since it can help generate more effective proposals to adapt to different situations. Unique proposals are able to better fit lane lines in their unique scenarios, thus increasing TP and reducing FP. However, directly using the distance between every two lane representations as mutual information makes it difficult to control the shape of decoded proposals.
Therefore, according to Equation~\eqref{equ:mutual_infomation} and \eqref{equ:probability}, we hope to reduce the mutual information between proposals by 1) enlarging the shape differences of proposals; 2) redistributing the probabilities of proposals. Although both the availability constraint and the diversity constraint aim to modulate the shape and location of the lanes, there is an essential difference between the two. The availability constraint $J_{ava}$ aims to supervise the location and shape attributes of a single lane proposal, while the diversity constraint $J_{div}$ is proposed to restrain the relationship of multiple lane proposals.
In this subsection, we will elaborate on how to enlarge the shape differences of proposals.

\begin{figure}[t]
    \centering
    \subfigure[with the upper limit]{
    	\includegraphics[width=3in]{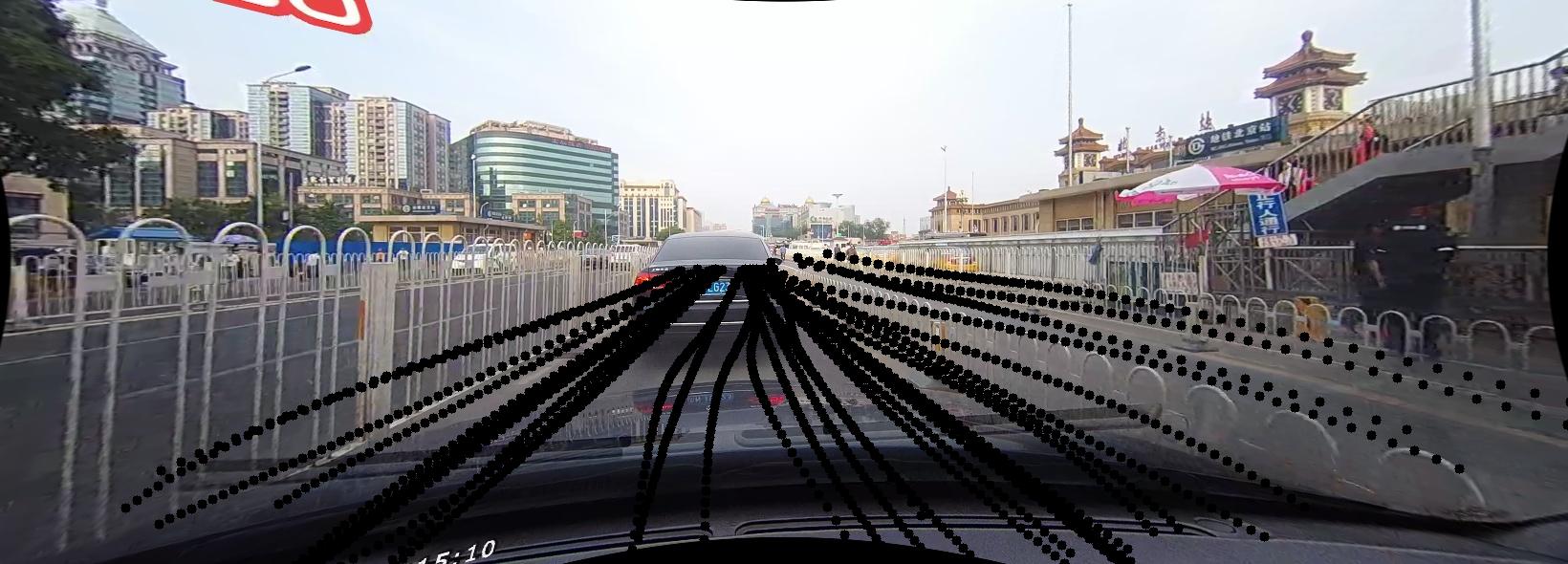}
    }
    \hspace{-3mm}
    \subfigure[without the upper limit]{
    	\includegraphics[width=3in]{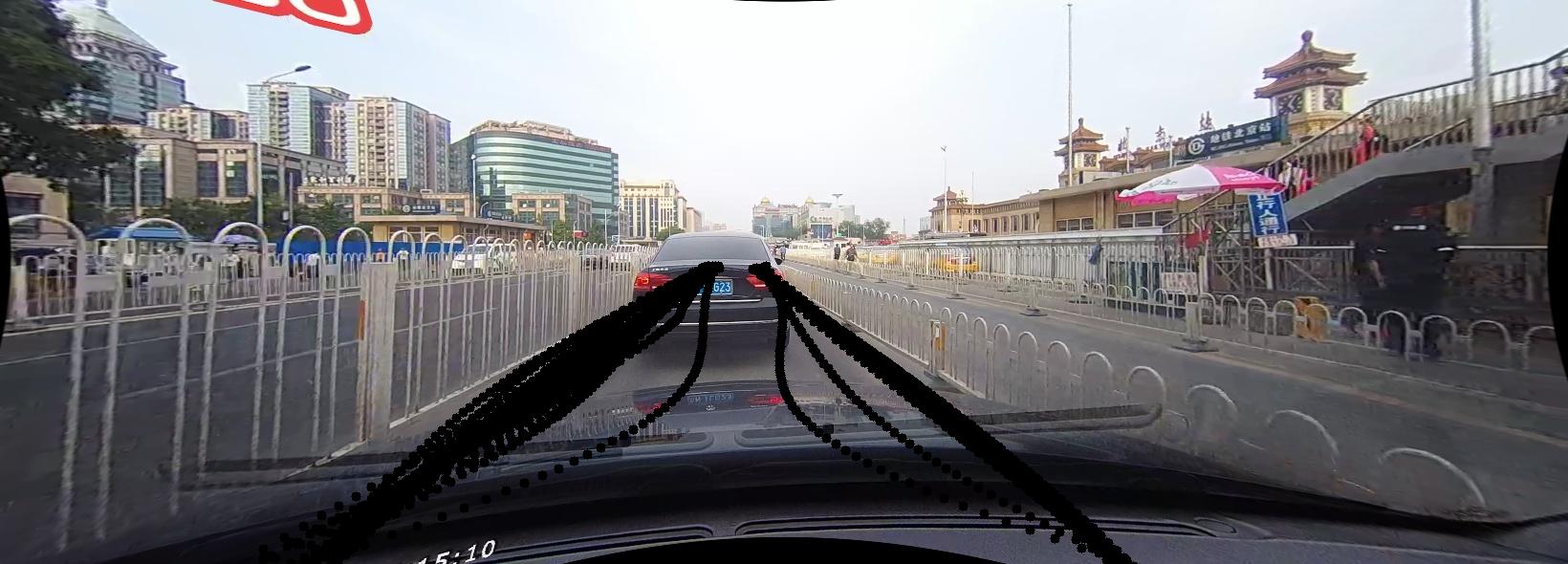}
    }
    \caption{The comparison between the diversity constraint with and without an upper limit.
    The upper limit aims to make the distribution of lines more uniform, as shown in (a).}
    \label{fig:upper-limit}
    \vspace{-4mm}
\end{figure}

To enlarge the shape differences, we propose the diversity constraint to diversify the shapes of different proposal lane lines. Specifically, we first build a symmetric matrix with $K \times K$ elements, where each element represents a pairwise shape difference. A simple way to measure the shape difference is to calculate the curvature difference between different proposals. Therefore, we can raise the shape diversity by enlarging the mean curvature difference:
\begin{equation}
    \begin{aligned}
    J_{div} = -\frac{2}{K(K-1)}\sum_{k=1}^{K}\sum_{i=k}^{K} |S_{k} - S_{i}|,
    \label{equ:ori_div}
    \end{aligned}
\end{equation}
where we only count the elements in the upper triangular matrix, and the diagonal elements are all zero.

However, forcing all proposals to have different shapes may have strong side effects since proposals matching different ground-truth lanes should not be computed together. To solve this issue, we instead compute the intra-cluster shape differences, i.e., we only calculate the shape difference between proposals that match the same ground truth according to Equation~\eqref{eq:rematch}, and improve Equation ~\eqref{equ:ori_div} as follows:
\begin{equation}
    \begin{aligned}
    J_{div} = -\frac{1}{K'}\sum\limits_{g^{(k)}=g^{(i)}} |S_{k} - S_{i}|,
    \end{aligned}
\end{equation}
where $card(\{(k,i)|g^{(k)}=g^{(i)}\})=K'$. To apply the intra-cluster diversity constraint for all groups, we generate a binary mask with $K \times K$ elements, of which only $K'$ elements are 1 and 0 otherwise. 
To illustrate the effect of $J_{div}$, we show the visual comparison between great and small shape differences in Fig.~\ref{fig:difference}. 

While $J_{div}$ achieves a promising overall arrangement of proposals, a few samples whose ground truth lanes are less than $3$ will lead to unstable training. Specifically, few ground truth lanes will cause a large number of proposals to come together. In other words, one ground truth will match multiple proposal lane lines. Therefore, we limit the number of proposed lane lines that match a single ground truth. For example, when there are only two ground truth lane lines in Fig.~\ref{fig:upper-limit} (b), all proposals will be forced to locate near these two lane lines by Equation~\eqref{equ:location_loss}. Setting a higher number will effectively solve this problem without degrading performance. We conduct $J_{loc}$ and $J_{div}$ only on the proposal lane lines in set $\{p_1,p_2,\cdots,p_U | (p_1,p_2,\cdots) = sort(Q_{(i,k)}), g^{(k)}=i\}$. The comparison between the proposals with and without an upper limit is shown in Figure \ref{fig:upper-limit}. It can be concluded that when there is an upper limit, all proposed lane lines can still appear in reasonable locations. To avoid the generalization problem that may be caused by manually setting the upper limit, we set the upper limit value according to the number of proposals.

\begin{table*}[tb]
  \begin{center}
  \setlength{\tabcolsep}{3mm}{
    \caption{Details of Datasets}
    \begin{tabular}{ccccccccc}
      \toprule 
      Dataset & \#Total  &  Train & Validation & Test & Resolution & \#Lines & \#Scenarios & Environment\\ 
      \midrule 
      TuSimple~\cite{tusimple} & 6408 & 3268 & 358 & 2782 &  1280×720 & $\leq5$ & 1 & highway\\
      CULane~\cite{culane} & 133235 & 88880 & 9675 & 34680 & 1640×590 & $\leq4$  & 9 & urban and highway\\
      LLAMAS~\cite{llamas} & 100042 & 58269 & 20844 & 20929 &  1276×717 & $\leq4$  & 1 & highway\\
      CurveLanes~\cite{CurveLane} & 150000 & 100000 & 20000 & 30000 &  2560×1440\&1570×660 & $\leq14$  & 1 & urban and highway\\
      \bottomrule
    \end{tabular}
    \label{tab:datasets}}
  \end{center}
  \vspace{-4mm}
\end{table*}

\begin{figure}[t]
    \centering
    \setlength{\tabcolsep}{0.8pt}
    \begin{tabular}{cc} 
    	\includegraphics[width=1.7in,totalheight=0.8in]{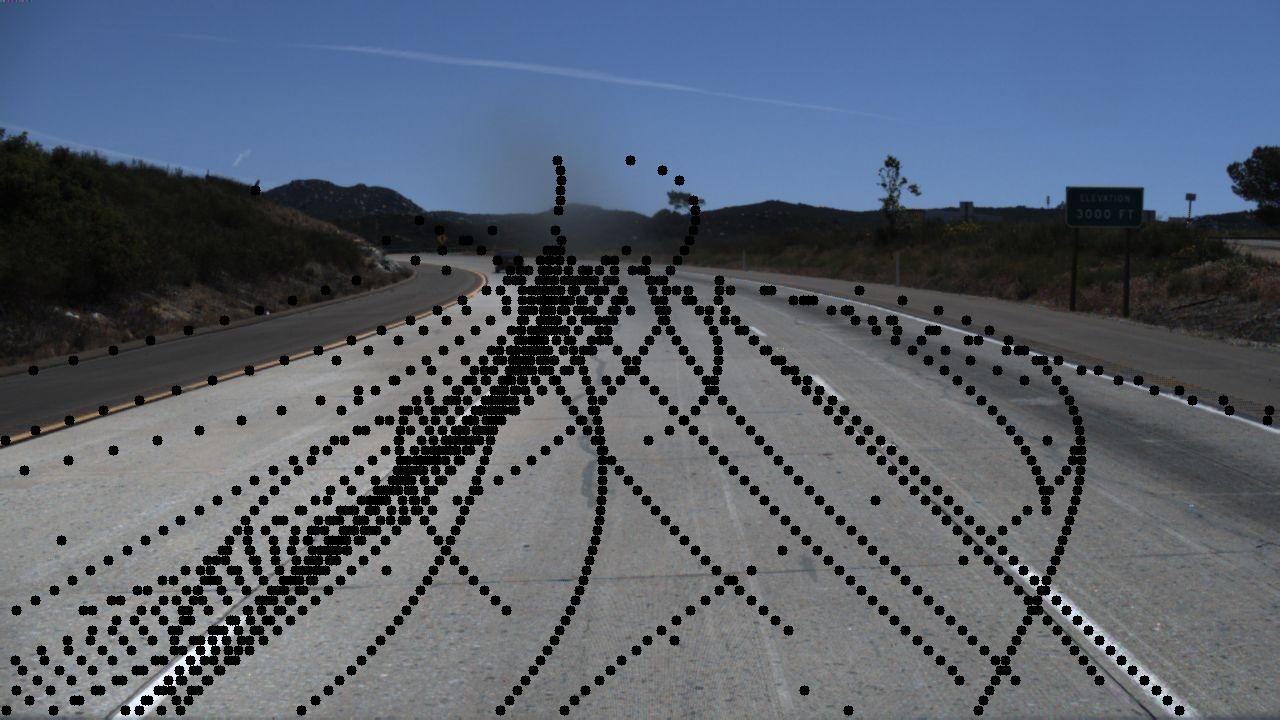}&
    	\includegraphics[width=1.7in,totalheight=0.8in]{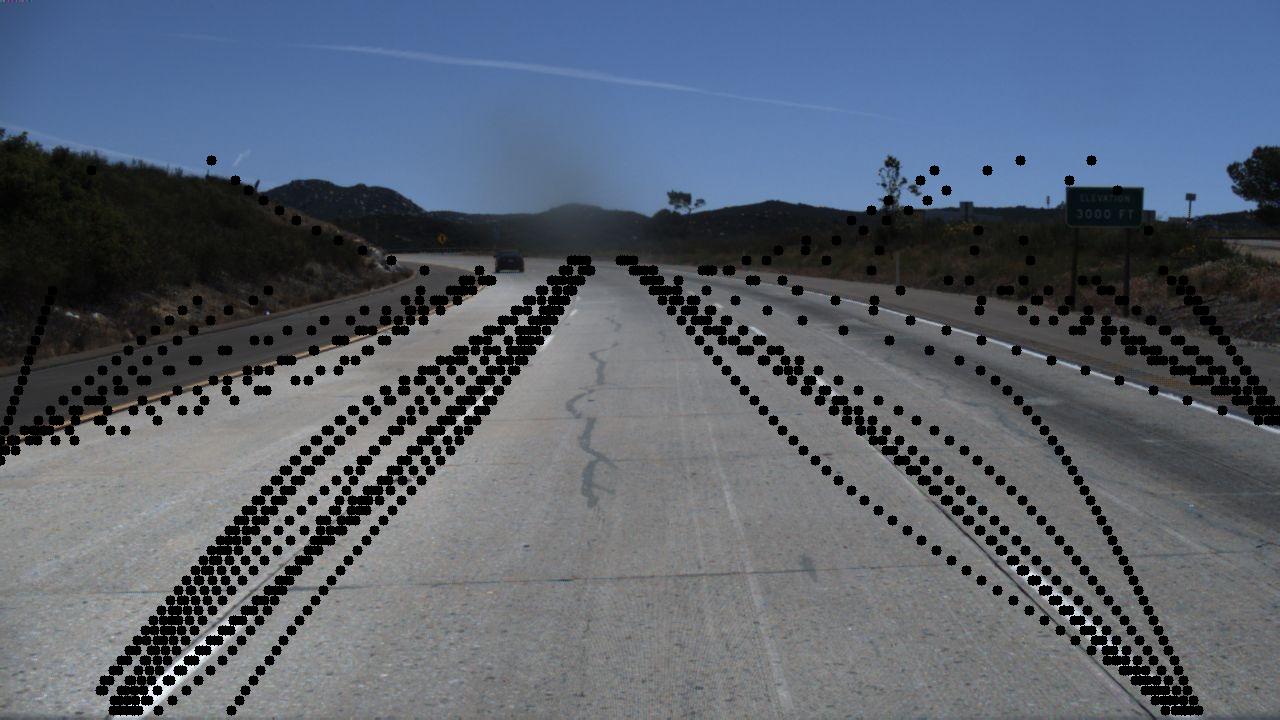}\\
    	\includegraphics[width=1.7in,totalheight=0.8in]{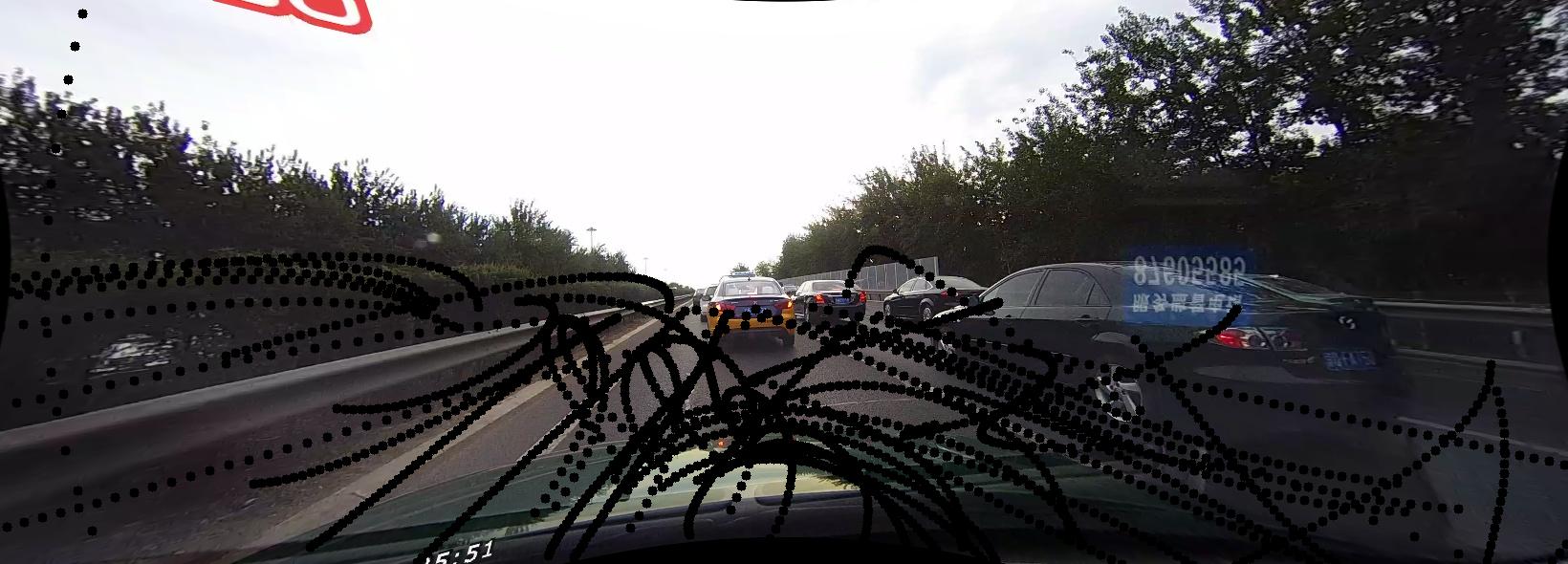}&
    	\includegraphics[width=1.7in,totalheight=0.8in]{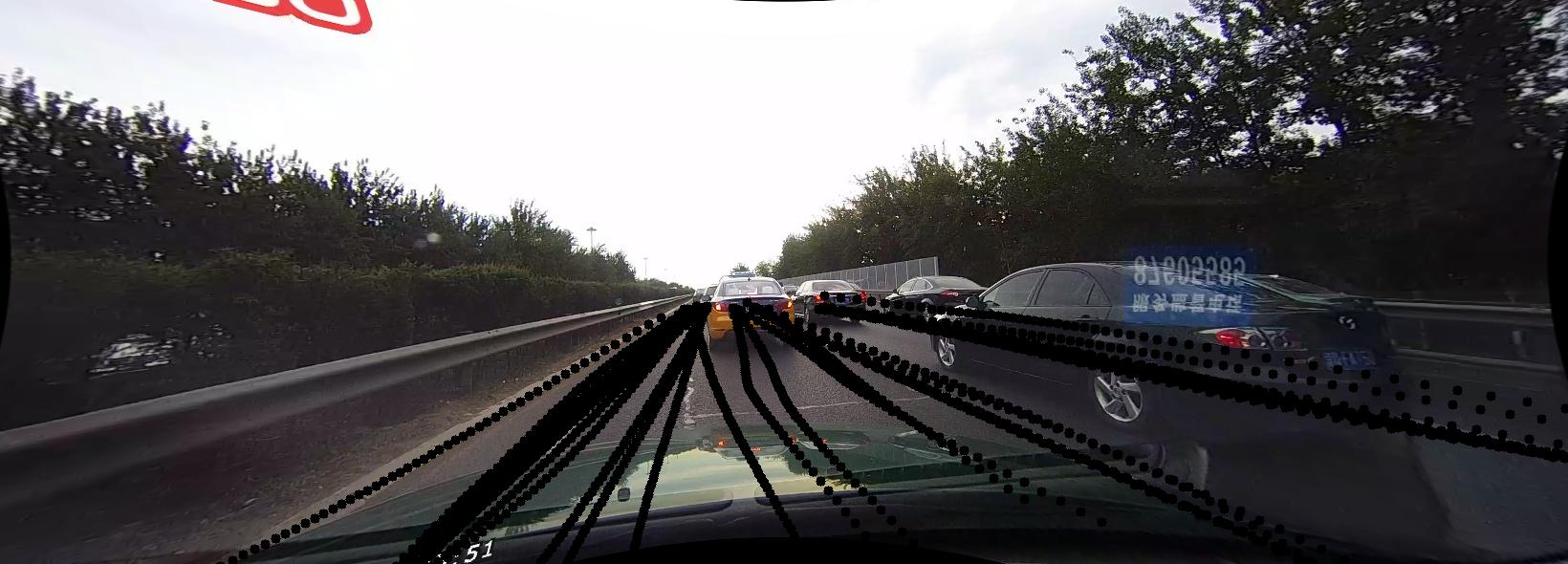}\\
    	\includegraphics[width=1.7in,totalheight=0.8in]{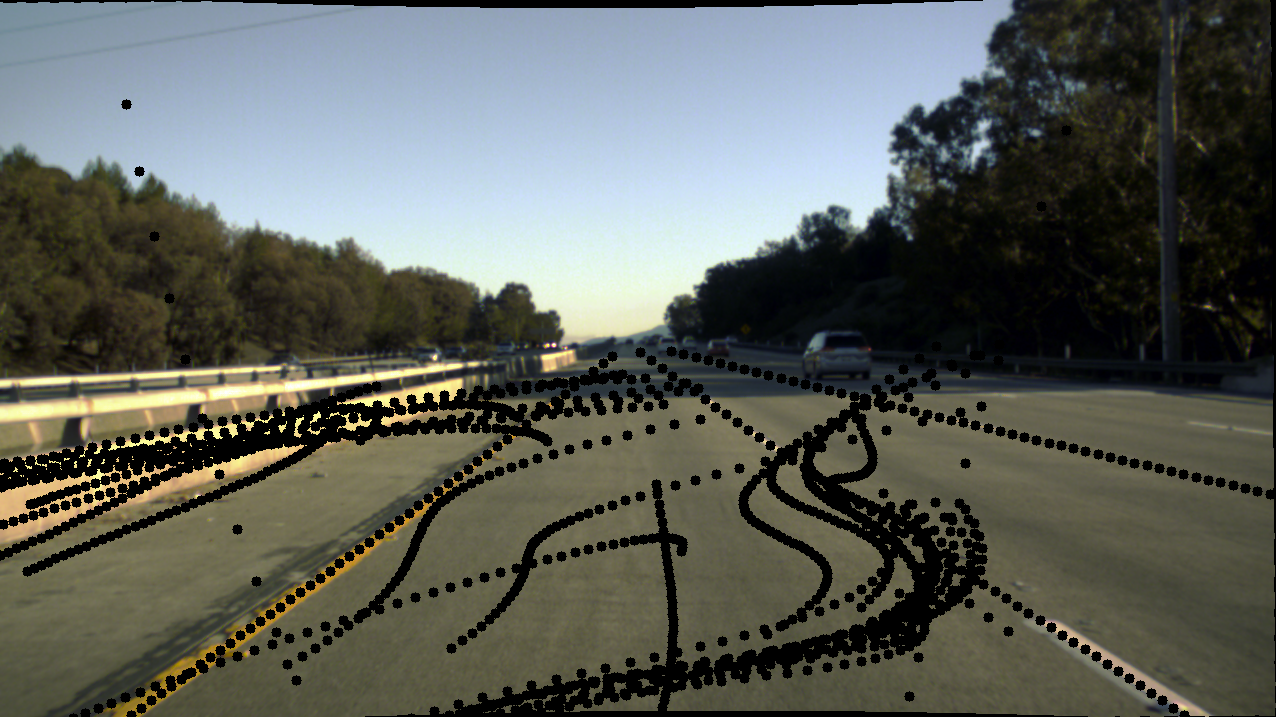}&
    	\includegraphics[width=1.7in,totalheight=0.8in]{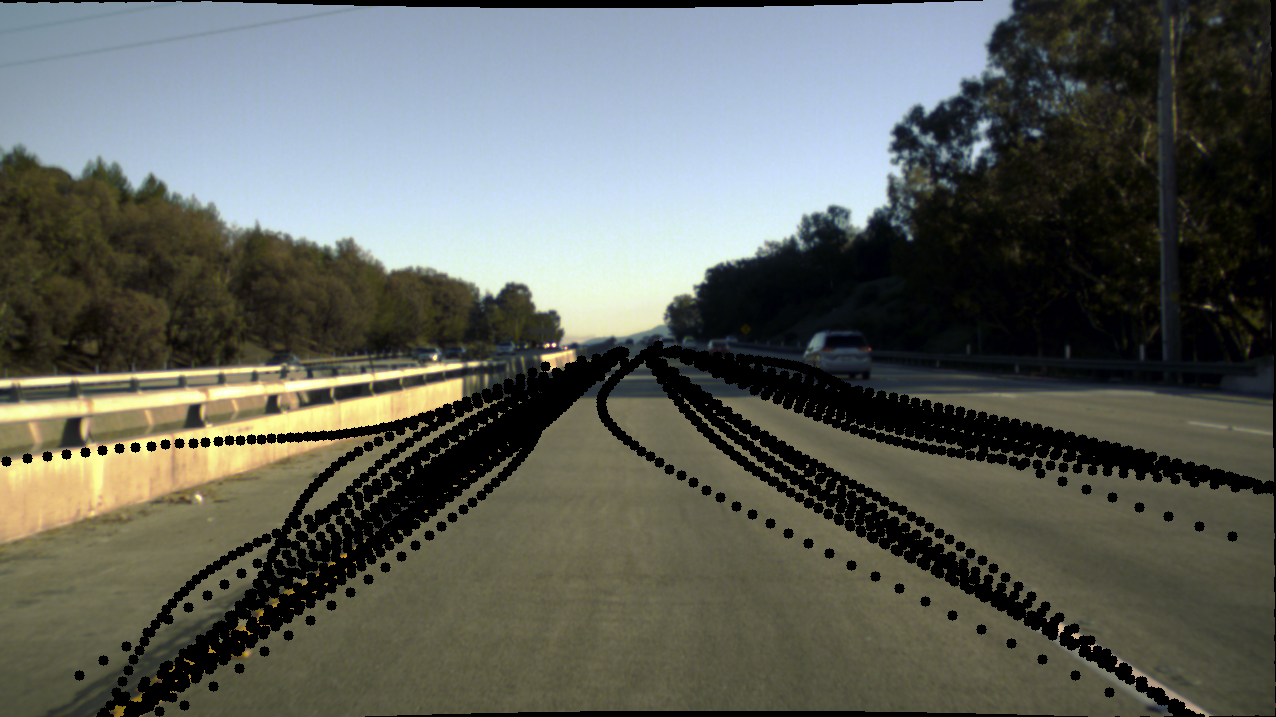}\\
    	BézierLaneNet~\cite{bezier} & ours
    \end{tabular}
    \caption{Proposals comparison between BézierLaneNet~\cite{bezier} and ours.
     From top to bottom, the images are from TuSimple~\cite{tusimple}, CULane~\cite{culane}, and LLAMAS~\cite{llamas} dataset.}
    \label{fig:50show}
    \vspace{-3mm}
\end{figure}

\subsection{Discrimination Constraint}
As mentioned above, existing methods separate confidence loss from proposal quality. Optimal binary matching algorithms such as the Hungarian algorithm ~\cite{wang2021end} only match a relatively close proposal lane line for each ground truth in the image, regardless of the distance between the matched proposal and the ground truth. Matching proposals will have the classification label set to 1, and the rest of the proposals will be assigned 0. This training strategy inevitably produces some high-confidence but low-quality proposals, which eventually become FP. During the testing phase, a proposal with a lower confidence value will be considered a negative prediction, even if it is close to the ground truth. In general, existing methods handle all positive proposals in a fixed manner, i.e., enforcing their confidence level to 1 while ignoring topological and spatial quality.

To address this, we reform the binary cross-entropy loss to boost the discriminative power. During training, different from existing methods that set each matched proposal confidence label to 1, we generate lane-aware ground truths as follows:
\begin{equation}
\begin{aligned}
J_{dis} = -{(l'\log(p) + w(1 - l')\log(1 - p))},
\end{aligned}
\end{equation}
   \begin{eqnarray}\label{equ:softlabel}
      l' =
      \left\{  
                   \begin{array}{lr}  
                    \gamma + (1 - \gamma) \times \mathbf{e}^{-\beta J_{reg}}, ~~~~~~ if~ l=1,   \\  
                    0 , ~~~~~~~~~~~~~~~~~~~~~~~~~~~~~~~ if~ l=0,
                   \end{array}  
      \right. 
\end{eqnarray}
where $l$ means the original existence label, and $l'$ means the new confidence label. $J_{reg}$ is the regression loss of selected proposal, which is set to be a constant variable. $\beta$ and $\gamma$ are used to balance $J_{reg}$, which are set to 10 and 0.5, respectively.
When all proposals are insufficient to match the ground truth, this discriminative constraint will give lower confidence to the closest proposal. As shown in Fig.~\ref{fig:fusion_strategy}(e) and (f), due to the discrimination constraints during training, the three proposals on the right get higher confidence, while the left-most proposal gets lower confidence since their shape and location are not good enough. It proves that our discrimination constraint can effectively help screen out close but unqualified proposals, thereby effectively reducing FP. More importantly, we can determine a threshold in the testing phase to distinguish between true and false predictions with less human effort.
Experimental results in Section~\ref{sec:experiment} show that $J_{dis}$ successfully improved the performance from the perspective of the confidence distribution of proposals.

\subsection{Overall Loss}
Different from the general curve-based models~\cite{bezier,LSTR}, we add the availability constraint, diversity constraint, and discrimination constraint together to modulate the quality of proposals. Our overall loss is a weighted sum of five losses:
{
\begin{align}
    J_{dense} = \lambda_{1}J_{reg} + \lambda_{2}J_{seg} + \lambda_{3}J_{ava} + \lambda_{4}J_{div} + \lambda_{5}J_{dis},
\label{eq:lossdense}
\end{align}}
where $\lambda_1, \lambda_2, \lambda_3, \lambda_4, \lambda_5$ are set to $1, 0.1, 0.0005, 0.0001, 0.75$, respectively. These coefficients are determined by performing 5-fold cross-validation. Compared with the sparse constraints in existing methods, our dense hybrid modulation significantly improves lane proposals' topological and spatial qualities without introducing new parameters. We compare proposal quality with our baseline BézierLaneNet in Fig.~\ref{fig:50show}, and the comparison shows that our proposals outperform the baselines on all datasets.

\begin{table*}[t]
    \begin{center}
    \caption{Quantitative Results on the CULane~\cite{culane} Dataset}
    \renewcommand{\arraystretch}{1.3}
    \begin{threeparttable}
        \begin{tabular}{lccccccccccc} 
            \toprule
            \textbf{Method}  & \textbf{Total} & \textbf{Normal} & \textbf{Crowded} & \textbf{Dazzle} & \textbf{Shadow} & \textbf{No line} & \textbf{Arrow} & \textbf{Curve}  & \textbf{Crossroad \ddag} $\downarrow$ & \textbf{Night} \\ 
            \cline{1-11}
            \textbf{Segmentation-based Method} &        &        &         &        &        &        &        &        &       &       \\
            \cline{1-1}
            SCNN (LargeFOV)~\cite{culane}                 & 71.60  & 90.60  & 69.70   & 58.50  & 66.90  & 43.40  & 84.10  & 64.40  & 1990  & 66.10 \\
            SAD (ENet)~\cite{ENet-SAD}                & 70.80  & 90.10  & 68.80   & 60.20  & 65.90  & 41.60  & 84.00  & 65.70  & 1998  & 66.00 \\
            SIM-CycleGAN (ERFNet)~\cite{SIM}            & 73.90  & 91.80  & 71.80   & 66.40  & 76.20  & 46.10  & 87.80  & 67.10  & 2346  & 69.40 \\
            RESA (ResNet-34)~\cite{RESA}       & 74.50  & 91.90  & 72.40   & 66.50  & 72.00  & 46.30  & 88.10  & 68.60  & 1896  & 69.80 \\
            \cline{1-11}
            \textbf{Anchor-based Methods}      &        &        &         &        &        &        &        &        &       &       \\
            \cline{1-1}
            UFLD (ResNet-18) ~\cite{UFAST}      & 68.40  & 87.70  & 66.00   & 58.40  & 62.80  & 40.20  & 81.00  & 57.90  & 1743  & 62.10 \\
            UFLD (ResNet-34) ~\cite{UFAST}      & 72.30  & 90.70  & 70.20   & 59.50  & 69.30  & 44.40  & 85.70  & 69.50  & 2037  & 66.70 \\
            CurveLanes-NAS(ResNet-18)~\cite{CurveLane}  & 71.40  & 88.30  & 68.60   & 63.20  & 68.00  & 47.90  & 82.50  & 66.00  & 2817  & 66.20 \\
            CurveLanes-NAS (ResNet-34)~\cite{CurveLane}  & 73.50  & 90.20  & 70.50   & 65.90  & 69.30  & 48.80  & 85.70  & 67.50  & 2359  & 68.20 \\
            UFLDv2 (ResNet-18)~\cite{UFLDv2}    & 74.70  & 91.70  & 73.00   & 64.60  & 74.70  & 47.20  & 87.60  & 68.70  & 1998  & 70.20 \\
            UFLDv2 (ResNet-34)~\cite{UFLDv2}    & 75.90  & 92.50  & 74.90   & 65.70  & 75.30  & 49.00  & 88.50  & 70.20  & 1864  & 70.60 \\
            CondLaneNet (ResNet-18)~\cite{CondLaneNet}      & 78.14  & 92.87  & 75.79   & 70.72  & 80.01  & 52.39  & 89.37  & 72.40  & 1364  & 73.23 \\
            CLRNet (ResNet-18)~\cite{CLRNet}                & 79.58  & 93.30  & 78.33   & 73.71  & 79.66  & 53.14  & 90.25  & 71.56  & 1321  & 75.11 \\
            LaneATT (ResNet-18)~\cite{LaneATT}             & 75.13  & 91.17  & 72.71   & 65.82  & 68.03  & 49.13  & 87.82  & 63.75  & 1020  & 68.58 \\
            \textbf{ours-LaneATT (ResNet-18)}               & 75.42  & 91.12  & 72.88   & 66.49  & 70.61  & 48.74  & 86.84  & 65.29  & 979   & 69.73  \\
            \cline{1-11}
            \textbf{Curve-based Methods}            &        &        &        &        &        &         &        &        &      &      \\
            \cline{1-1}
            LSTR(ResNet-18)*~\cite{LSTR}            & 68.72  & 86.78  & 67.34  & 56.63  & 59.82  & 40.10   & 78.66  & 56.64  & 1166 & 59.92 \\
            BézierLaneNet (ResNet-18)~\cite{bezier} & 73.67  & 90.22  & 71.55  & 62.49  & 70.91  & 45.30   & 84.09  & 58.98  & 996  & 68.70 \\
            BézierLaneNet (ResNet-34)~\cite{bezier} & 75.57  & 91.59  & 73.20  & 69.20  & 76.74  & 48.05   & 87.16  & 62.45  & 888  & 69.90 \\ 
            \textbf{ours-BézierLaneNet (ResNet-18)}               & 74.59  & 90.55  & 73.25  & 63.49  & 65.53  & 45.11   & 84.83  & 60.19  & 701  & 68.94  \\
            \textbf{ours-BézierLaneNet (ResNet-34)}               & 76.21  & 91.50  & 74.19  & 68.11  & 74.85  & 48.25   & 87.38  & 60.41  & 872  & 71.51  \\
            \bottomrule
        \end{tabular}
    \label{tab:culane-compare}
    \begin{tablenotes}
    \item[] *Since the result of CULane~\cite{culane} is not given in the LSTR~\cite{LSTR} paper, we used the results in the paper of BézierLaneNet~\cite{bezier} here.
    \item[] \ddag For the “Cross” category, only false positives are shown.
    \end{tablenotes}
    \end{threeparttable}
    \end{center}
    \vspace{-5mm}
\end{table*}

\section{Experiments}
\label{sec:experiment}
In this section, we conduct various experiments to evaluate the proposed dense proposal modulation method. We perform an ablation study to show the effectiveness of each design through qualitative and quantitative analysis.
\subsection{Experimental Setting}
\subsubsection{\textbf{Datasets}}
To evaluate our methods, we conduct experiments on four commonly used benchmark datasets, including TuSimple~\cite{tusimple}, CULane~\cite{culane}, LLAMAS~\cite{llamas} and CurveLanes~\cite{CurveLane} datasets. TuSimple~\cite{tusimple} is collected on the highway in the daytime with fair weather conditions. CULane~\cite{culane} is collected by cameras installed on six different vehicles in Beijing, with 9 different scenarios such as night, crowded, and dazzle. LLAMAS~\cite{llamas} is a newly formed large dataset with over 100k annotated images, which is collected using Lidar maps. {CurveLanes is a recently released dataset whose driving scenarios are full of complex topologies such as curve lines and dense lines. Detailed information about the datasets can be found in Table \ref{tab:datasets}.}

\begin{figure*}[h]
    \centering
    \setlength{\tabcolsep}{0.8pt}
    \begin{tabular}{cccc} 
        \rotatebox{90}{\scriptsize{~~~~~~~\textbf{Night}}}
    	\includegraphics[width=1.7in,totalheight=0.8in]{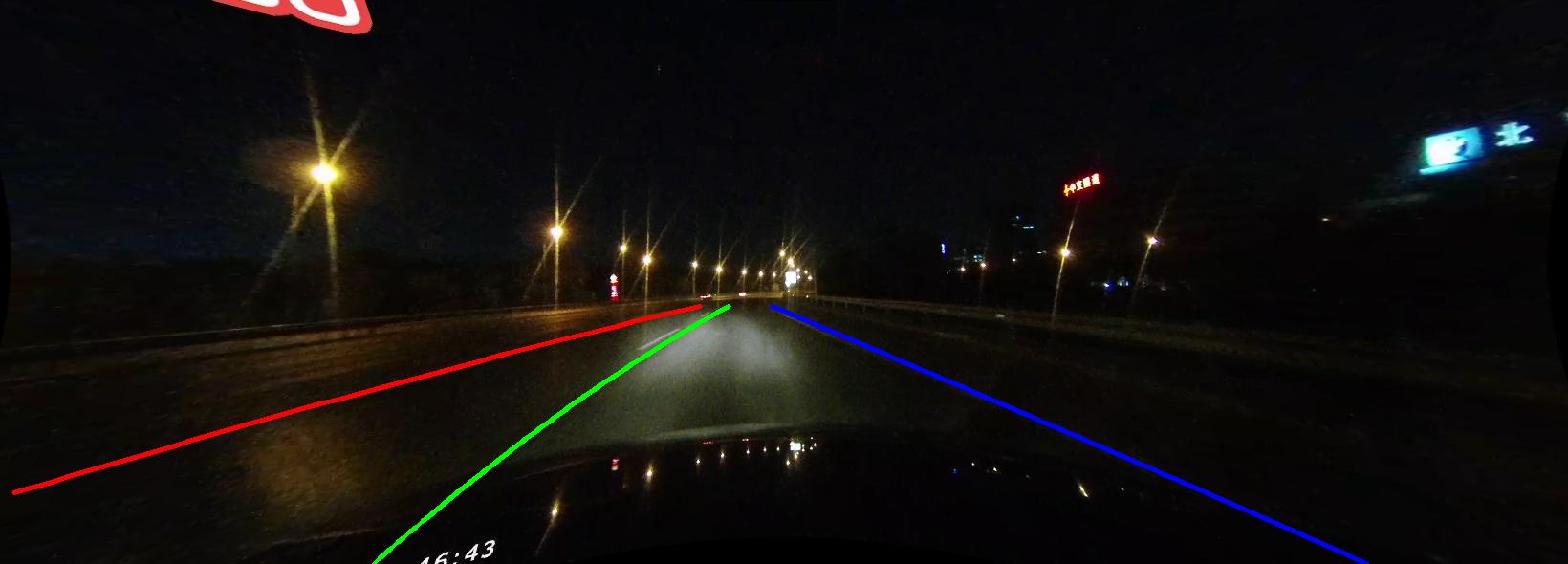}&
    	\includegraphics[width=1.7in,totalheight=0.8in]{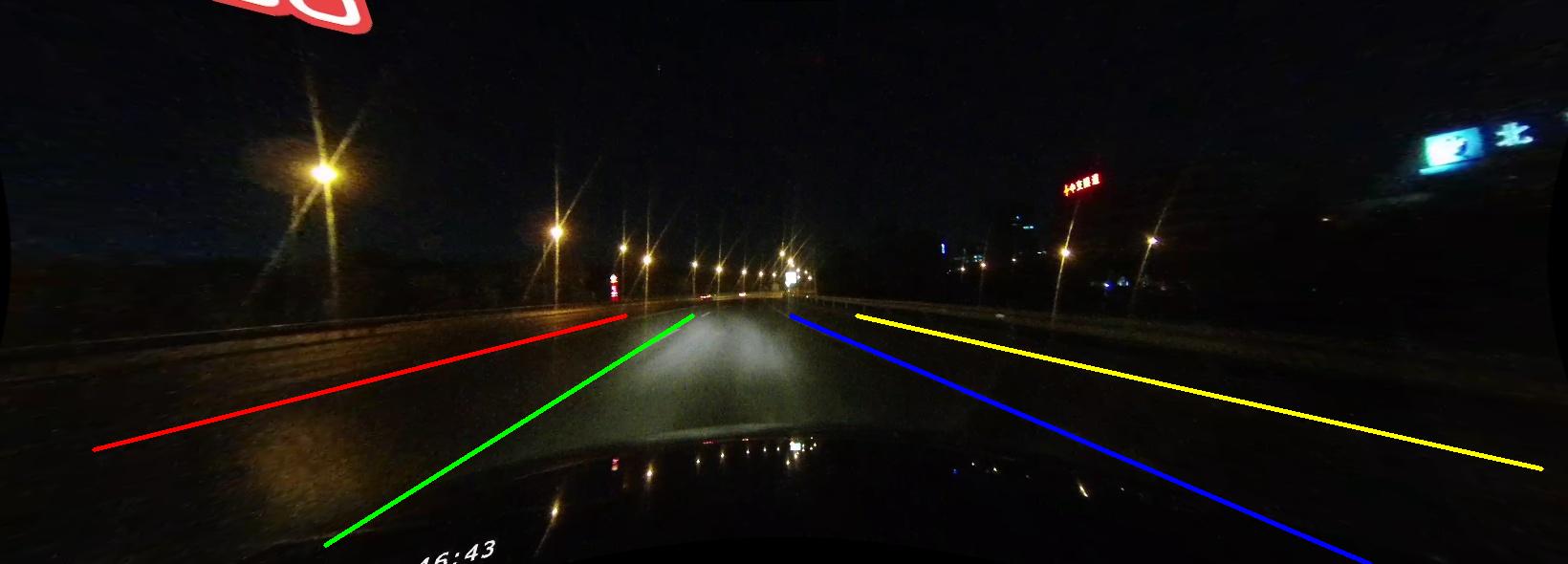}&
    	\includegraphics[width=1.7in,totalheight=0.8in]{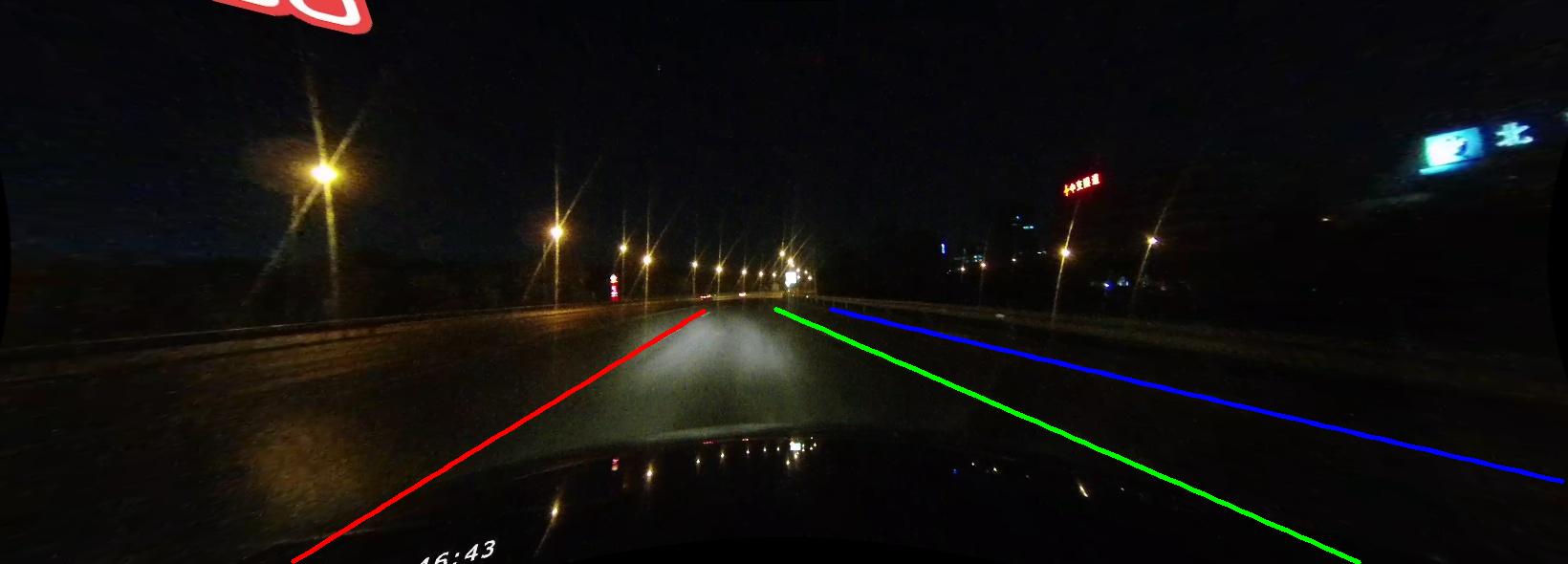}&
    	\includegraphics[width=1.7in,totalheight=0.8in]{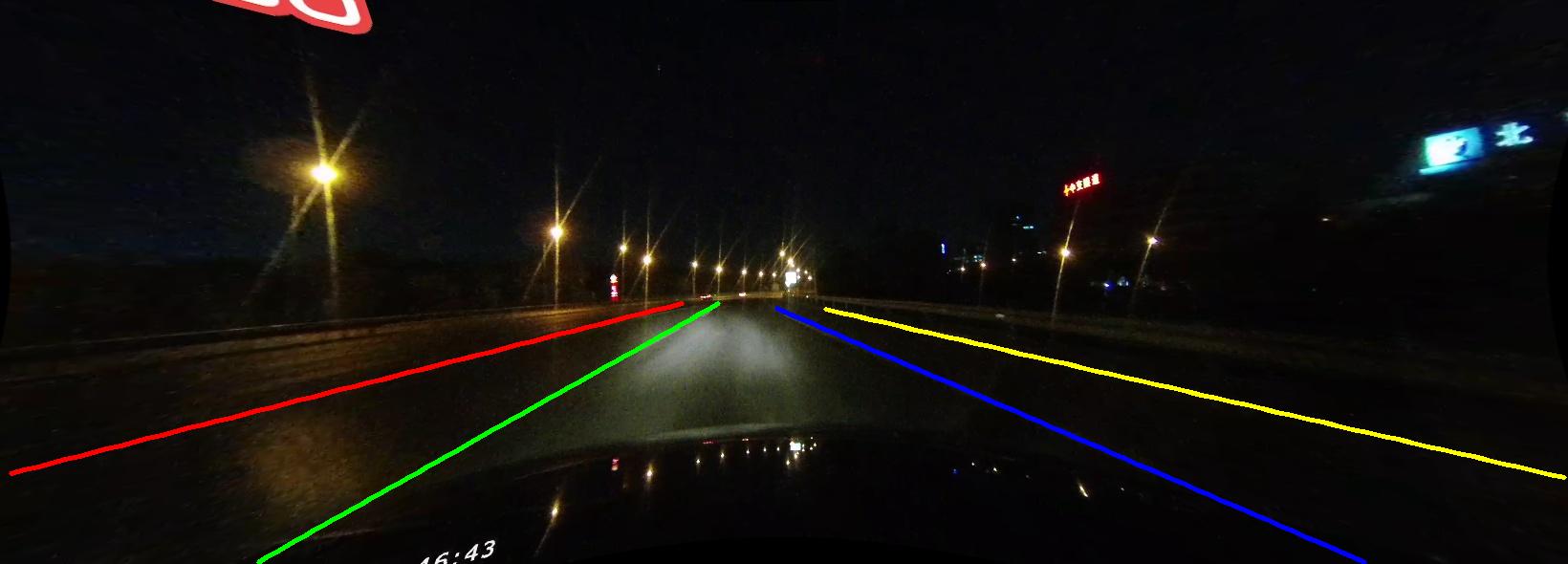}\\
    	\rotatebox{90}{\scriptsize{~~~~~~~\textbf{Shadow}}}
    	\includegraphics[width=1.7in,totalheight=0.8in]{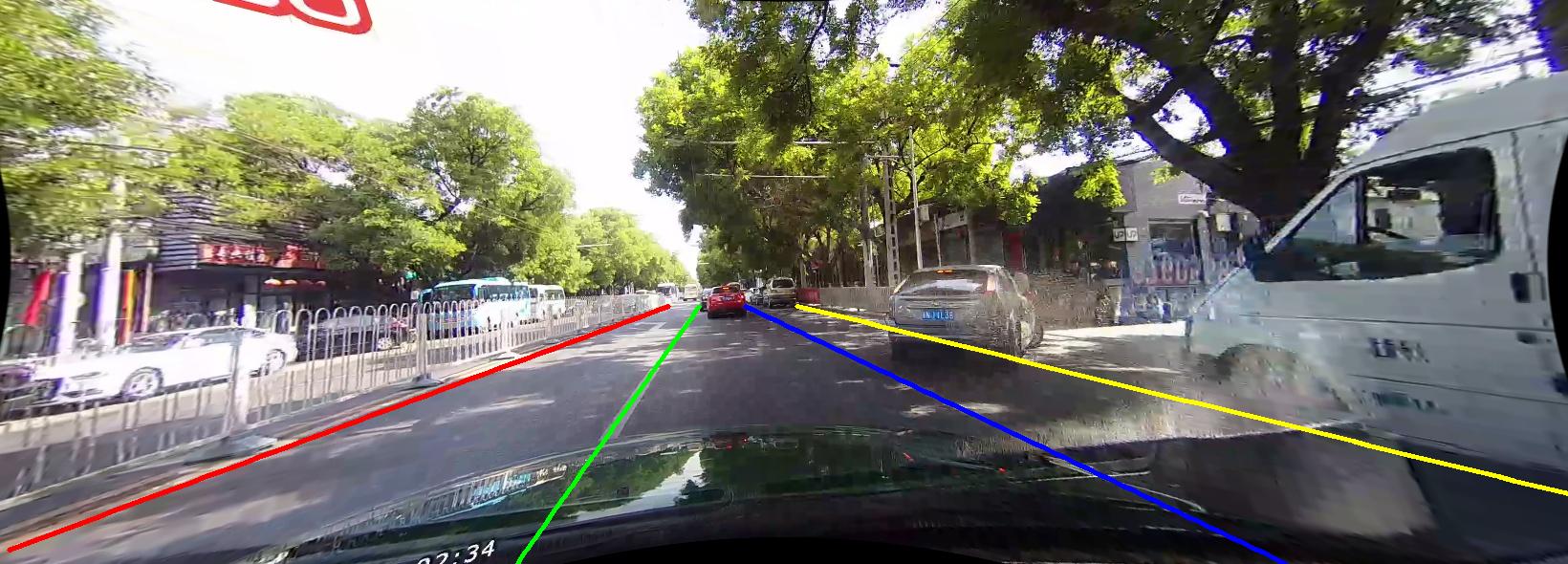}&
    	\includegraphics[width=1.7in,totalheight=0.8in]{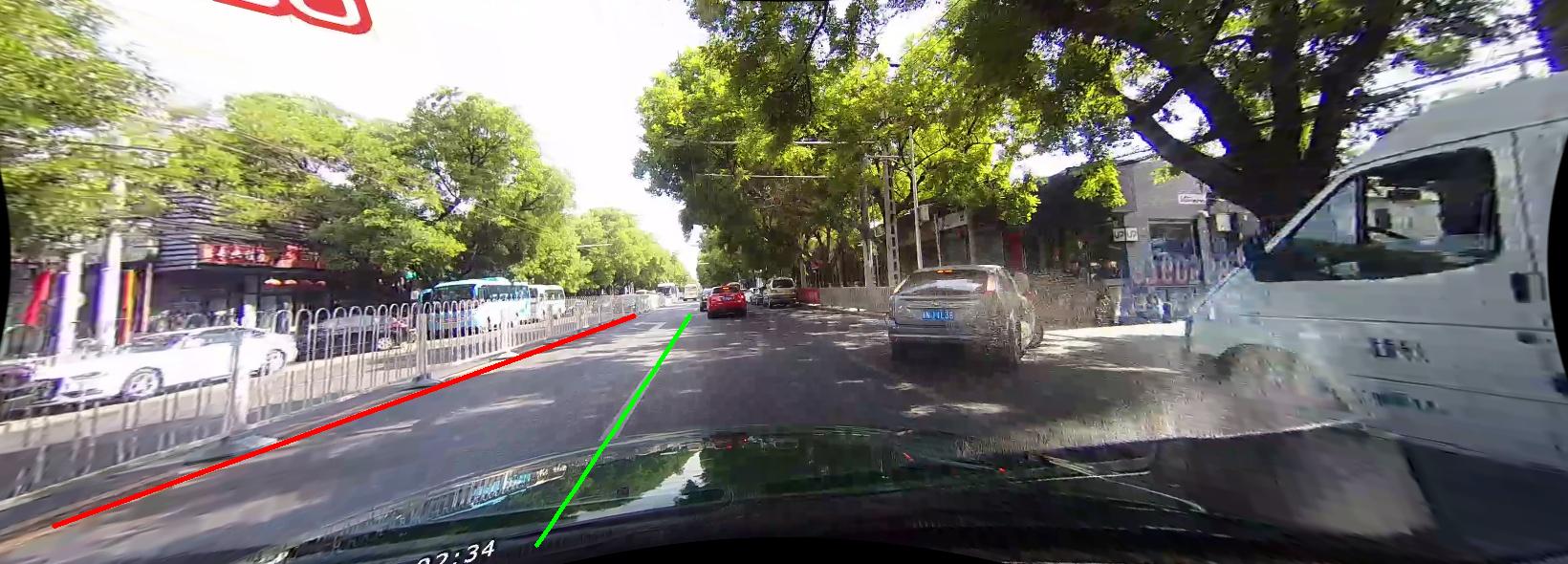}&
    	\includegraphics[width=1.7in,totalheight=0.8in]{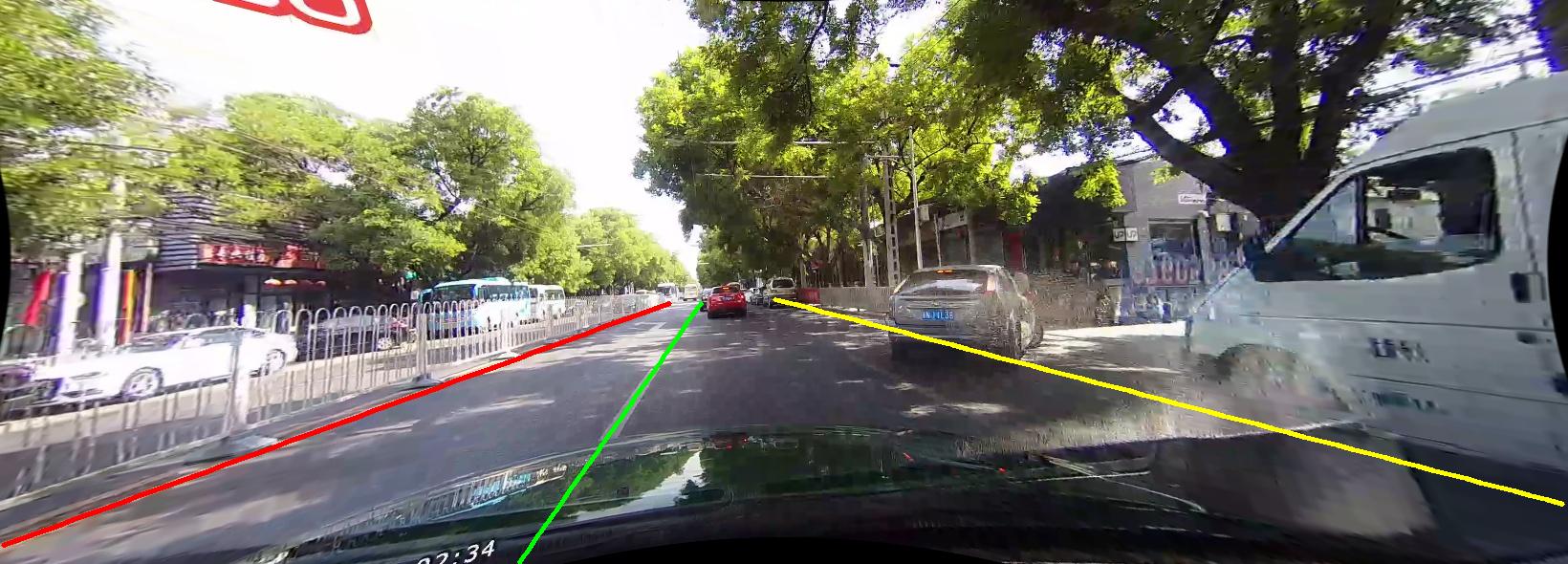}&
    	\includegraphics[width=1.7in,totalheight=0.8in]{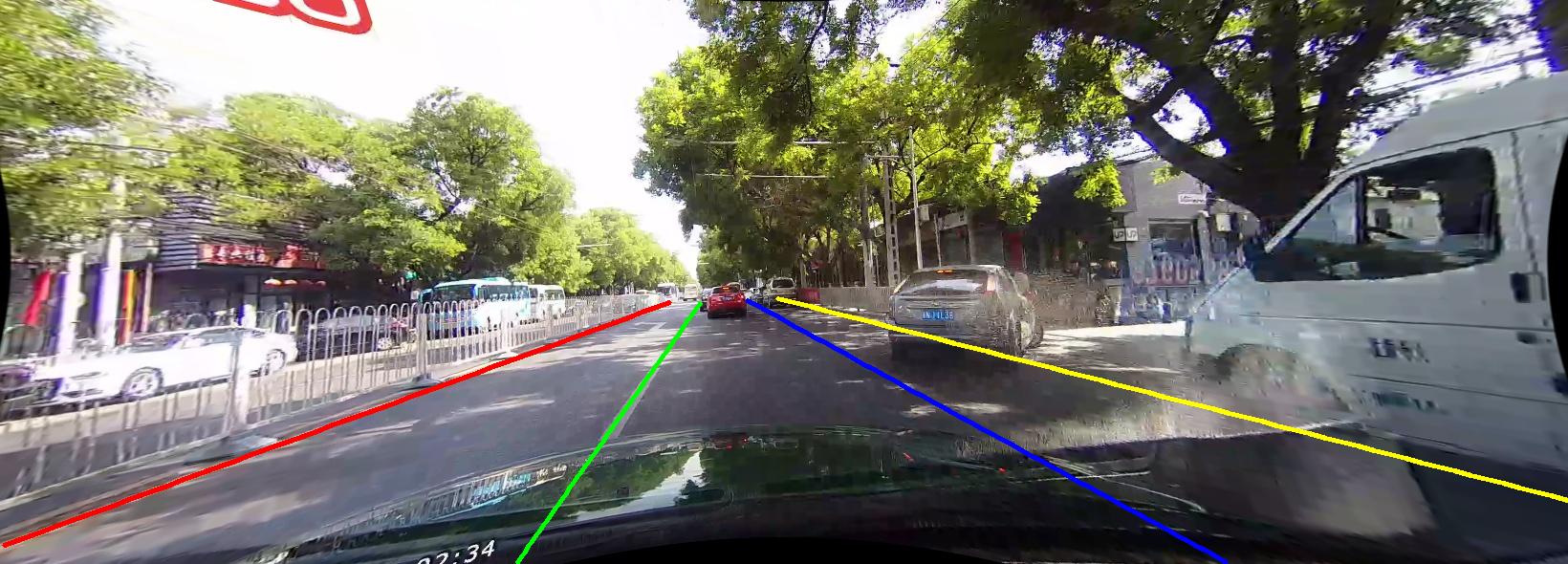}\\
    	\rotatebox{90}{\scriptsize{~~~~~~~\textbf{Crowded}}}
    	\includegraphics[width=1.7in,totalheight=0.8in]{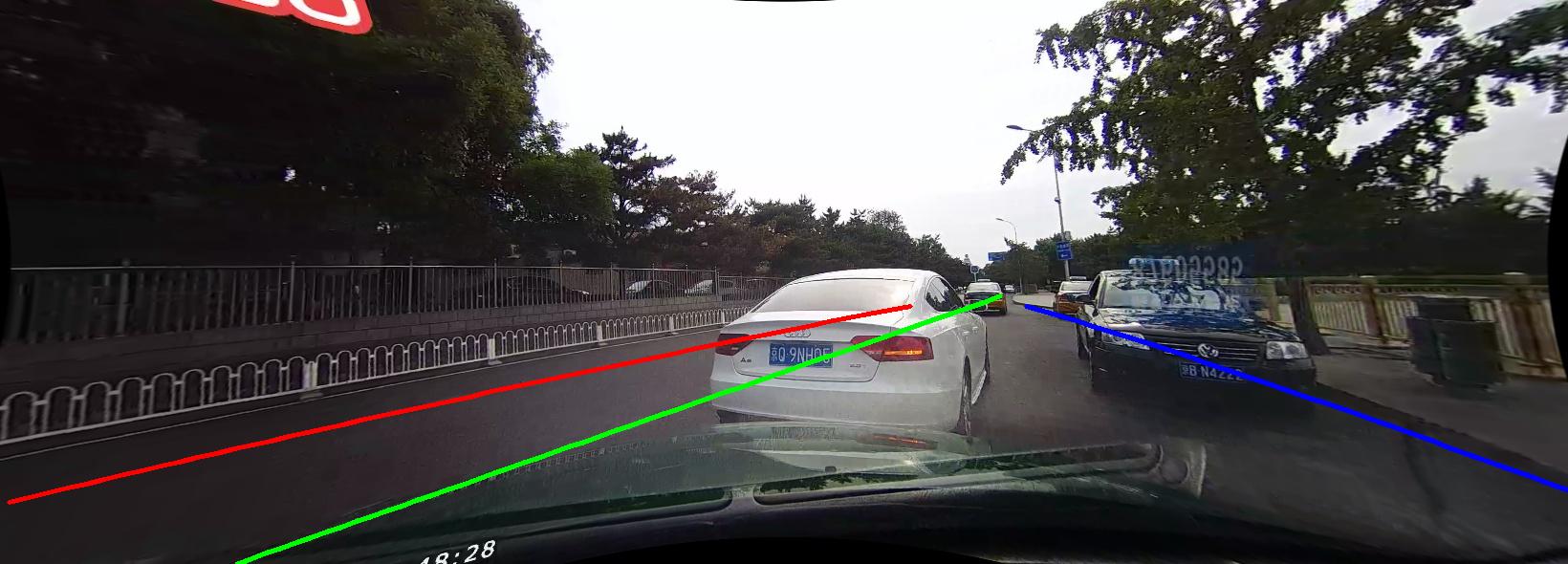}&
    	\includegraphics[width=1.7in,totalheight=0.8in]{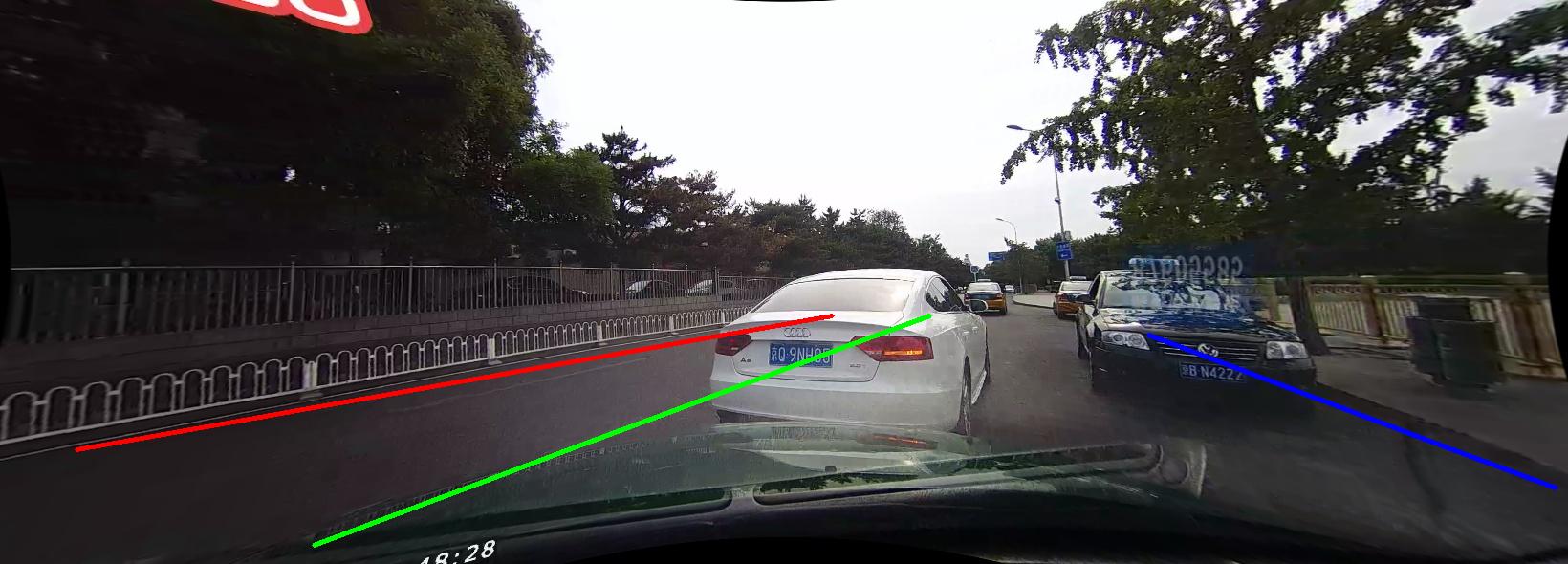}&
    	\includegraphics[width=1.7in,totalheight=0.8in]{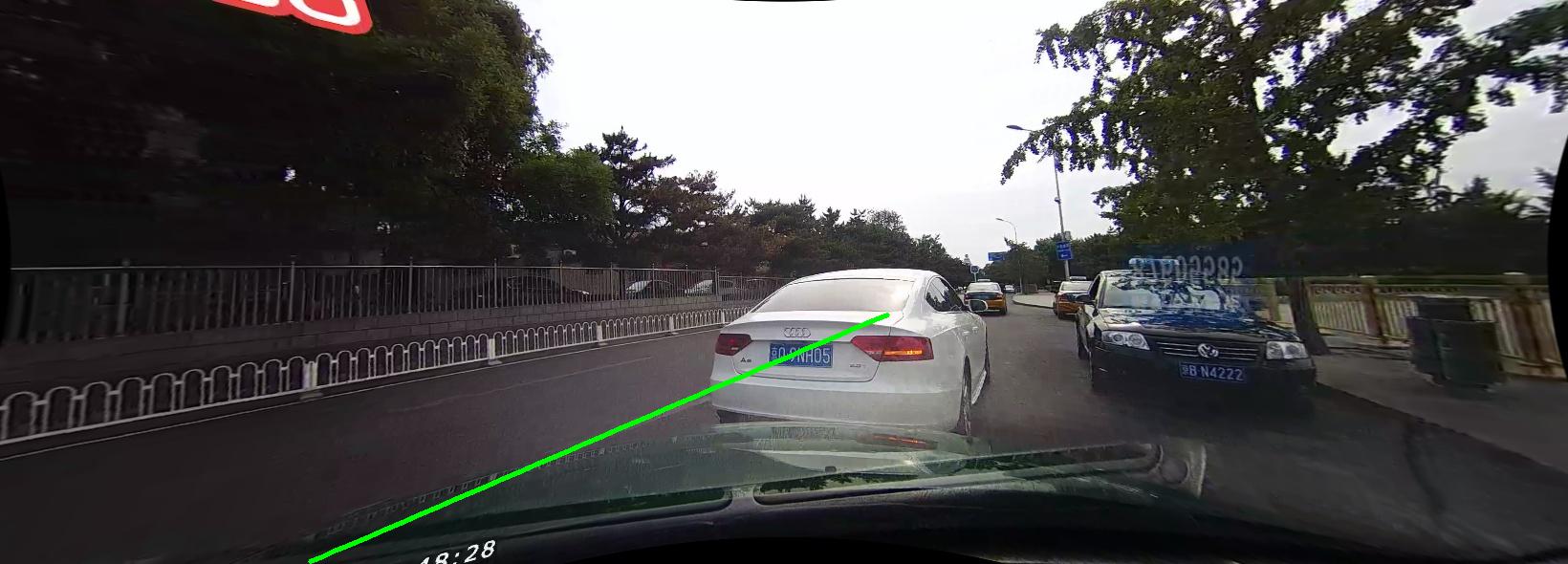}&
    	\includegraphics[width=1.7in,totalheight=0.8in]{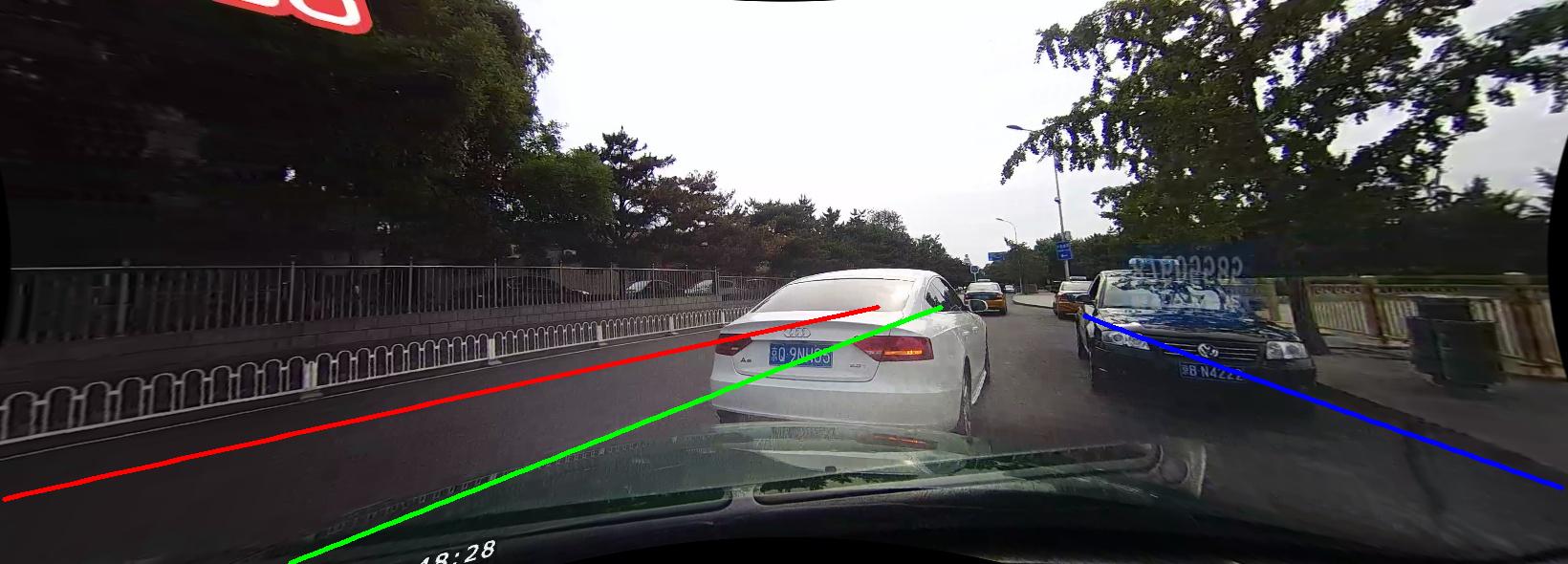}\\
    	\rotatebox{90}{\scriptsize{~~~~~~~\textbf{Normal}}}
    	\includegraphics[width=1.7in,totalheight=0.8in]{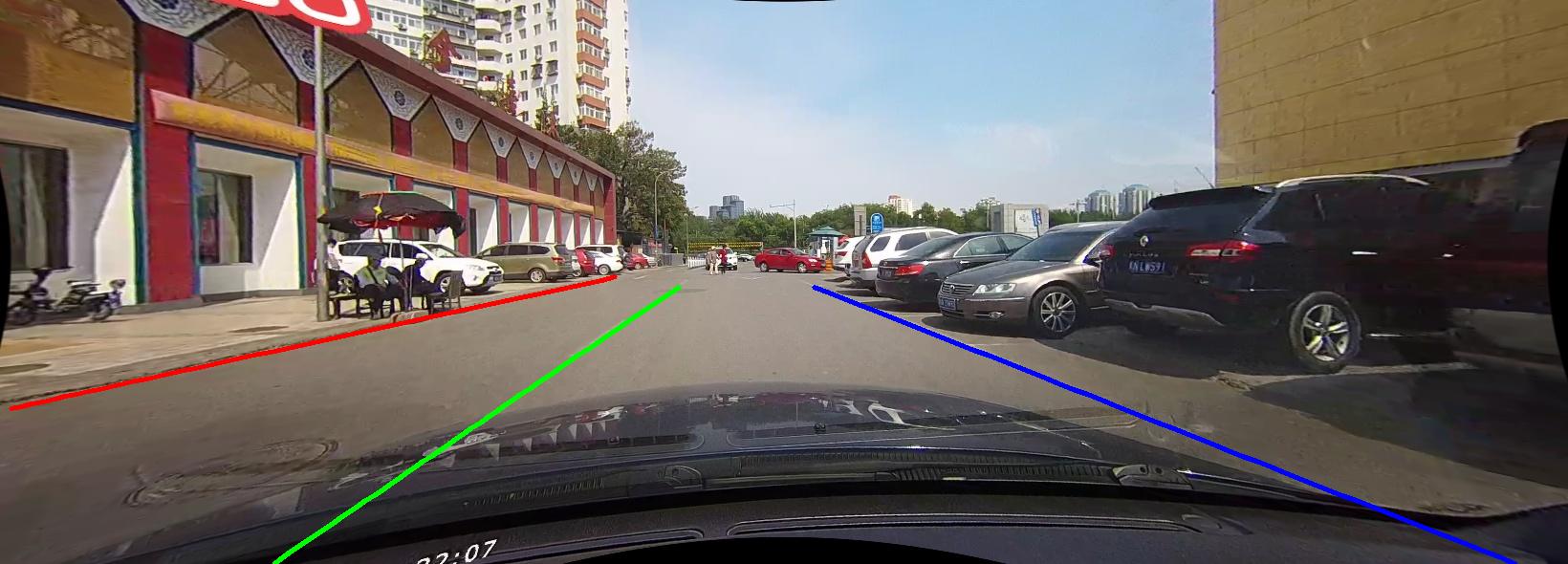}&
    	\includegraphics[width=1.7in,totalheight=0.8in]{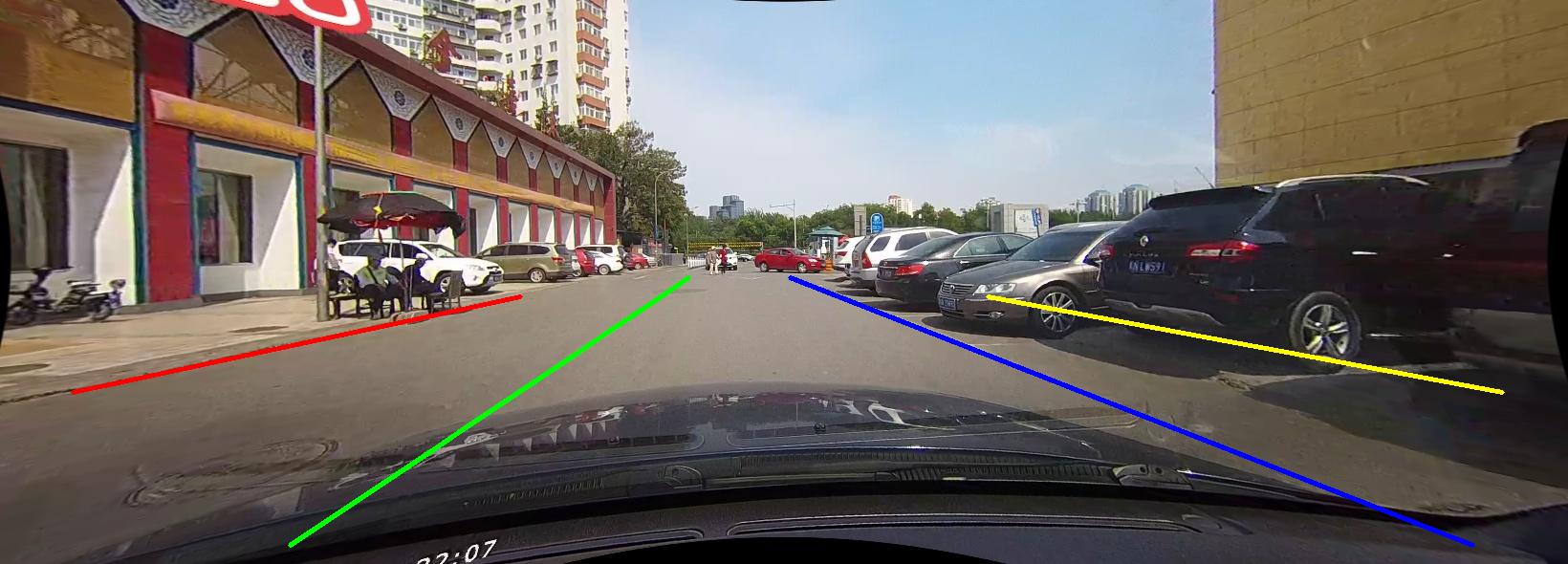}&
    	\includegraphics[width=1.7in,totalheight=0.8in]{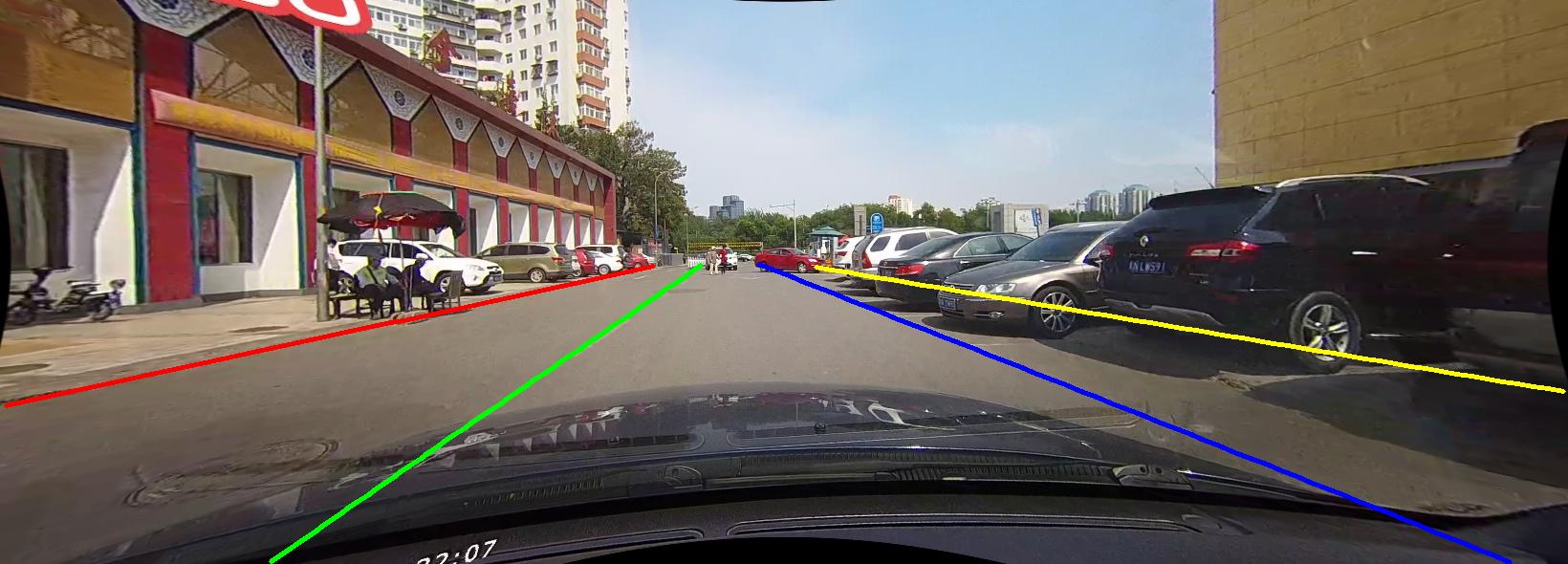}&
    	\includegraphics[width=1.7in,totalheight=0.8in]{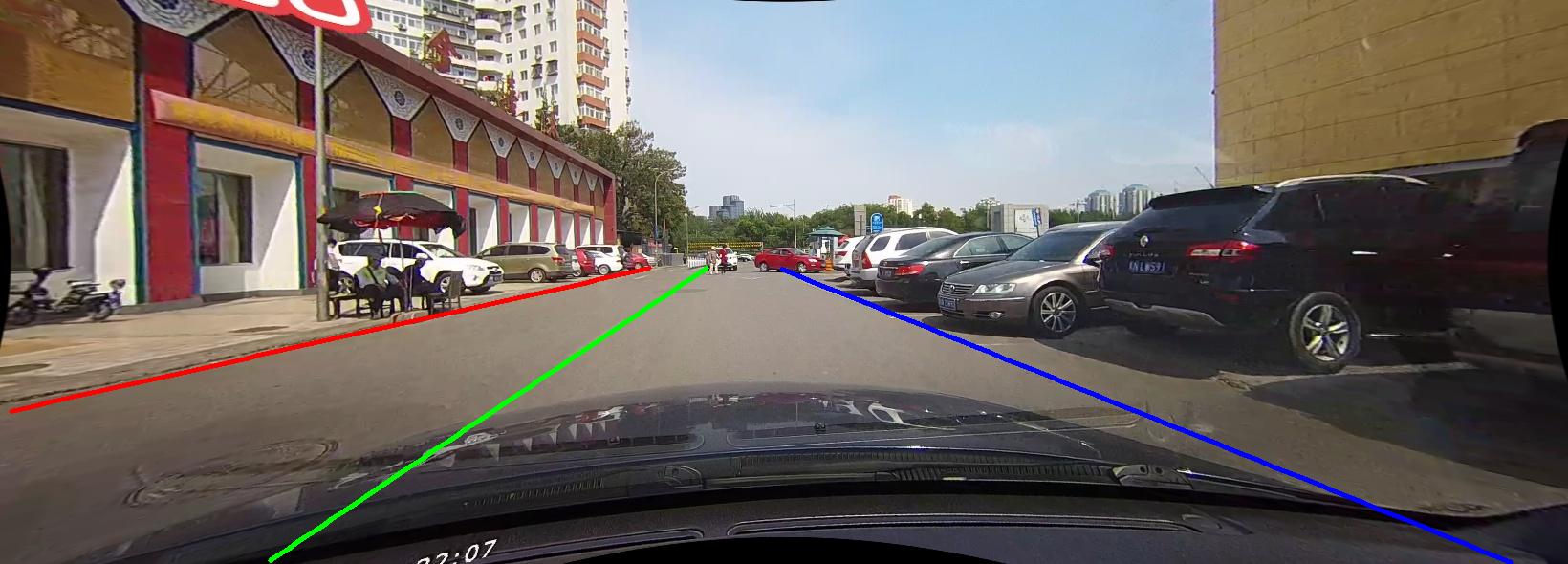}\\
    	{Ground truth} &{LSTR~\cite{LSTR}} &{BézierLaneNet~\cite{bezier}} &{Ours}\\
    \end{tabular}
    \caption{Qualitative results on the CULane~\cite{culane} dataset.}
    \label{fig:culane_resultshow}
    \vspace{-3mm}
\end{figure*}

\subsubsection{\textbf{Evaluation Metrics}}
For the TuSimple~\cite{tusimple} dataset, there are three official evaluation metrics: accuracy, false positive rate (FPR), and false negative rate (FNR). The accuracy is computed by:
\begin{equation}
    \begin{aligned}
    accuracy = \frac{\sum_{clip}C_{clip}}{\sum_{clip}S_{clip}},
    \end{aligned}
\end{equation}
where $C_{clip}$ is the number of correctly predicted lane line points, and $S_{clip}$ is the total number of ground truth points in each clip. Different from the TuSimple~\cite{tusimple} dataset, the official evaluation metric for CULane~\cite{culane}, LLAMAS~\cite{llamas} and CurveLanes~\cite{CurveLane} datasets is F1 score, which is calculated as follows:
\begin{equation}
    \begin{aligned}
    F1 = \frac{2 \times Precision \times Recall}{Precision+Recall},
    \end{aligned}
\end{equation}
\begin{equation}
    \begin{aligned}
    Precision = \frac{TP}{TP+FP},
    Recall = \frac{TP}{TP+FN}.
    \end{aligned}
\end{equation}
The prediction is considered true positive ($TP$) when the intersection-over-union (IoU) between the prediction and ground truth exceeds $0.5$.

\begin{figure}[t]
    \centering
    \subfigure[Statistics of times each proposal is selected in BézierLaneNet~\cite{bezier}]{
    	\includegraphics[width=3.2in]{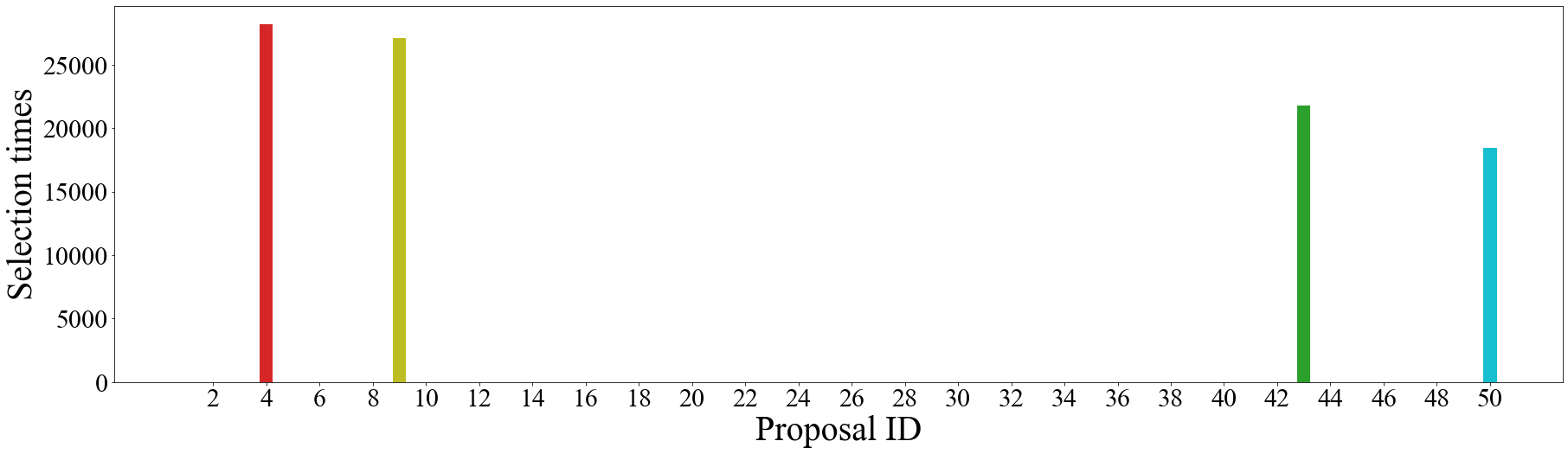}
    }
    \subfigure[Statistics of times each proposal is selected in our method]{
    	\includegraphics[width=3.2in]{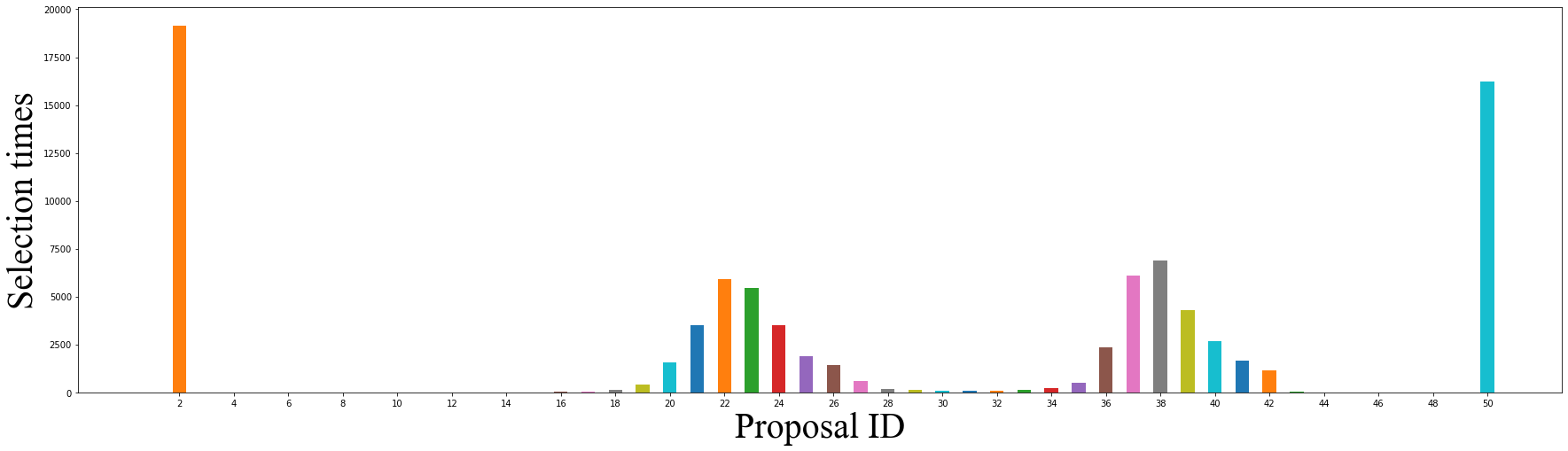}
    }
    \subfigure[Selected proposal statistics of BézierLaneNet~\cite{bezier}]{
    	\includegraphics[width=1.3in]{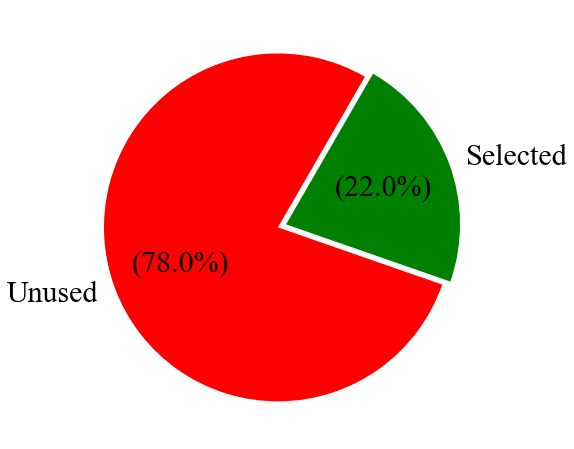}
    }
    \subfigure[Selected proposal statistics of our methods]{
    	\includegraphics[width=1.3in]{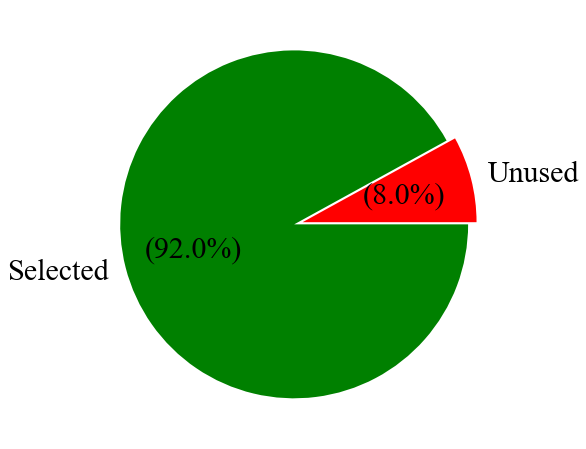}
    }
    \caption{The comparison of selected proposal statistics between BézierLaneNet~\cite{bezier} and our method. Statistics are tested using the ResNet-18 backbone on the test set of CULane~\cite{culane}. The comparison demonstrates that our proposals participate more in the testing phase than its counterpart model.}
    \label{fig:pp-statistics}
    \vspace{-3mm}
\end{figure}

\begin{figure*}[t]
    \centering
    \setlength{\tabcolsep}{0.8pt}
    \begin{tabular}{cccc} 
    	\includegraphics[width=1.7in,totalheight=0.8in]{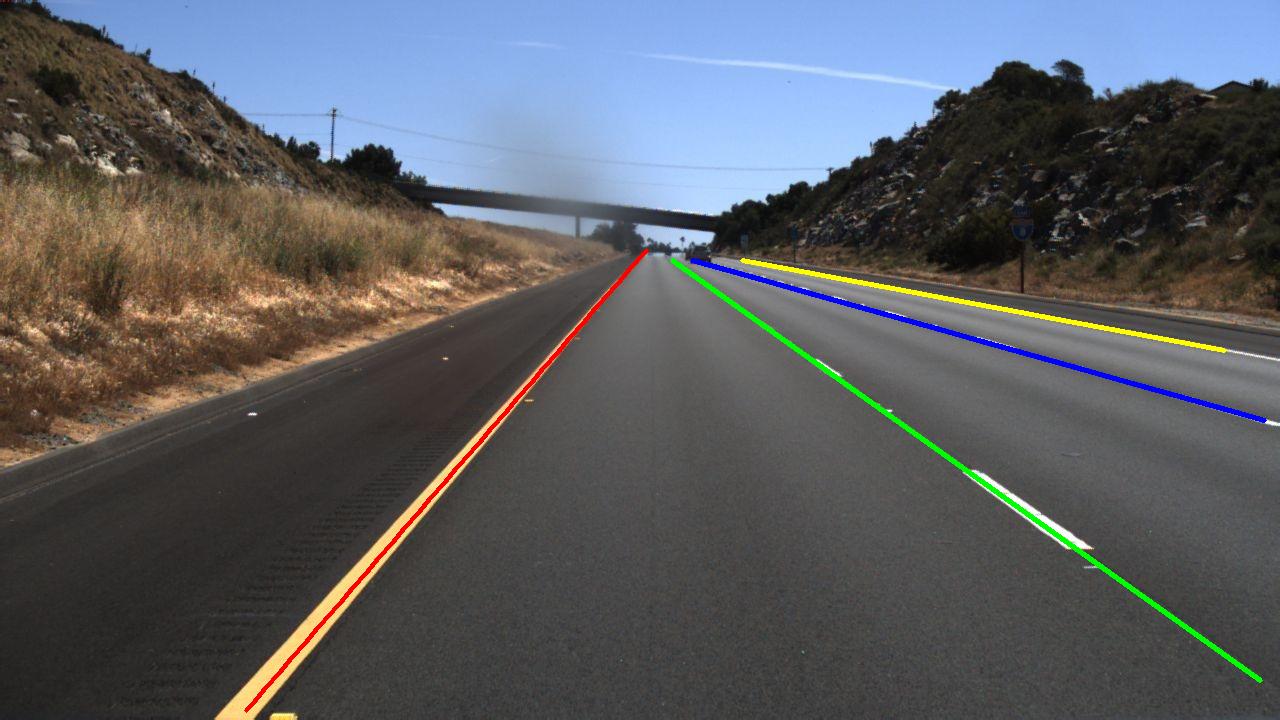}&
    	\includegraphics[width=1.7in,totalheight=0.8in]{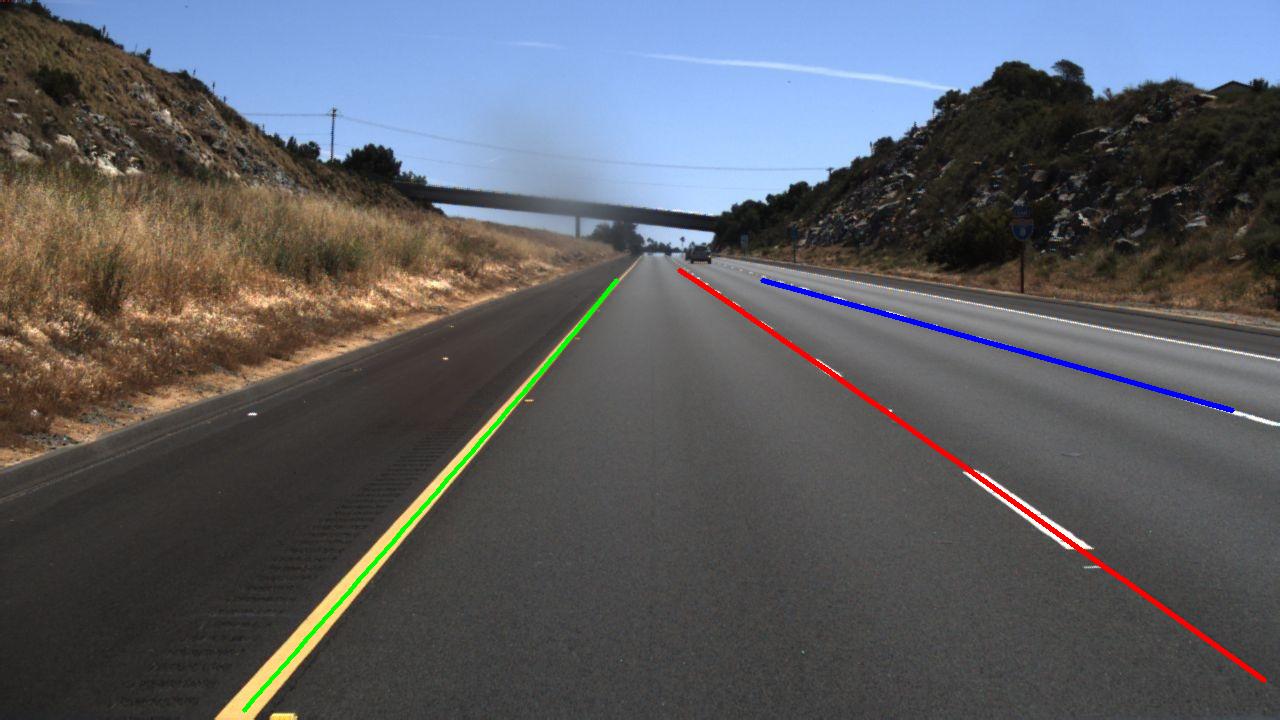}&
    	\includegraphics[width=1.7in,totalheight=0.8in]{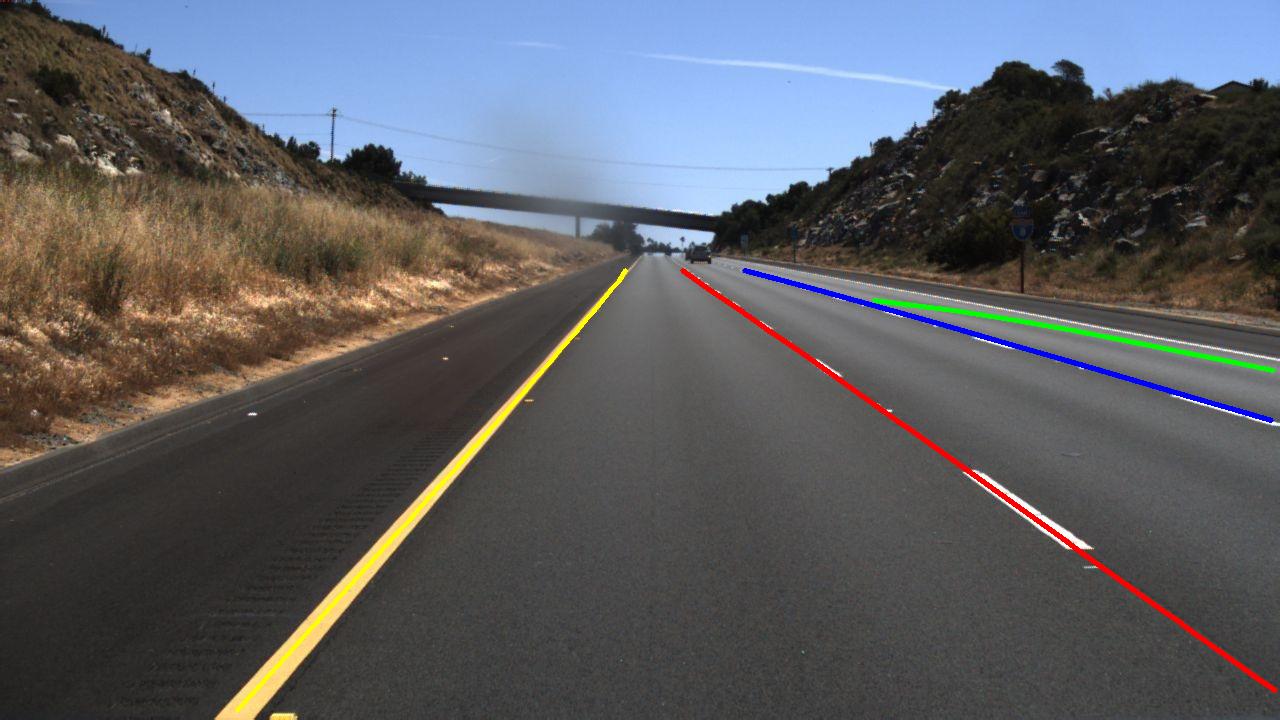}&
    	\includegraphics[width=1.7in,totalheight=0.8in]{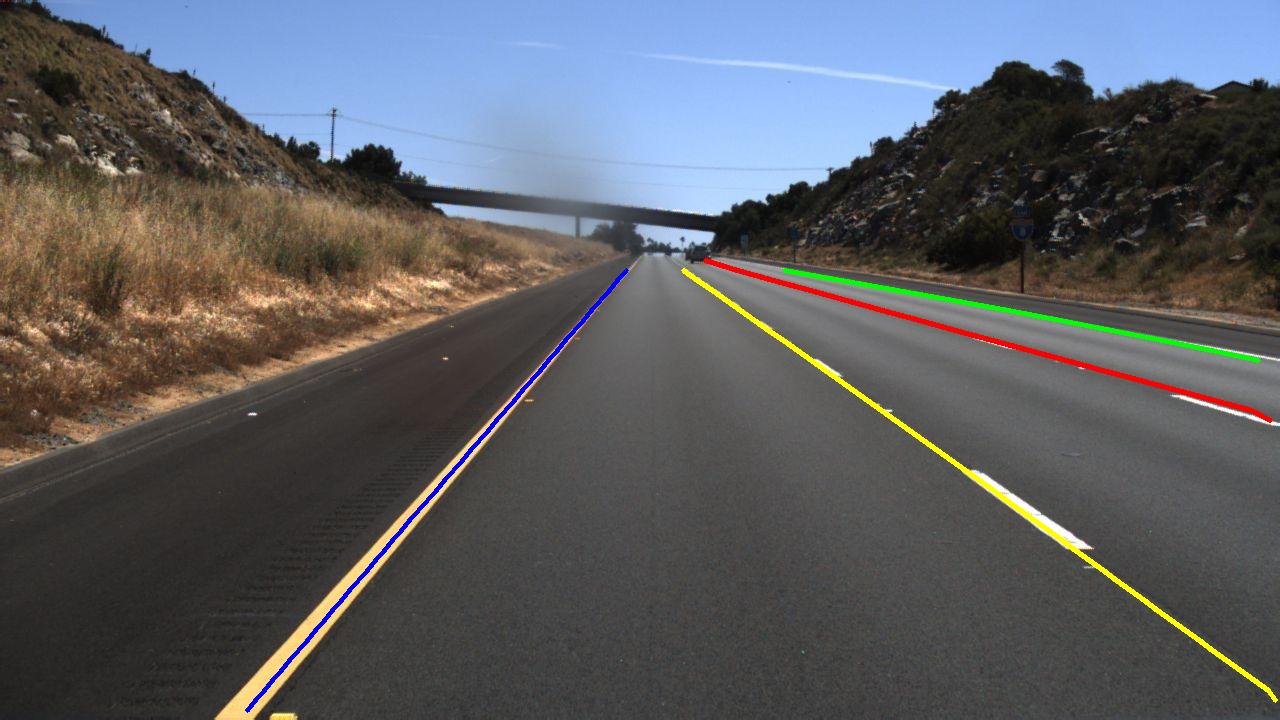}\\
    	\includegraphics[width=1.7in,totalheight=0.8in]{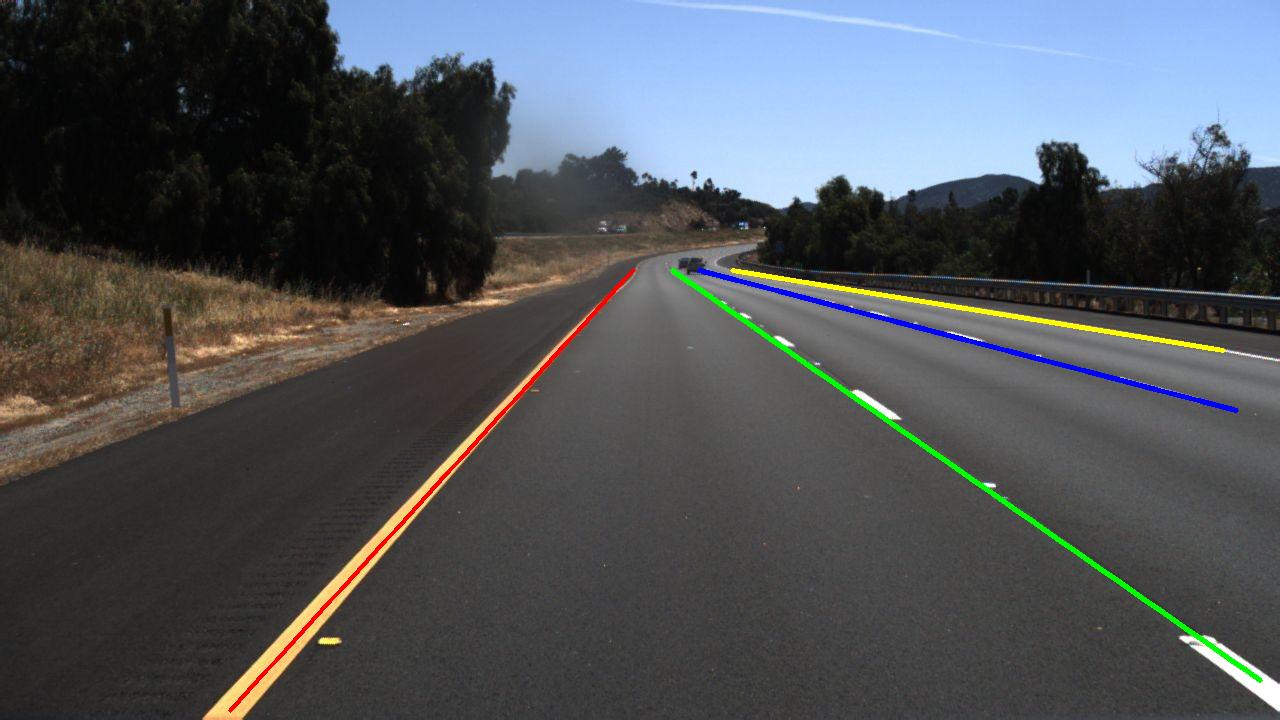}&
    	\includegraphics[width=1.7in,totalheight=0.8in]{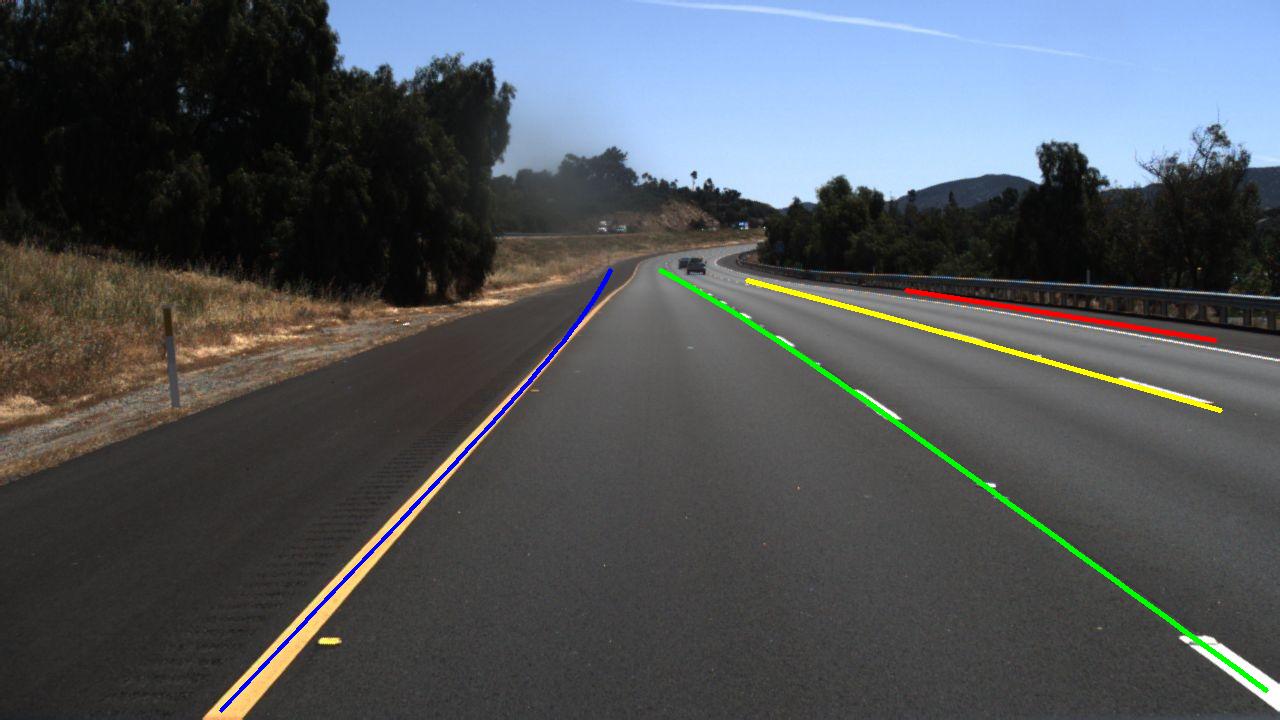}&
    	\includegraphics[width=1.7in,totalheight=0.8in]{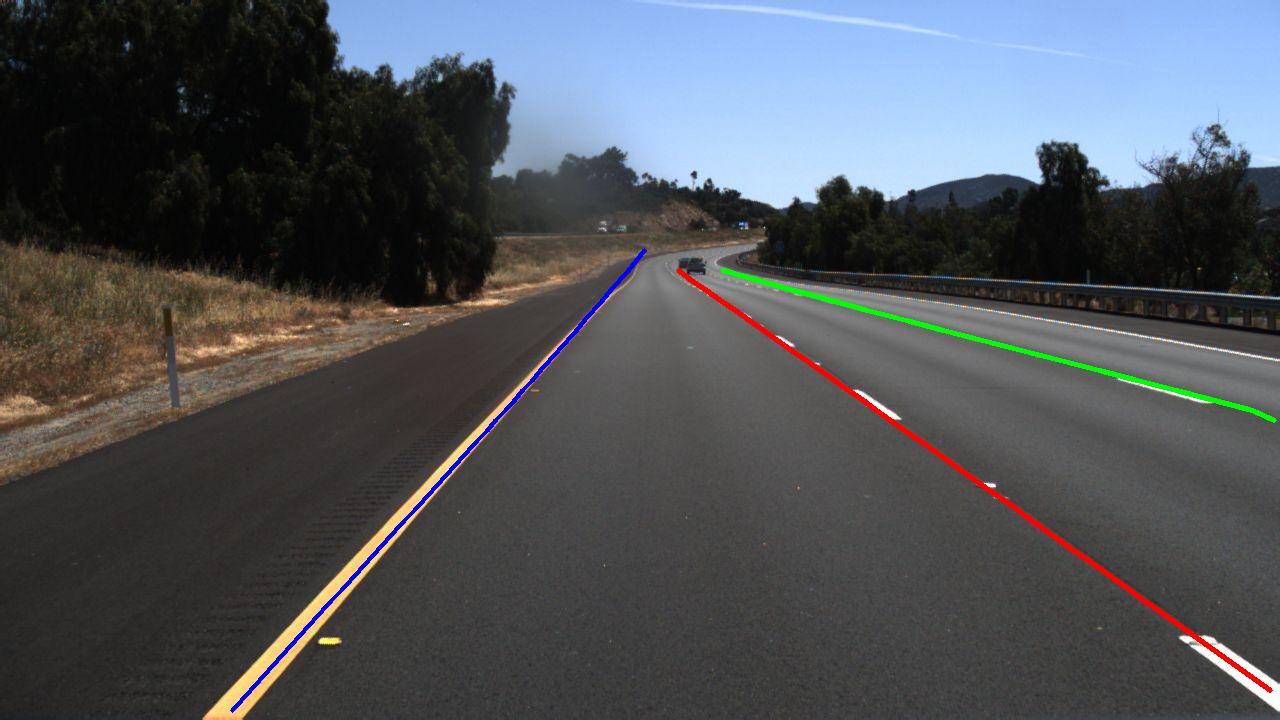}&
    	\includegraphics[width=1.7in,totalheight=0.8in]{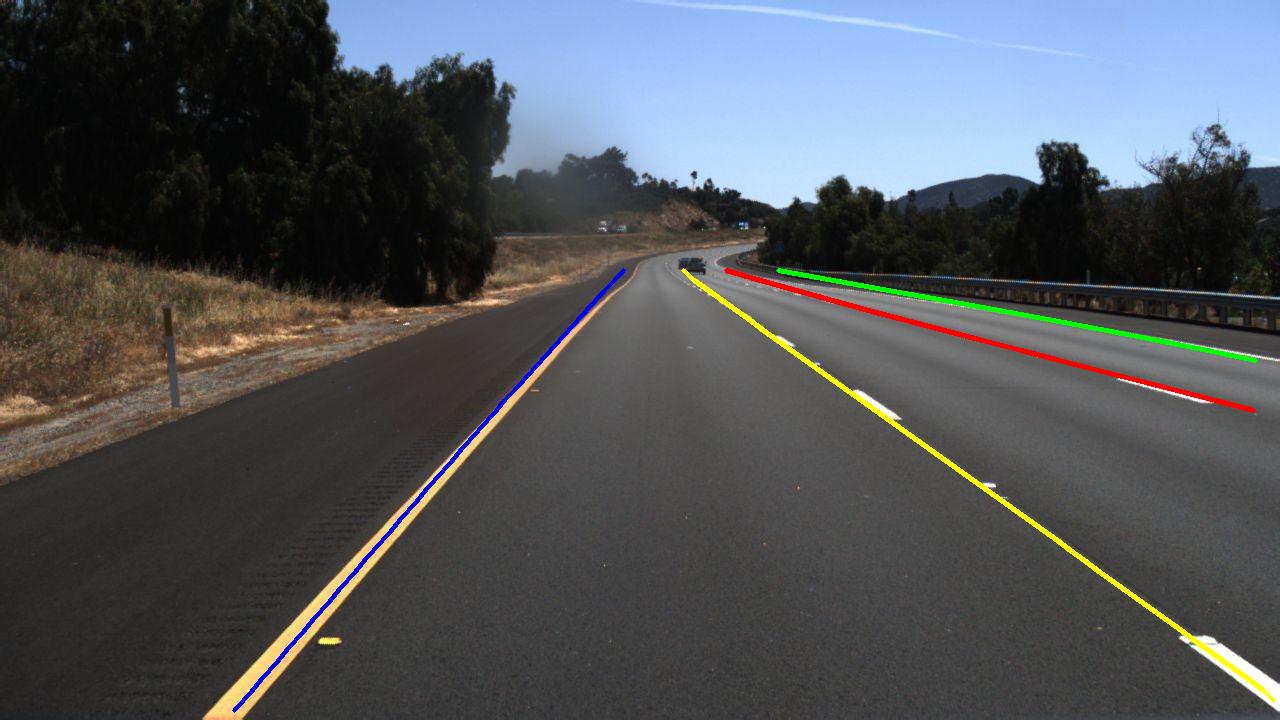}\\
    	\includegraphics[width=1.7in,totalheight=0.8in]{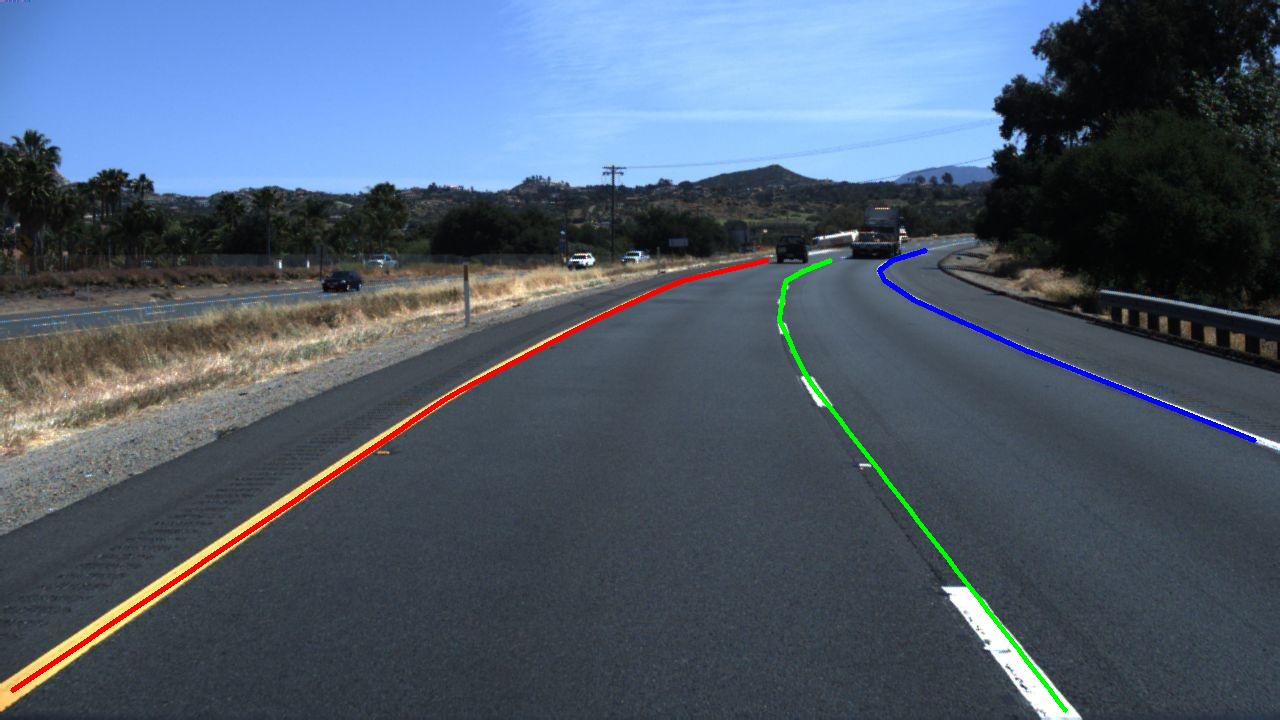}&
    	\includegraphics[width=1.7in,totalheight=0.8in]{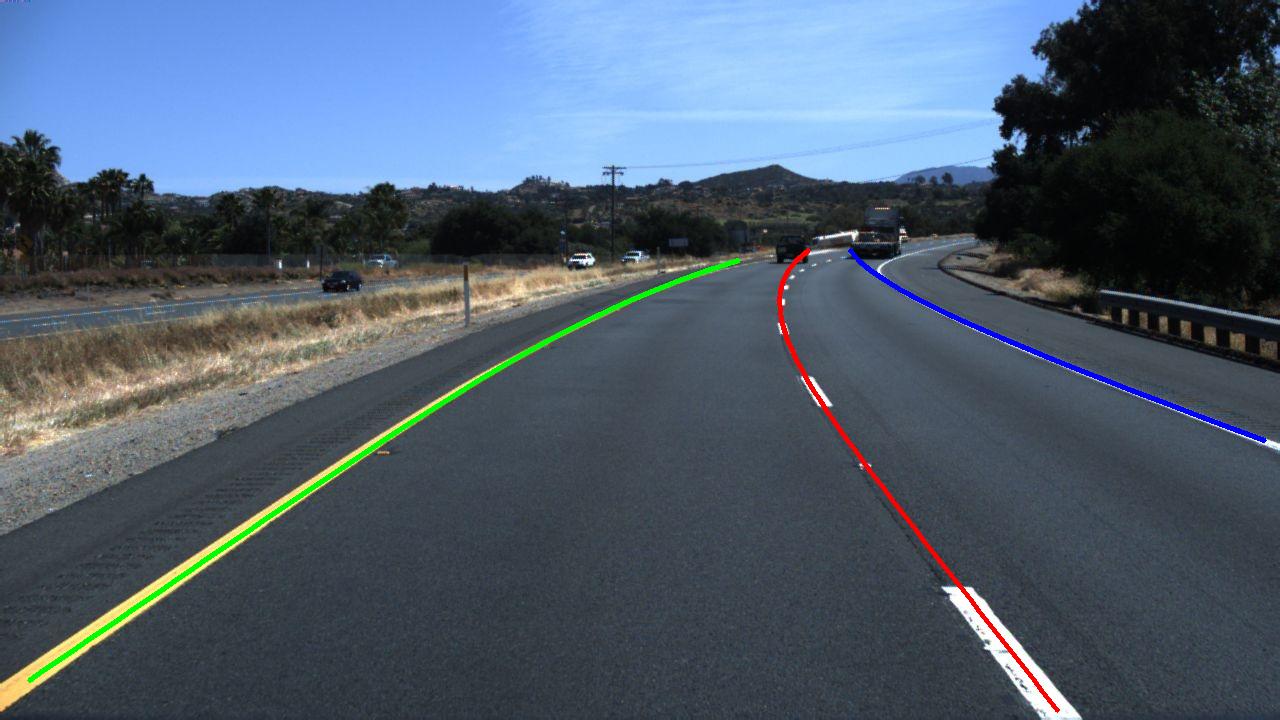}&
    	\includegraphics[width=1.7in,totalheight=0.8in]{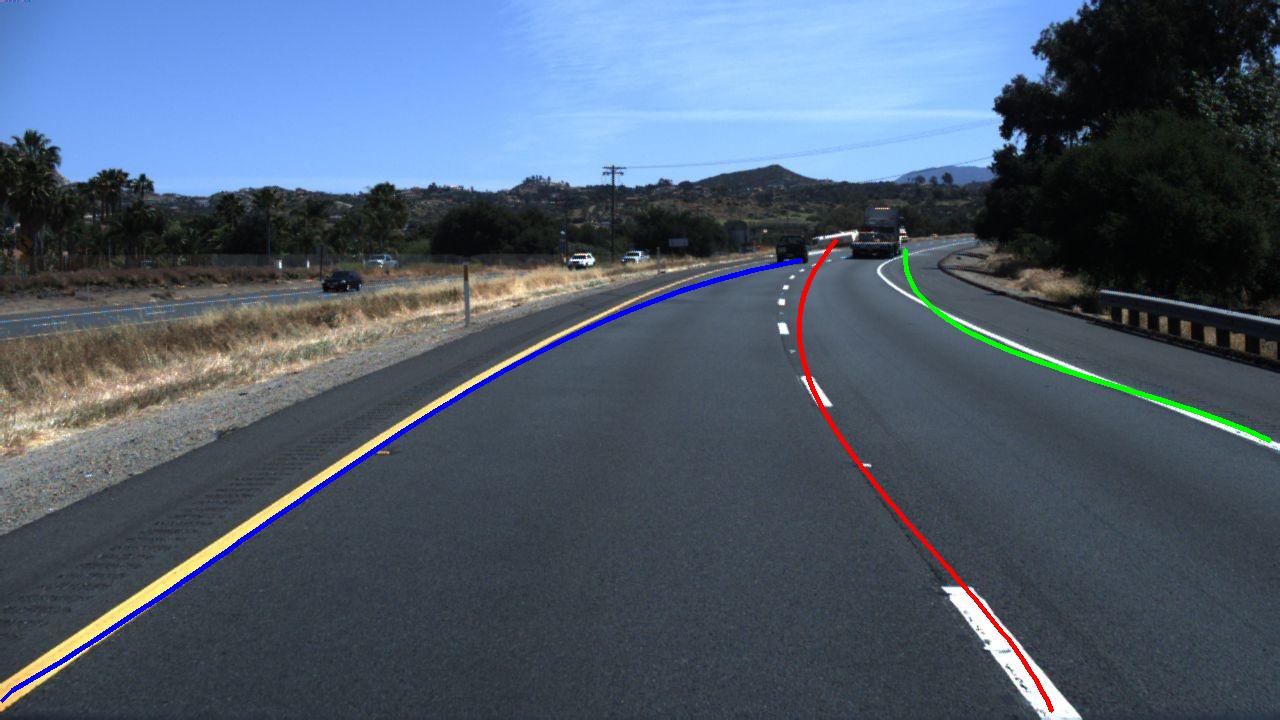}&
    	\includegraphics[width=1.7in,totalheight=0.8in]{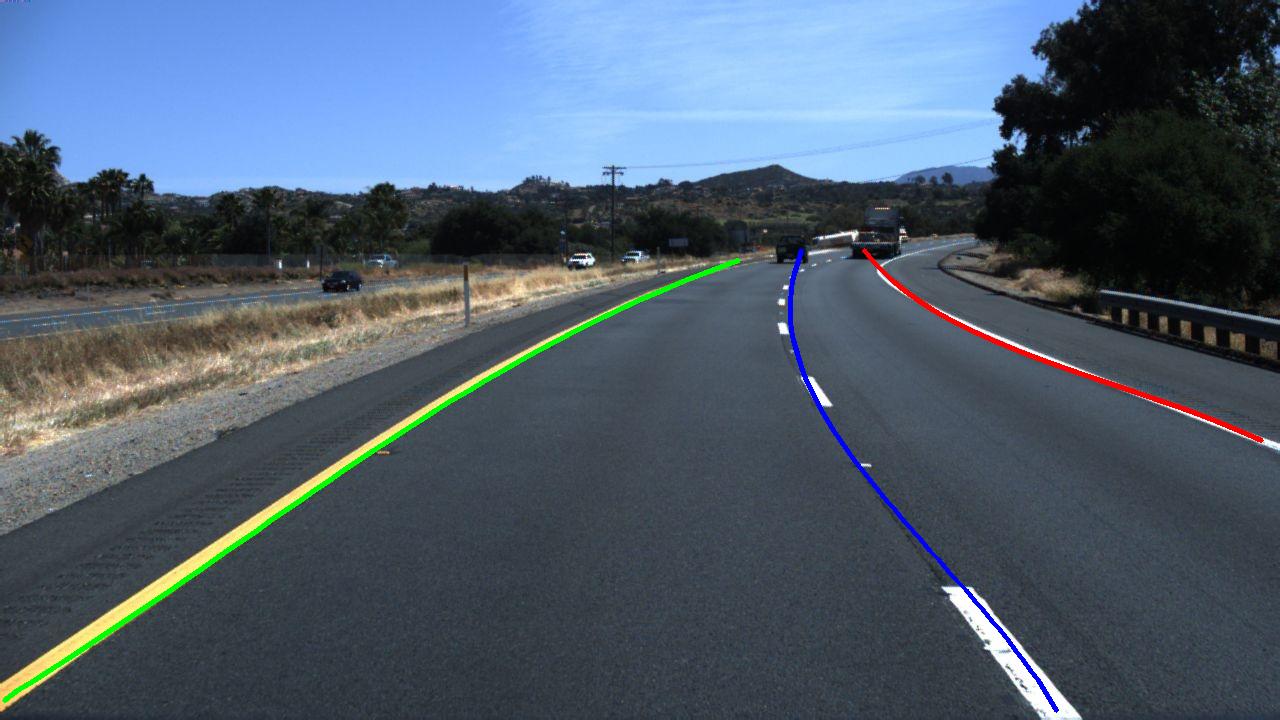}\\
    	{Ground truth} &{LSTR~\cite{LSTR}} &{BézierLaneNet~\cite{bezier}} &{Ours} \\
    \end{tabular}
    \caption{Qualitative results on the TuSimple~\cite{tusimple} test set.}
    \label{fig:tusimple_resultshow}
    \vspace{-4mm}
\end{figure*}

\subsubsection{\textbf{Implementation Details}}
We use the original architecture of BézierLaneNet~\cite{bezier} without changing any architecture, and we apply our DHPM with the three constraints mentioned in Section \ref{sec:model} during the training phase. The batch size is set to be 20 for all datasets, and we train for 400, 36, 20, and 36 epochs for TuSimple~\cite{tusimple}, CULane~\cite{culane}, LLAMAS~\cite{llamas} and CurveLanes~\cite{CurveLane}, respectively. The input images are resized to $360 \times 640$ for TuSimple~\cite{tusimple} and LLAMAS~\cite{llamas}, $720\times1280$ for CurveLanes~\cite{CurveLane}, while $288\times800$ for CULane~\cite{culane}. Given the number of proposals $K$, the upper limits in our diversity constraint were set to $U=\lfloor \frac{4K}{25} \rfloor$.
We use Adam optimizer and Cosine Annealing learning rate schedule~\cite{LaneATT} with an initial learning rate of $6\times10^{-4}$. All experiments were based on PyTorch. The data augmentation contains random affine transforms, the random horizontal flip of $50 \%$ probability, and color jitter, where random affine transforms include random rotation within $10$ degrees, random sampling ratio from $0.8$ to $1.2$.

\subsection{Main Results}
In this subsection, we compare various state-of-the-art methods on four large-scale datasets. Following BézierLaneNet~\cite{bezier}, we use ResNet-18 and ResNet-34 as our backbones. 

\begin{figure*}[t]
    \centering
    \setlength{\tabcolsep}{0.8pt}
    \begin{tabular}{ccccc} 
    	\includegraphics[width=1.36in,totalheight=0.68in]{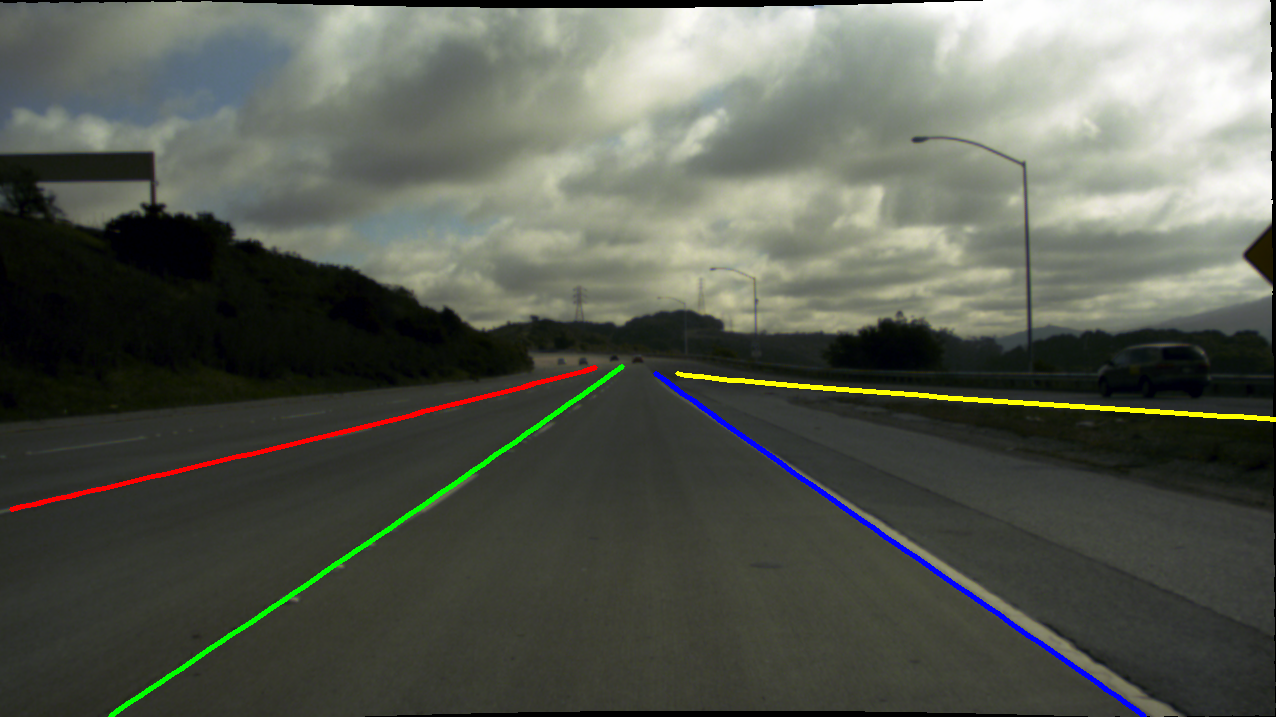}&
    	\includegraphics[width=1.36in,totalheight=0.68in]{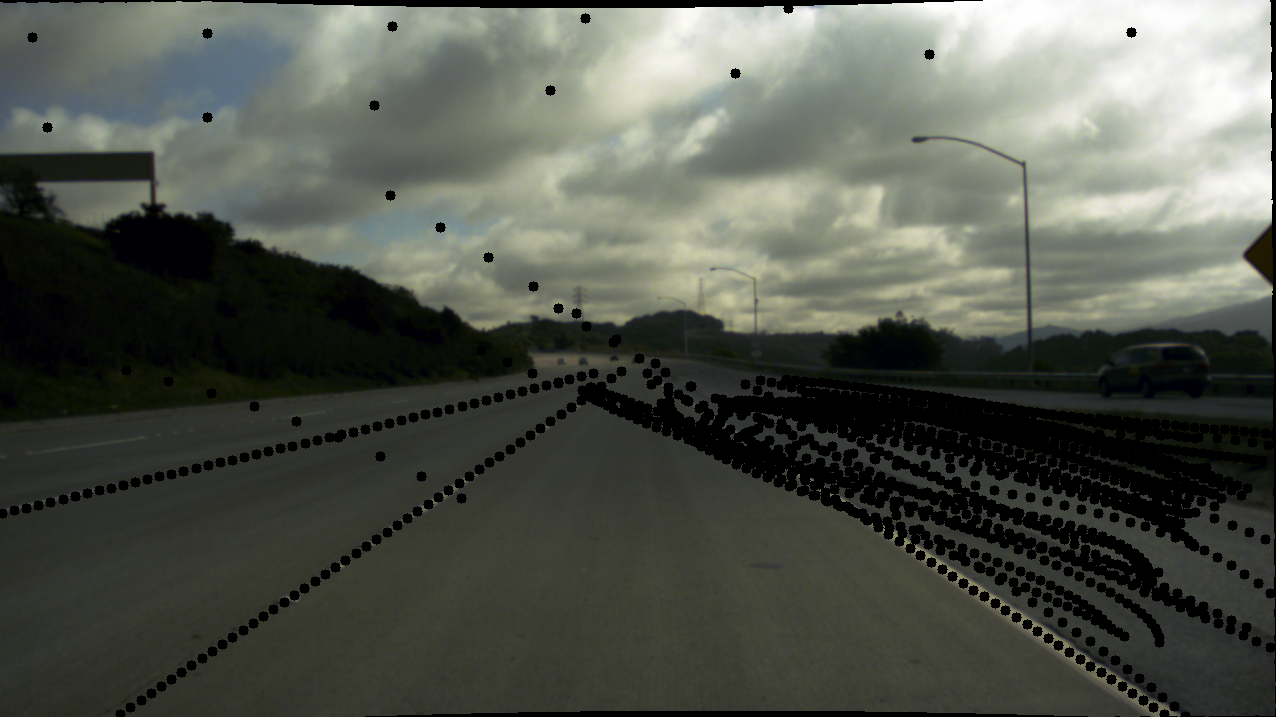}&
    	\includegraphics[width=1.36in,totalheight=0.68in]{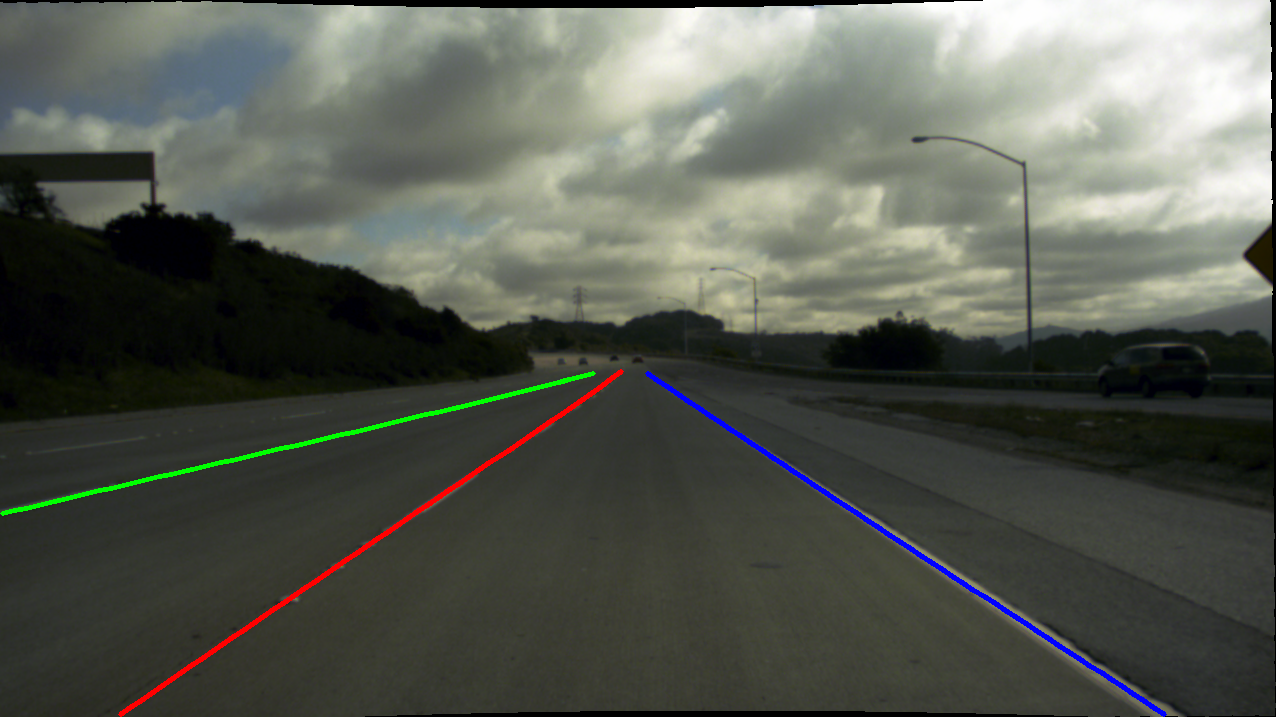}&
    	\includegraphics[width=1.36in,totalheight=0.68in]{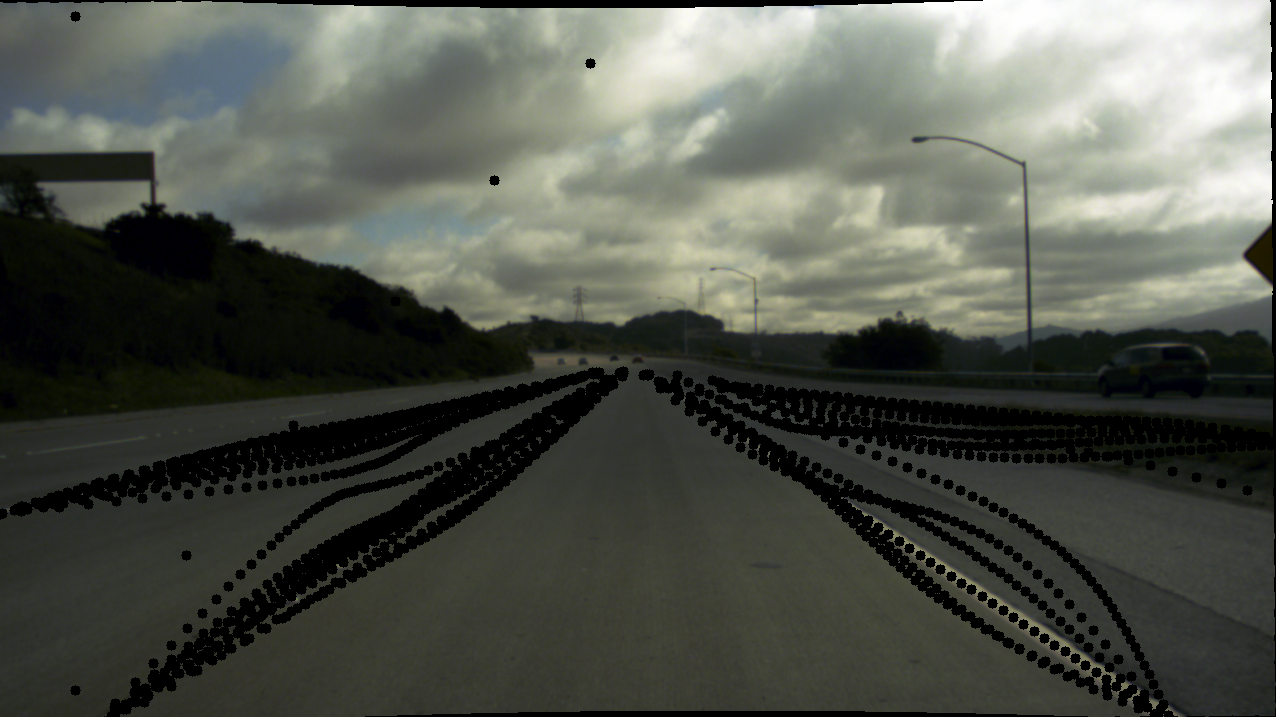}&
    	\includegraphics[width=1.36in,totalheight=0.68in]{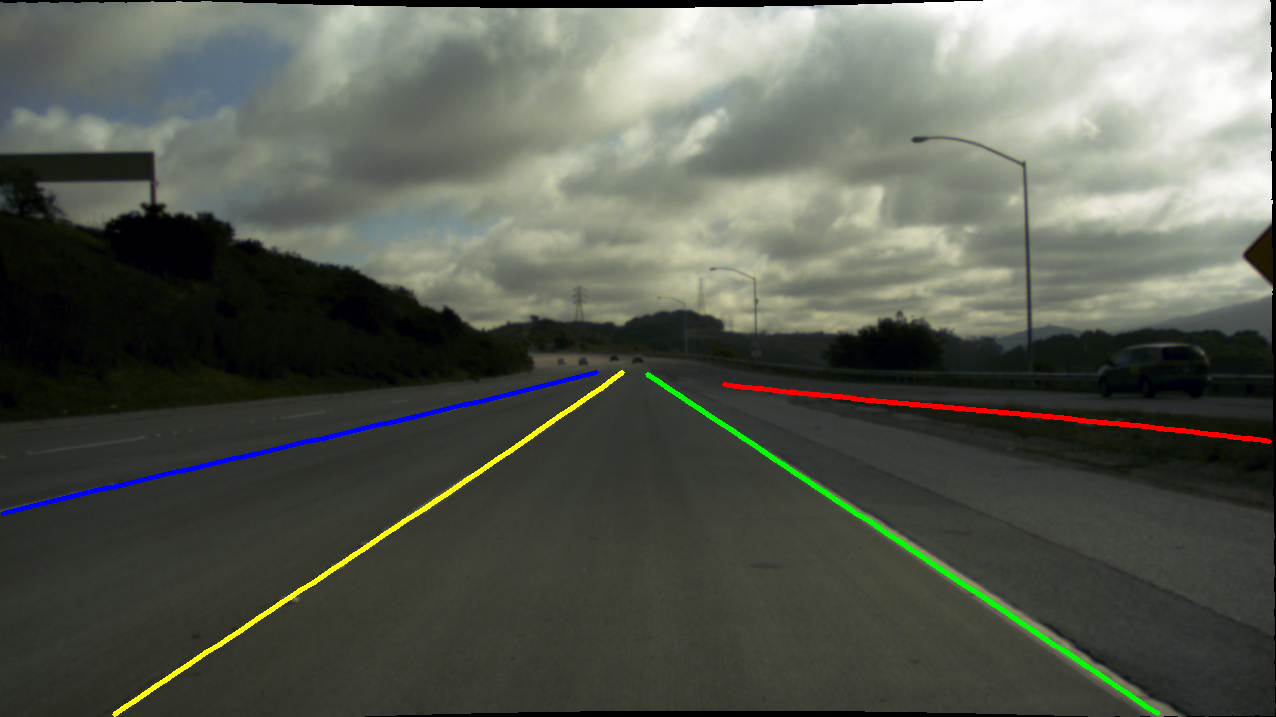}\\
    	\includegraphics[width=1.36in,totalheight=0.68in]{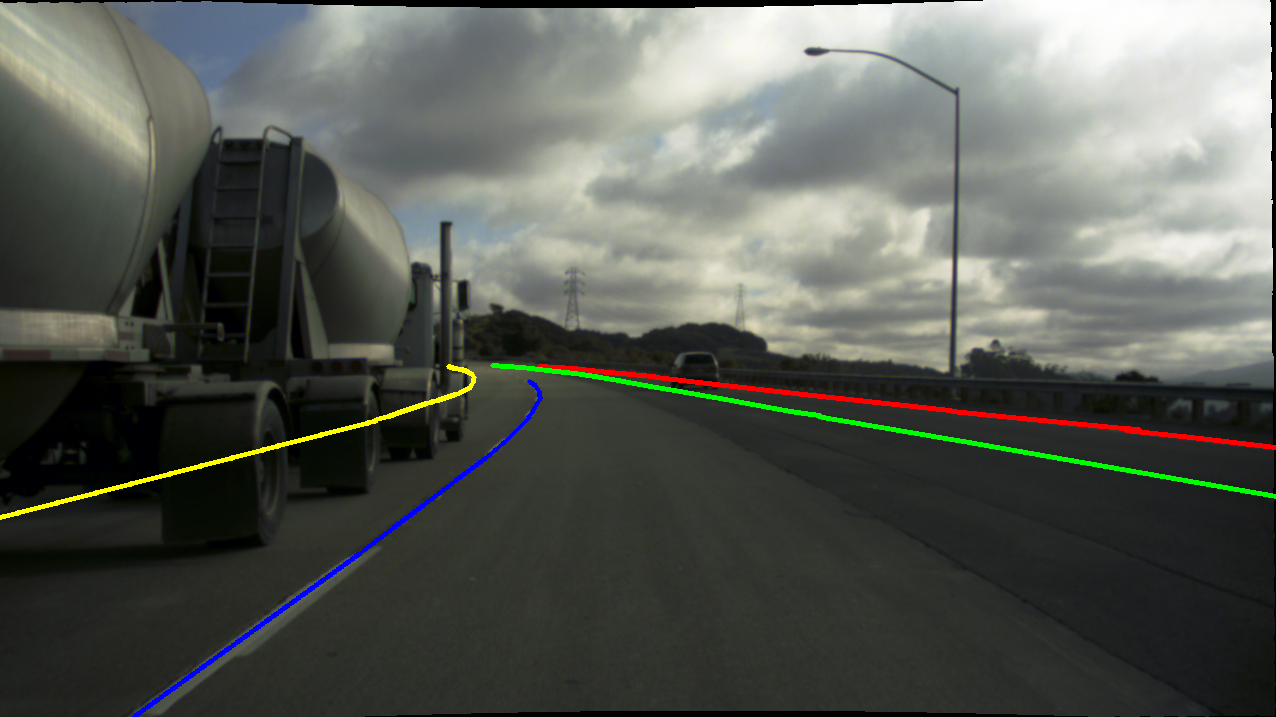}&
    	\includegraphics[width=1.36in,totalheight=0.68in]{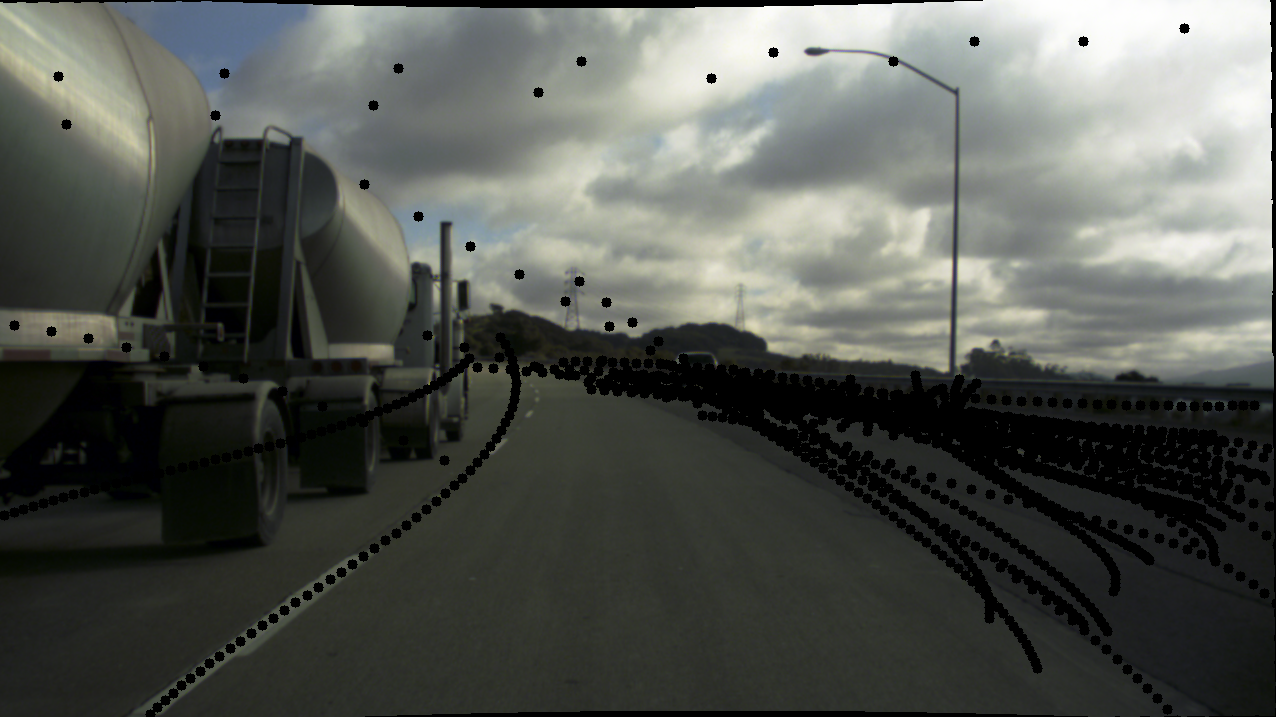}&
    	\includegraphics[width=1.36in,totalheight=0.68in]{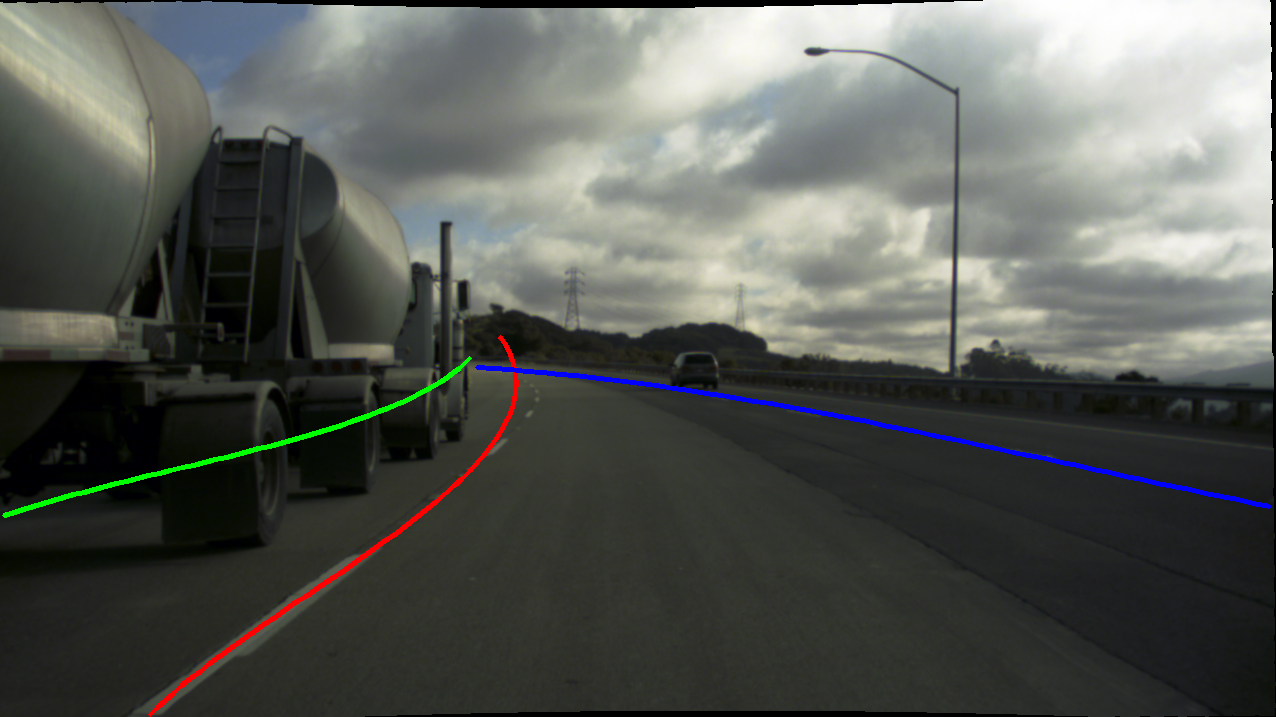}&
    	\includegraphics[width=1.36in,totalheight=0.68in]{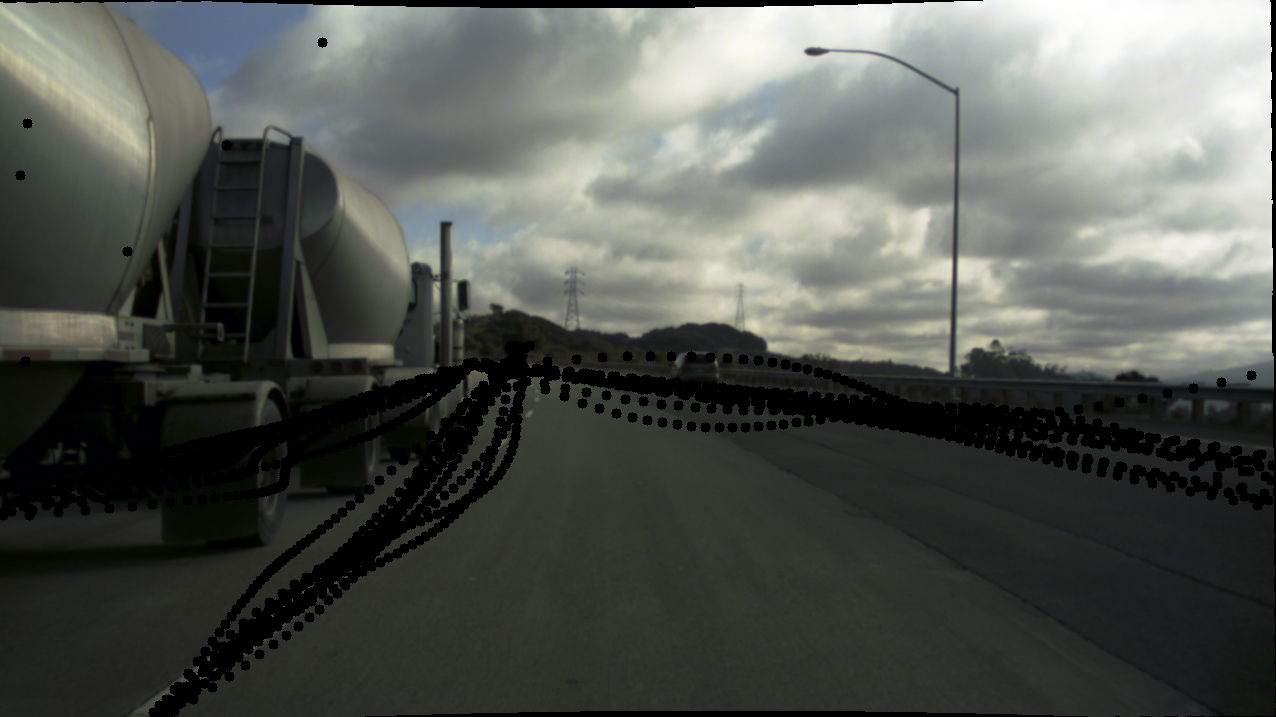}&
    	\includegraphics[width=1.36in,totalheight=0.68in]{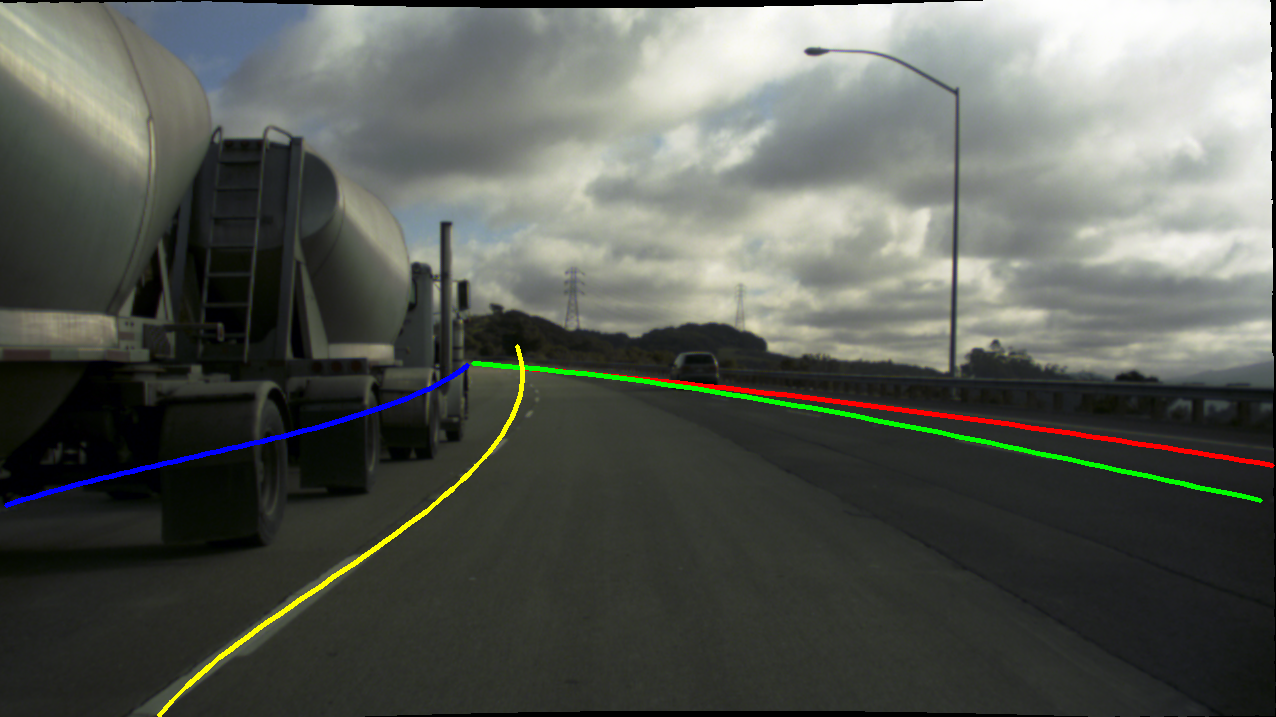}\\
    	\includegraphics[width=1.36in,totalheight=0.68in]{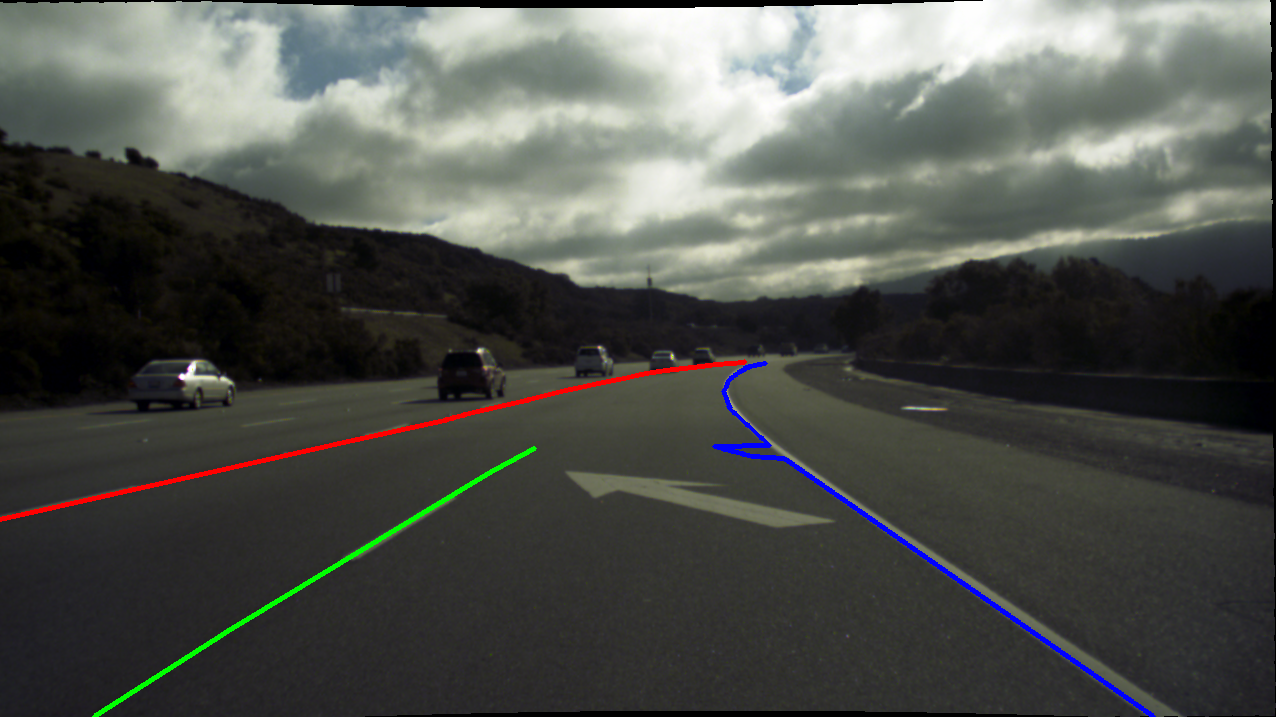}&
    	\includegraphics[width=1.36in,totalheight=0.68in]{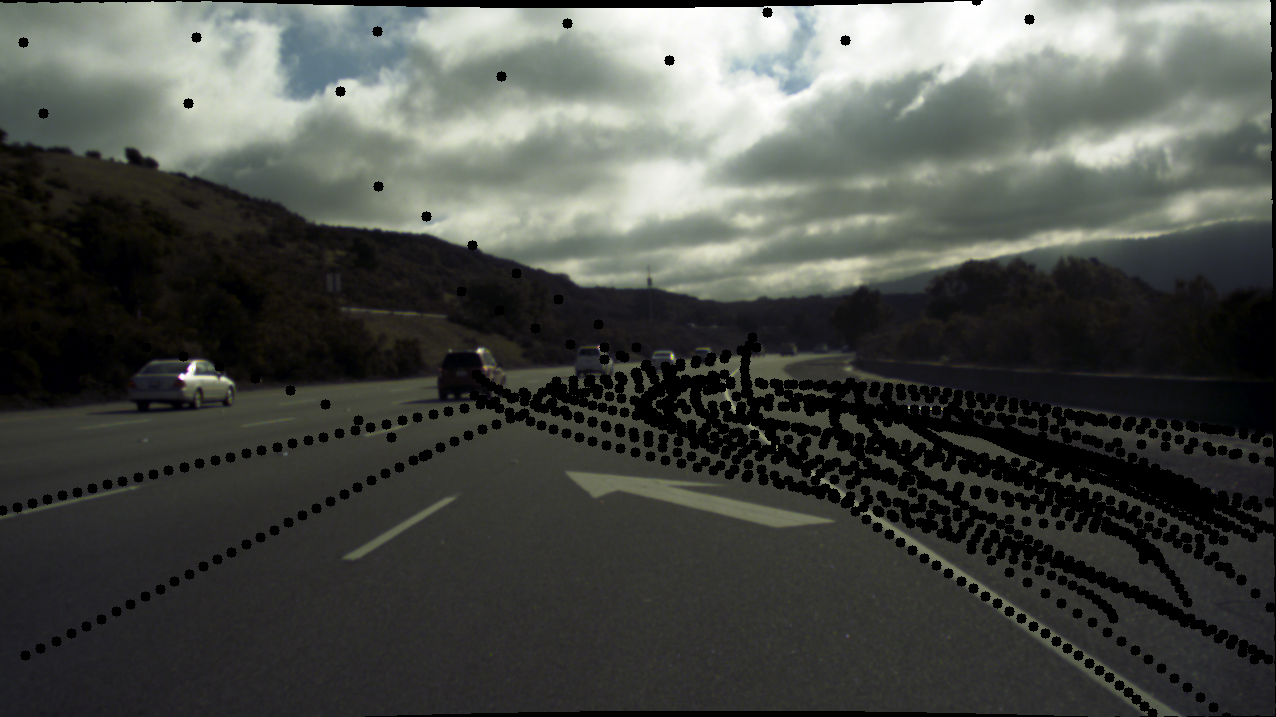}&
    	\includegraphics[width=1.36in,totalheight=0.68in]{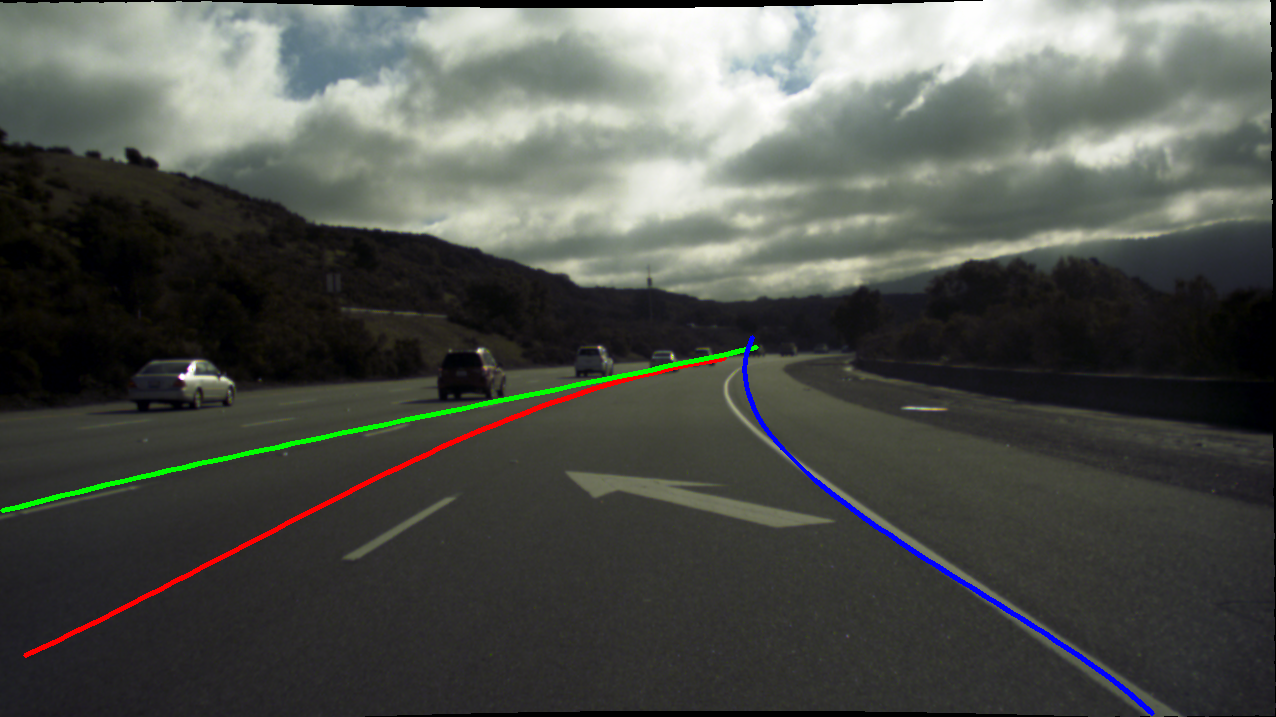}&
    	\includegraphics[width=1.36in,totalheight=0.68in]{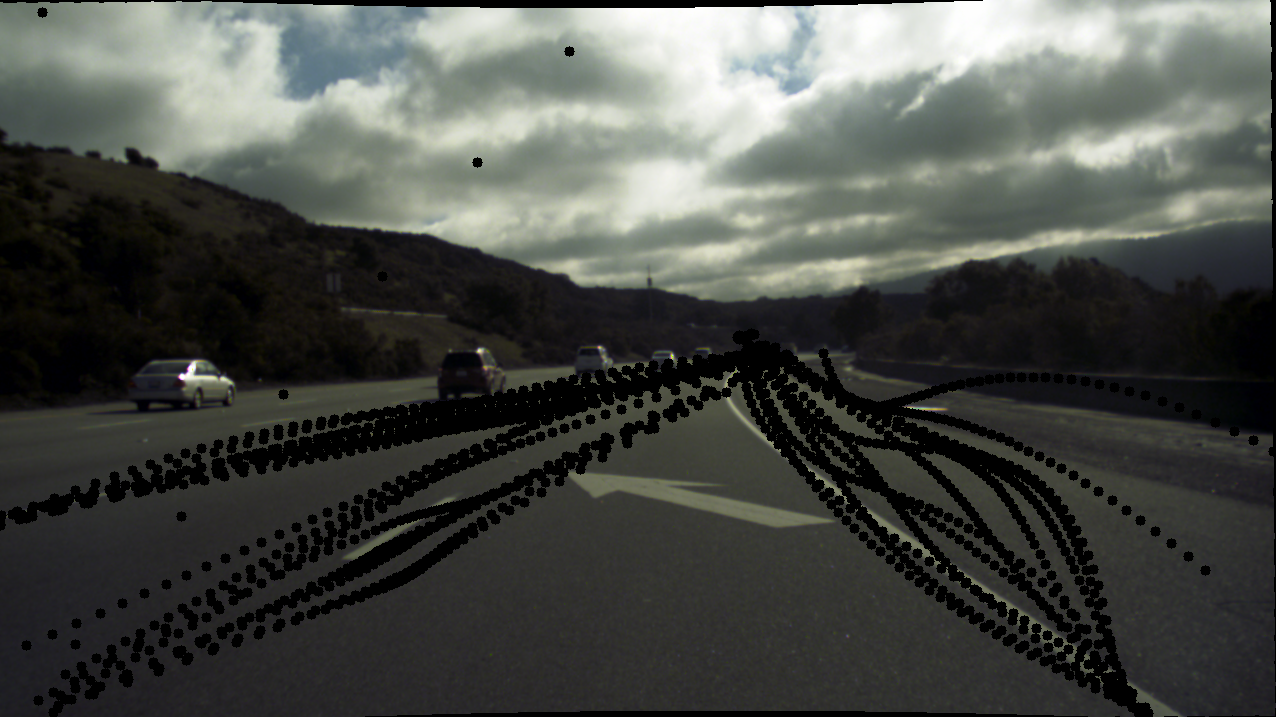}&
    	\includegraphics[width=1.36in,totalheight=0.68in]{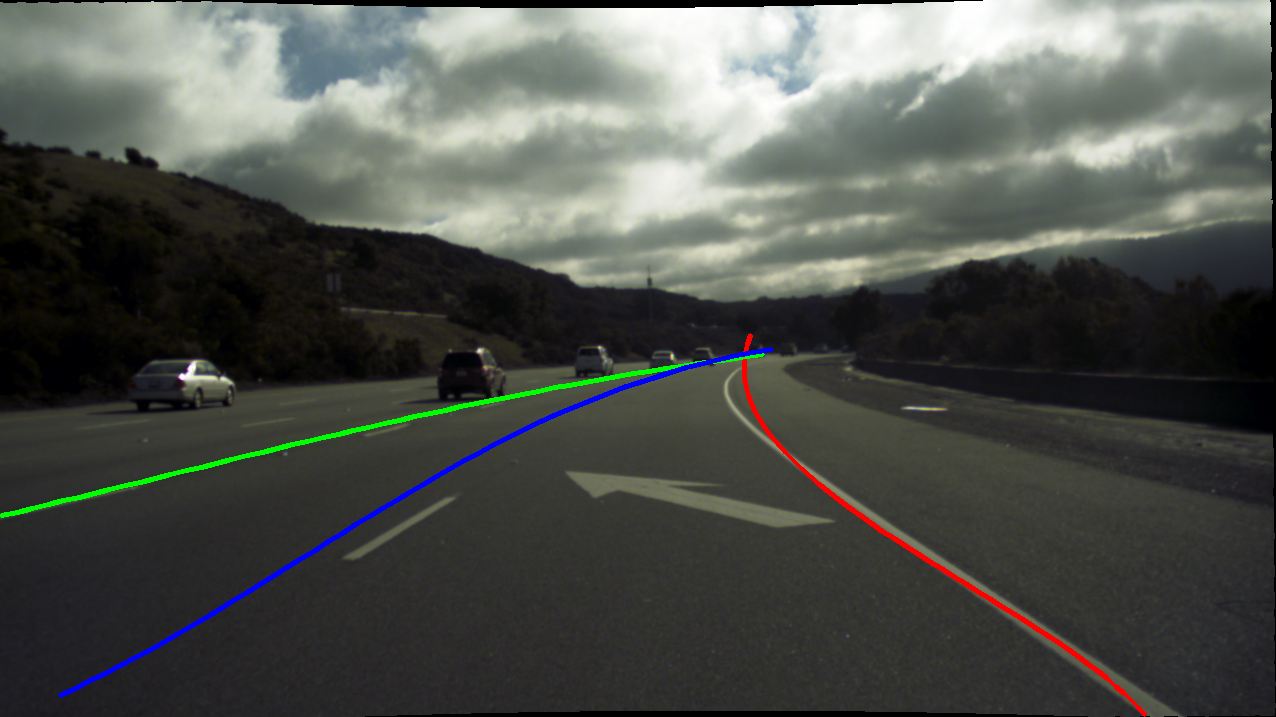}\\
    	\includegraphics[width=1.36in,totalheight=0.68in]{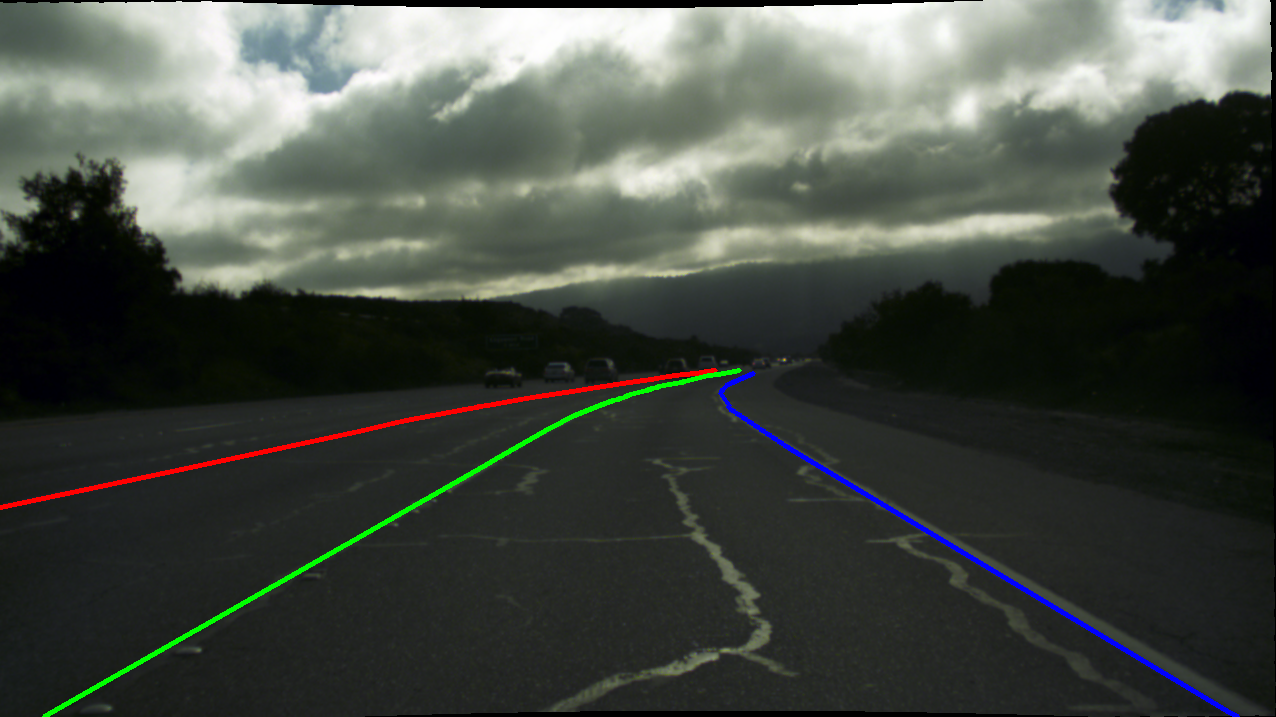}&
    	\includegraphics[width=1.36in,totalheight=0.68in]{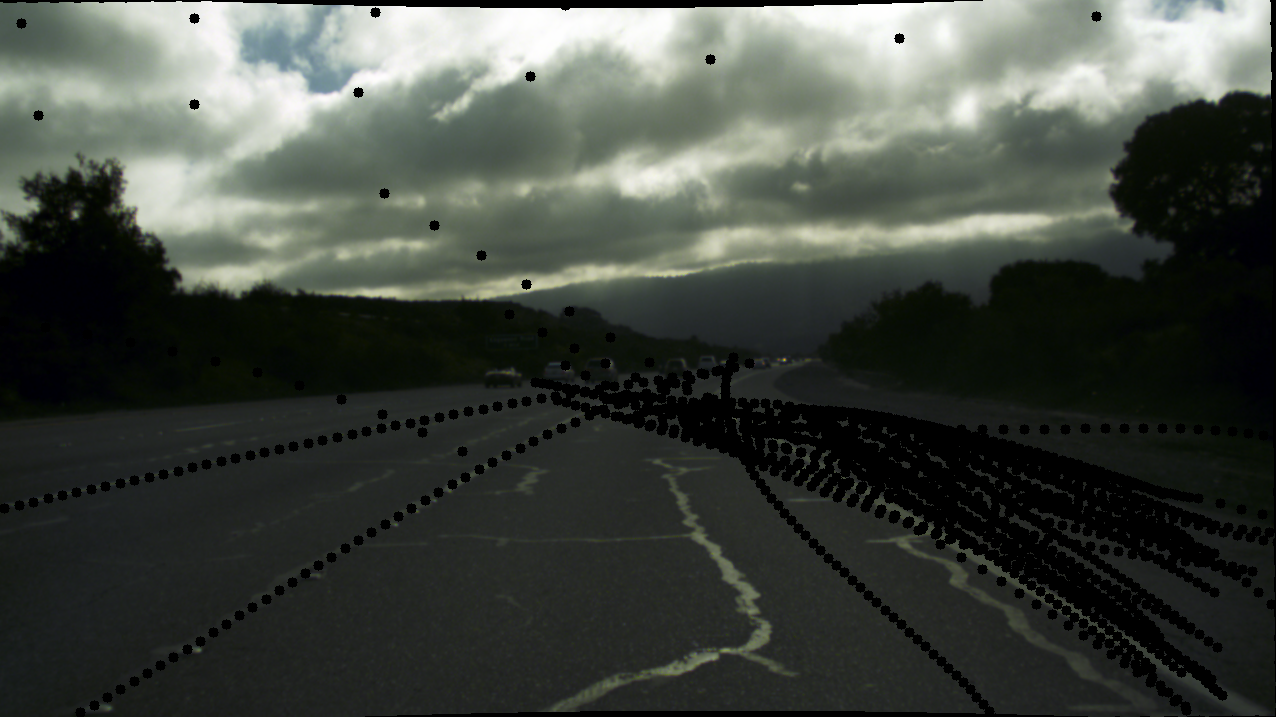}&
    	\includegraphics[width=1.36in,totalheight=0.68in]{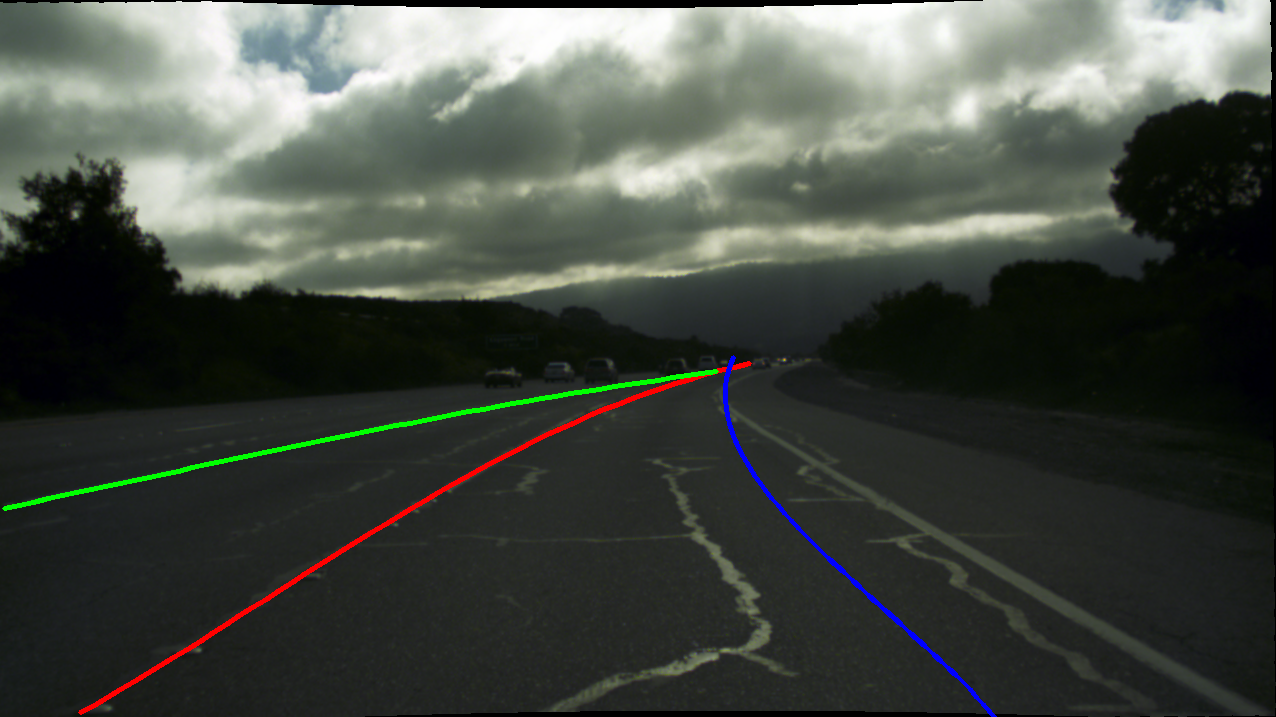}&
    	\includegraphics[width=1.36in,totalheight=0.68in]{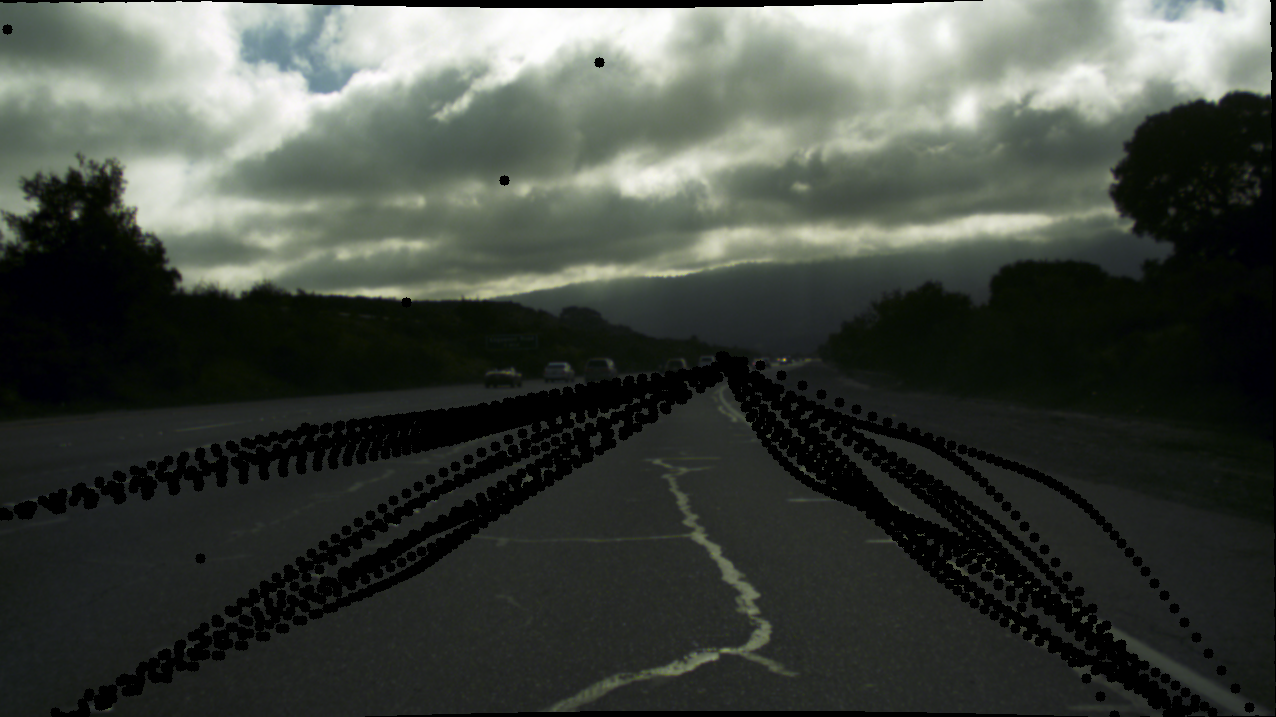}&
    	\includegraphics[width=1.36in,totalheight=0.68in]{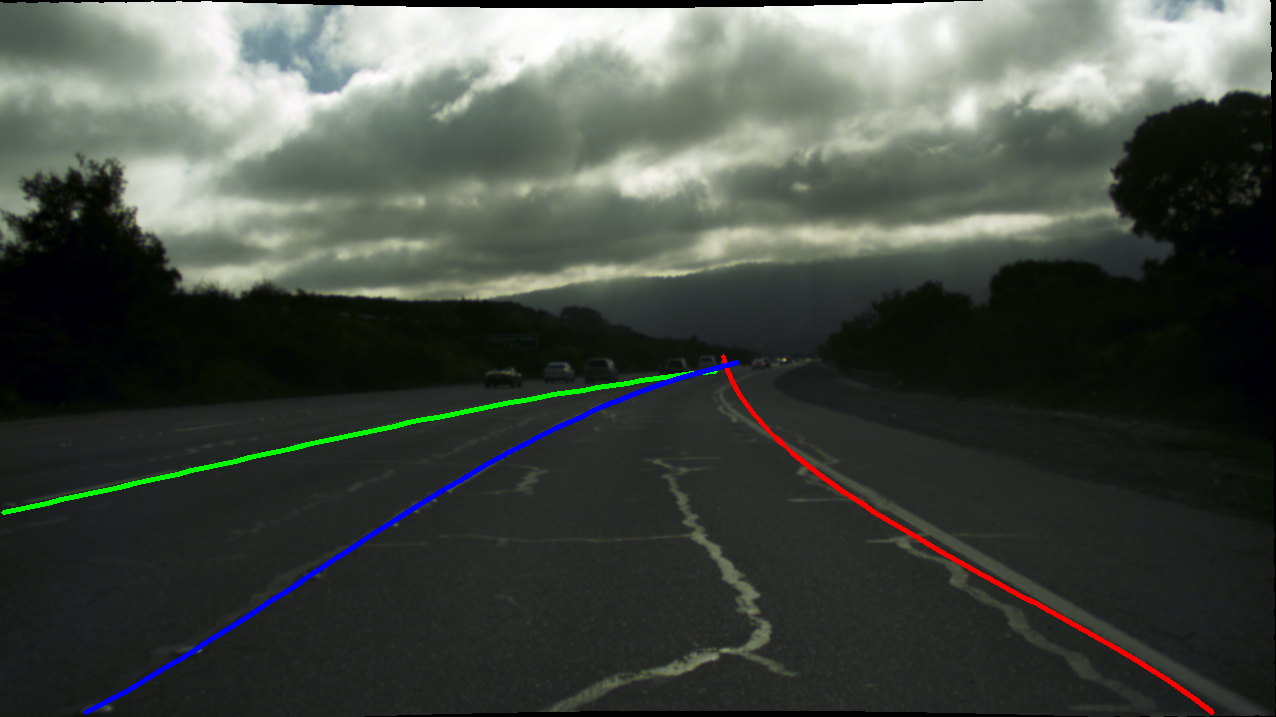}\\
    	{Ground truth} &{\cite{bezier}'s proposals} &{\cite{bezier}'s predictions} &{Our proposals} &{Our predictions} \\
    \end{tabular}
    \caption{Qualitative comparison on the LLAMAS~\cite{llamas} validation set.}
    \label{fig:llamas_resultshow}
    \vspace{-4mm}
\end{figure*}

\subsubsection{\textbf{CULane~\cite{culane}}}
On the CULane dataset, we compare our DHPM with other 12 methods, including SCNN~\cite{culane}, SAD~\cite{ENet-SAD}, UFLD~\cite{UFAST}, SIM-CycleGAN~\cite{SIM}, CurveLanes~\cite{CurveLane}, LaneATT~\cite{LaneATT}, RESA~\cite{RESA}, UFLDv2~\cite{UFLDv2}, CondLaneNet~\cite{CondLaneNet}, CLRNet~\cite{CLRNet}, LSTR (ResNet-18)*~\cite{LSTR}, and BézierLaneNet~\cite{bezier}. As shown in Table~\ref{tab:culane-compare}, our DHPM achieves state-of-the-art performance among the curve-based methods. In addition, we consistently outperform the baseline models across different backbones. Specifically, our lightweight model achieves a $0.92$ improvement on the F1 score with the same testing speed and without introducing new parameters to the model. In addition, {we achieve the lowest FP on the ``Crossroad" category across all kinds of methods,} where $295$ false positive predictions are reduced compared with the baseline. These results demonstrate the effectiveness of our dense hybrid modulation method. The visual comparisons between our DHPM and other methods are shown in Fig.~\ref{fig:culane_resultshow}. Besides, our method is not specific to curve-based methods. We also conduct experiments on LaneATT~\cite{LaneATT} as a new baseline. The experimental results show that our method also achieves a significant improvement over LaneATT, which demonstrates that our method has broad application prospects in the task of lane detection.

To prove the improvements in proposal quality, we compare the selected proposal statistics with the baseline on the test split of CULane. As shown in Fig.~\ref{fig:pp-statistics}, our proposals participate more in the testing phase than the counterpart model. More proposals are improved to have reasonable locations and shapes, which is the basis for contributing their value.

\subsubsection{\textbf{TuSimple~\cite{tusimple}}}
On the TuSimple dataset, we compare with current state-of-the-art methods and report the results in Table \ref{tab:tusimple-result}. We achieve higher accuracy and lower FN metrics with comparable FP compared to BézierLaneNet~\cite{bezier}. Although the metrics of TuSimple are almost saturated, {we still achieve very competitive performances among all kinds of methods.} For both backbones, we consistently improve the performance over the baseline counterparts, which proves the effective design of our method. The visual comparisons between our DHPM and other methods are shown in Fig.~\ref{fig:tusimple_resultshow}.

\begin{table}[tb]
  \begin{center}
    \caption{Quantitative Results on the Tusimple~\cite{tusimple} Dataset}
    \renewcommand{\arraystretch}{1.3}
        \begin{tabular}{lccc} 
            \toprule
            \textbf{Method}                               & \textbf{Accuracy} & \textbf{FP} & \textbf{FN}  \\ 
            \hline
            \textbf{Segmentation-based Method}            &                   &             &              \\
            \cline{1-1}
            SCNN (LargeFOV)~\cite{culane}                 & 96.53             & 6.17        & 1.80         \\
            EL-GAN~\cite{EL-GAN}                          & 96.39             & 4.12        & 3.36         \\
            SAD (ENet)~\cite{ENet-SAD}                    & 96.64             & 6.02        & 2.05         \\
            \cline{1-4}
            \textbf{Anchor-based Method}                  &                   &             &              \\
            \cline{1-1}
            PointLaneNet (MobileNet-v2)~\cite{PointLaneNet}& 96.34             & 4.67        & 5.18         \\
            LaneATT (ResNet-18) ~\cite{LaneATT}            & 95.57             & 3.56        & 3.01         \\
            UFLD (ResNet-18) ~\cite{UFAST}                 & 95.82             & 19.05       & 3.92         \\
            UFLD (ResNet-34) ~\cite{UFAST}                 & 95.86             & 18.91       & 3.75         \\
            UFLDv2 (ResNet-18)~\cite{UFLDv2}               & 95.50             & 3.06        & 4.82         \\
            UFLDv2 (ResNet-34)~\cite{UFLDv2}               & 95.56             & 3.18        & 4.37         \\
            CondLaneNet (ResNet-18)~\cite{CondLaneNet}     & 95.48             & 2.18        & 3.80         \\
            CLRNet (ResNet-18)~\cite{CLRNet}               & 96.84             & 2.28        & 1.92         \\
            \cline{1-4}
            \textbf{Curve-based Method}                    &                   &             &              \\
            \cline{1-1}
            PolyLaneNet (EfficientNet-b0)~\cite{PolyLaneNet}         & 93.36             & 9.42        & 9.33         \\
            LSTR (ResNet18)~\cite{LSTR}                    & 96.18             & 2.91        & 3.38         \\
            BézierLaneNet (ResNet-18)~\cite{bezier}        & 95.41             & 5.30        & 4.60        \\
            BézierLaneNet (ResNet-34)~\cite{bezier}        & 95.65             & 5.10        & 3.90        \\ 
            \textbf{ours-BézierLaneNet (ResNet-18)}        & 95.61             & 5.30        & 3.50        \\
            \textbf{ours-BézierLaneNet (ResNet-34)}        & 95.87             & 5.00        & 3.40        \\
            \bottomrule
        \end{tabular}
        \label{tab:tusimple-result}
    \end{center}
    \vspace{-5mm}
\end{table}

\subsubsection{\textbf{LLAMAS~\cite{llamas}}}
We also compare DHPM with recent top-performing approaches on the online benchmark of LLAMAS.
As shown in Table~\ref{tab:llamas-result}, we achieve very competitive results on F1 and Recall metrics. Compared to the baseline model, we consistently improve all metrics by a relatively clear advantage. Our ResNet-18 model is comparable to BézierLaneNet's ResNet-34 model, indicating that our proposal has better overall quality than BézierLaneNet. The visual comparisons between DHPM and others are shown in Fig.~\ref{fig:llamas_resultshow}.

\subsubsection{{\textbf{CurveLanes~\cite{CurveLane}}}}
\begin{figure*}[t]
\centering
    \setlength{\tabcolsep}{0.8pt}
    \begin{tabular}{ccc} 
        \includegraphics[width=2.0in,totalheight=0.9in]{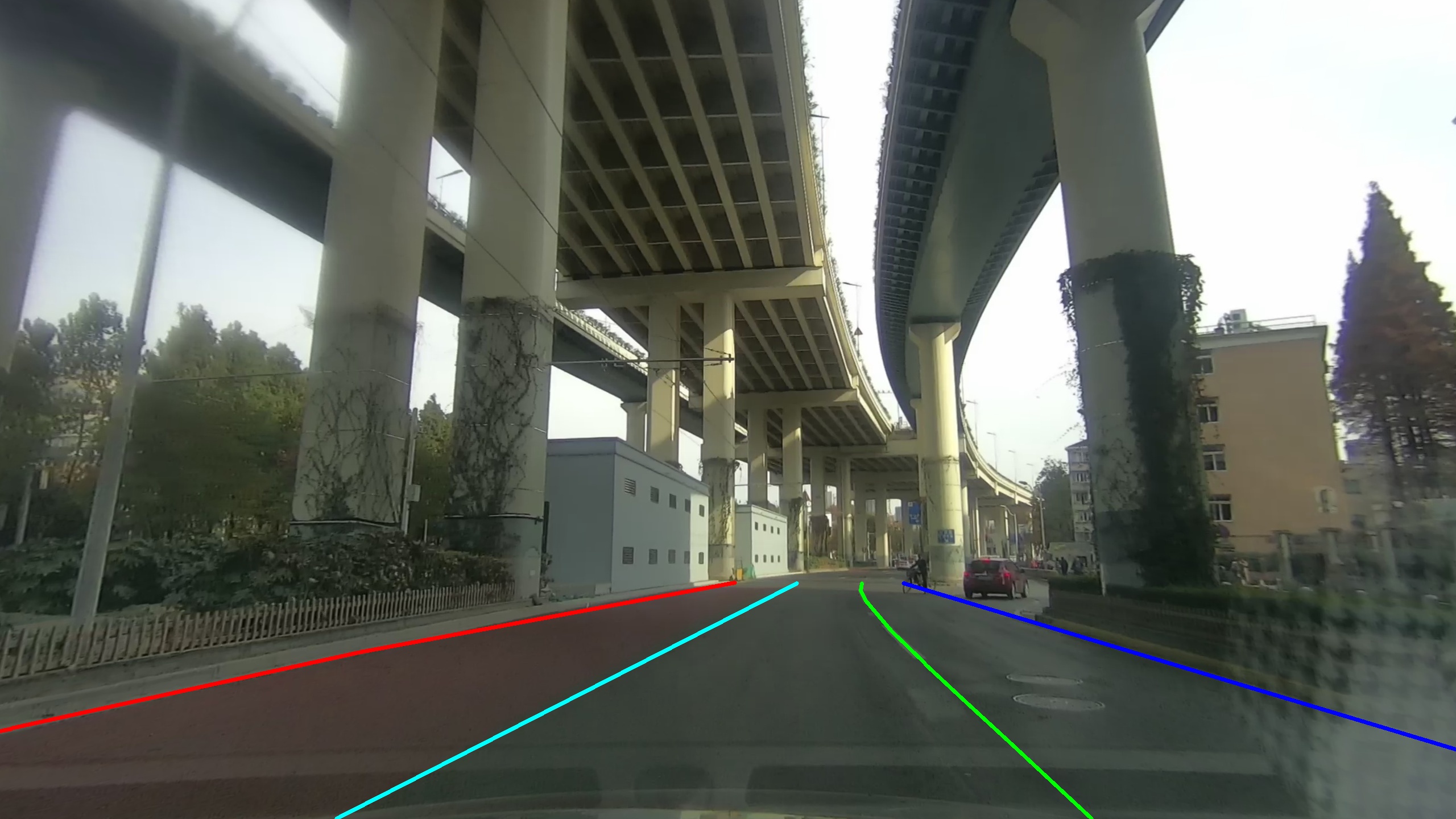}&
        \includegraphics[width=2.0in,totalheight=0.9in]{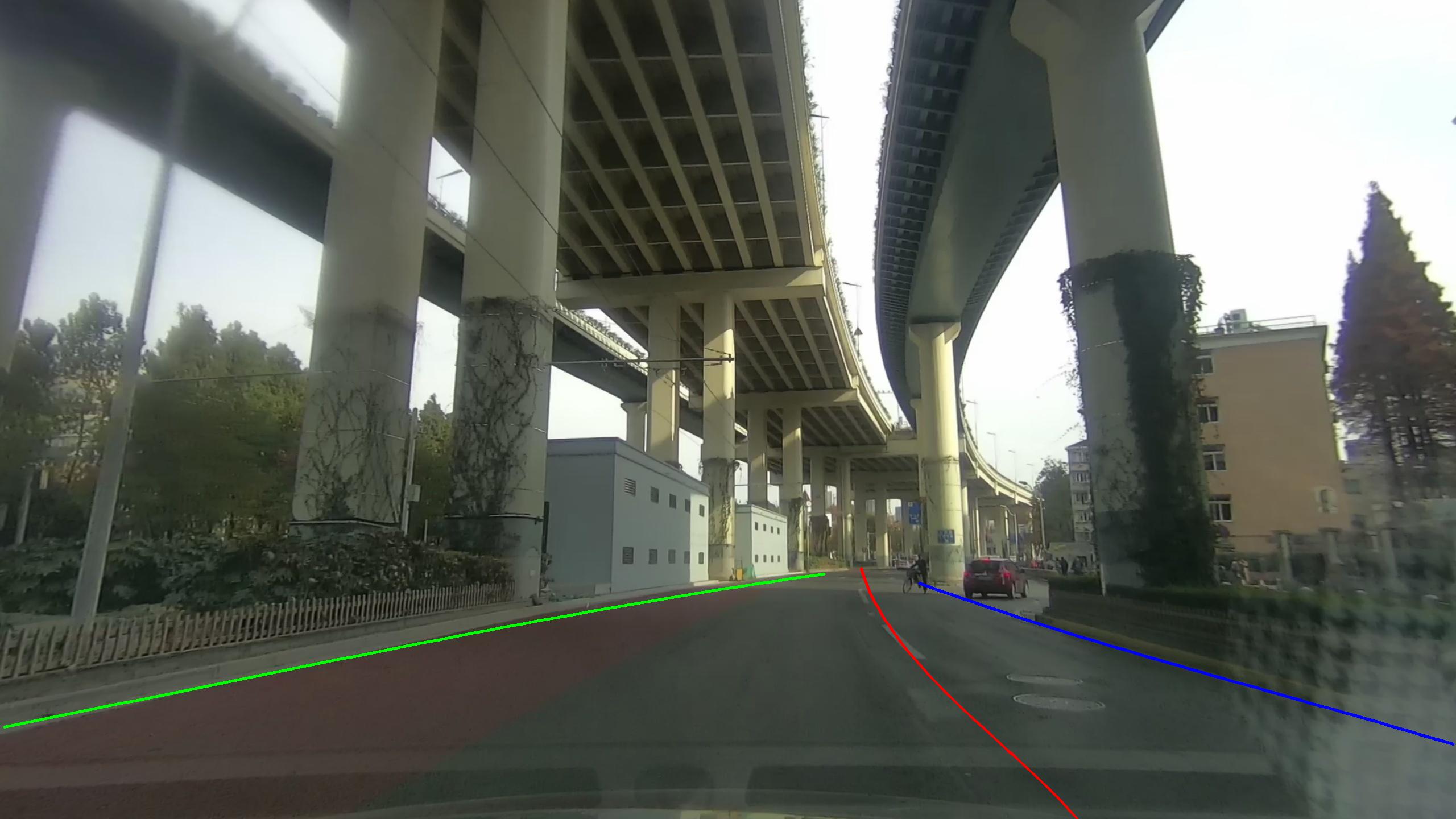}&
        \includegraphics[width=2.0in,totalheight=0.9in]{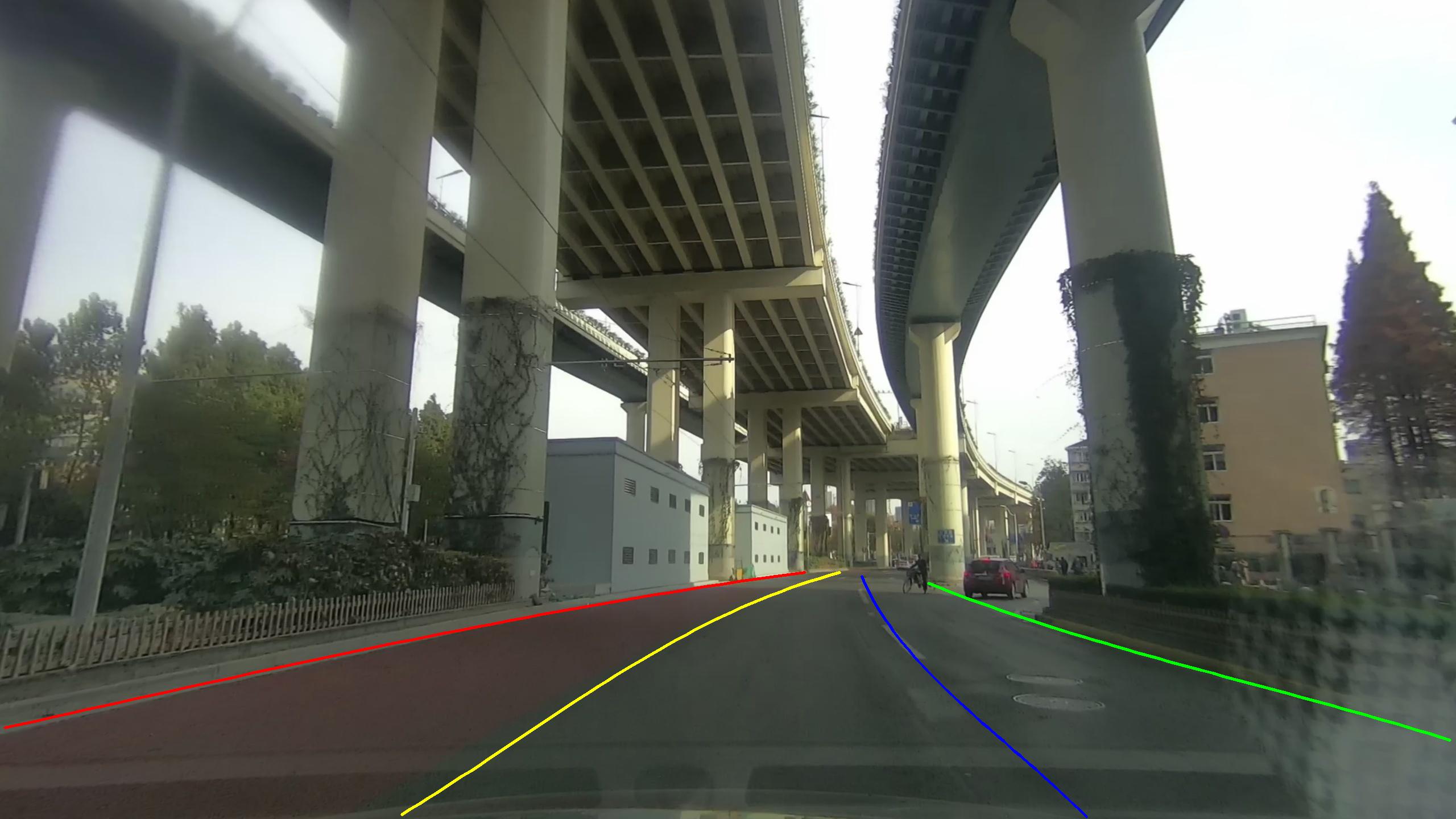}\\
        \includegraphics[width=2.0in,totalheight=0.9in]{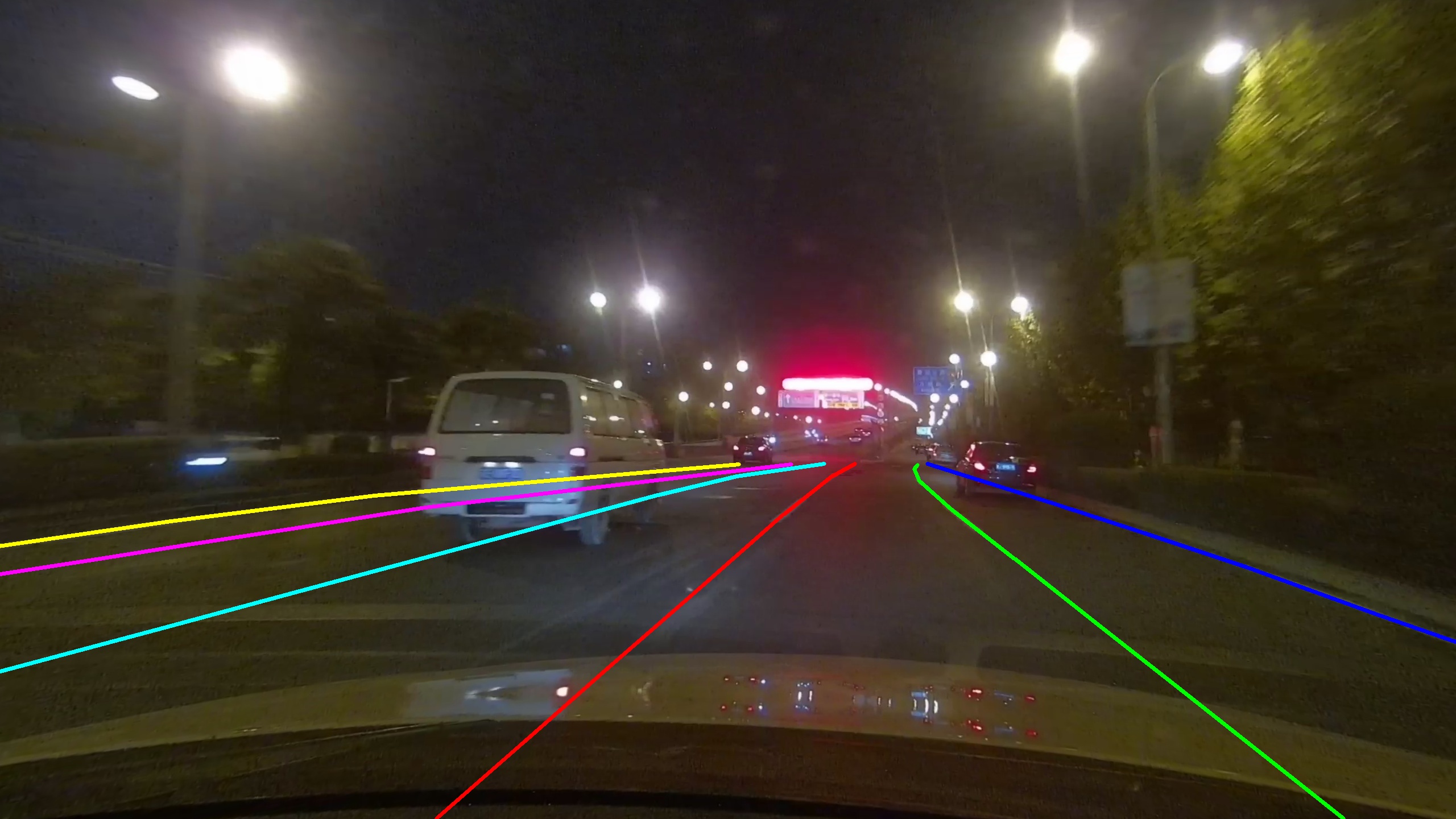}&
        \includegraphics[width=2.0in,totalheight=0.9in]{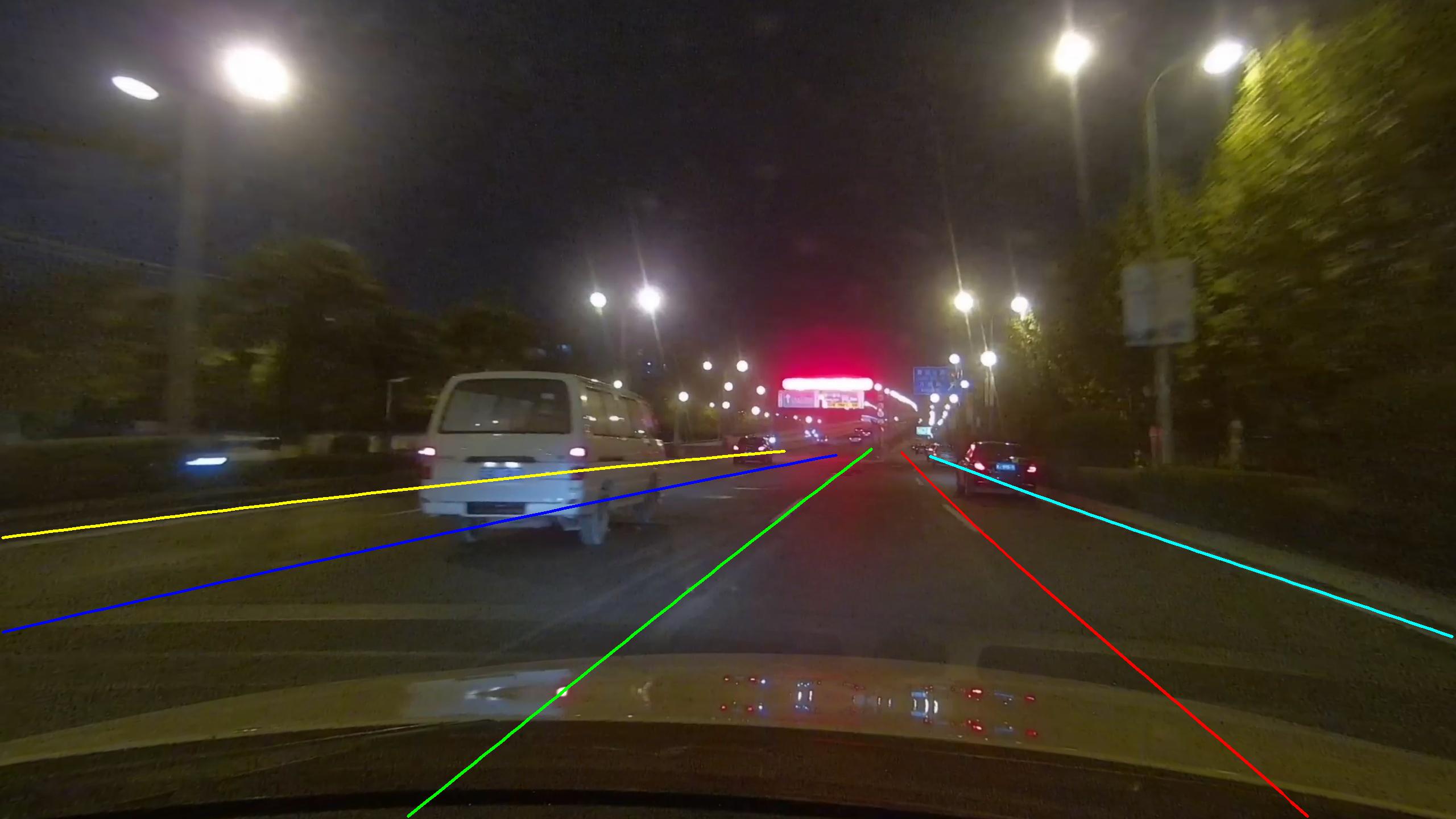}&
        \includegraphics[width=2.0in,totalheight=0.9in]{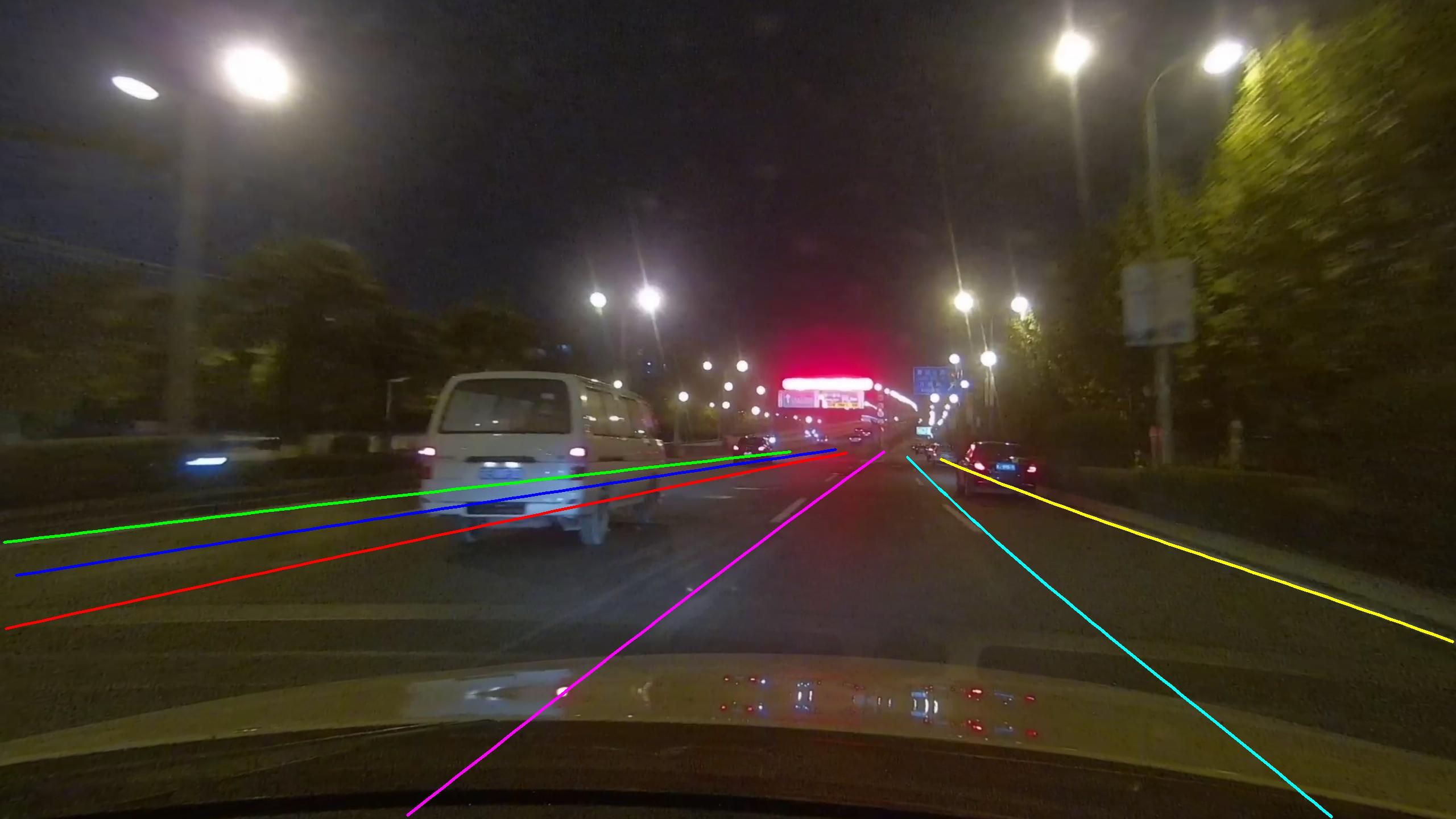}\\
        \includegraphics[width=2.0in,totalheight=0.9in]{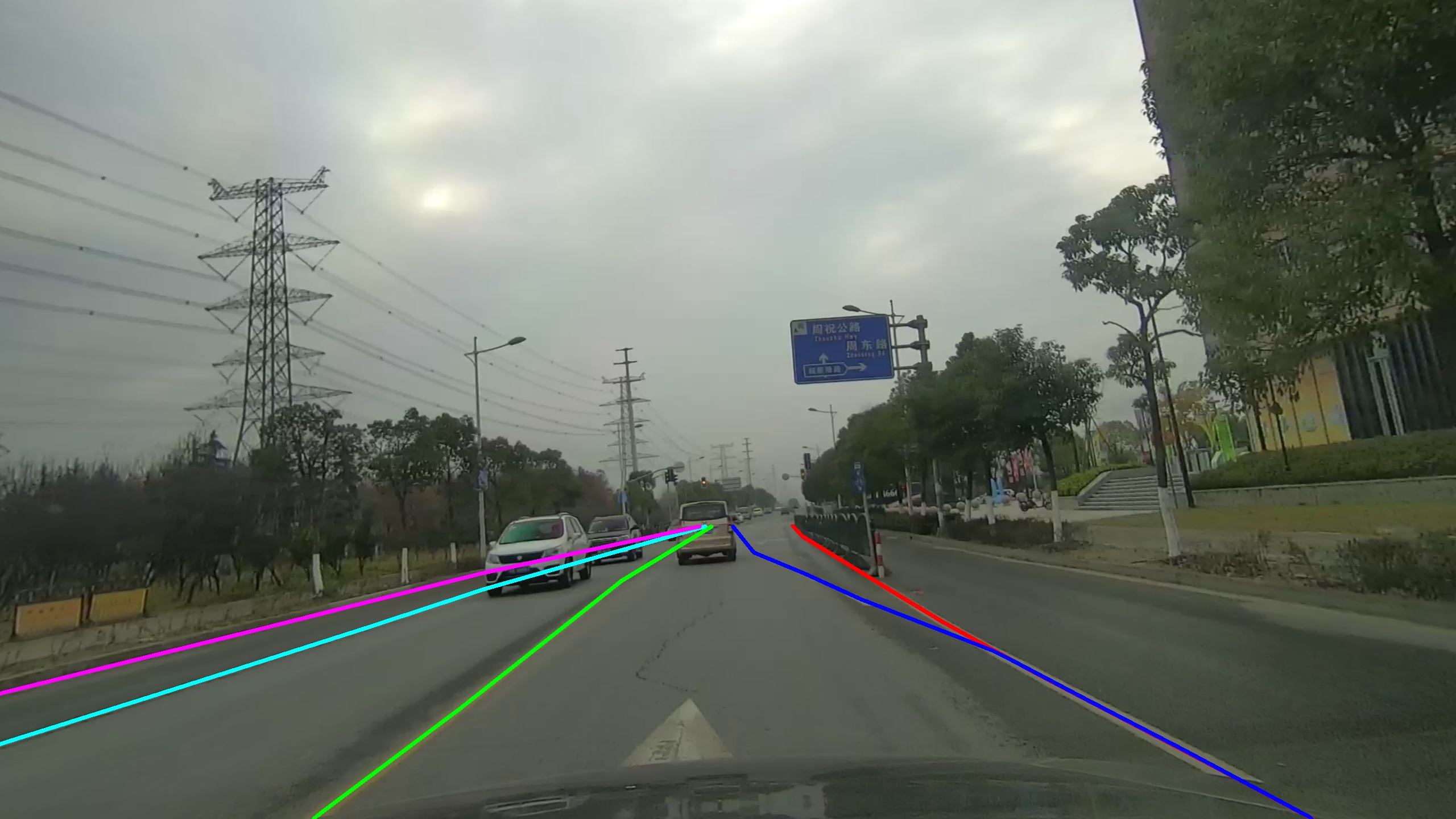}&
        \includegraphics[width=2.0in,totalheight=0.9in]{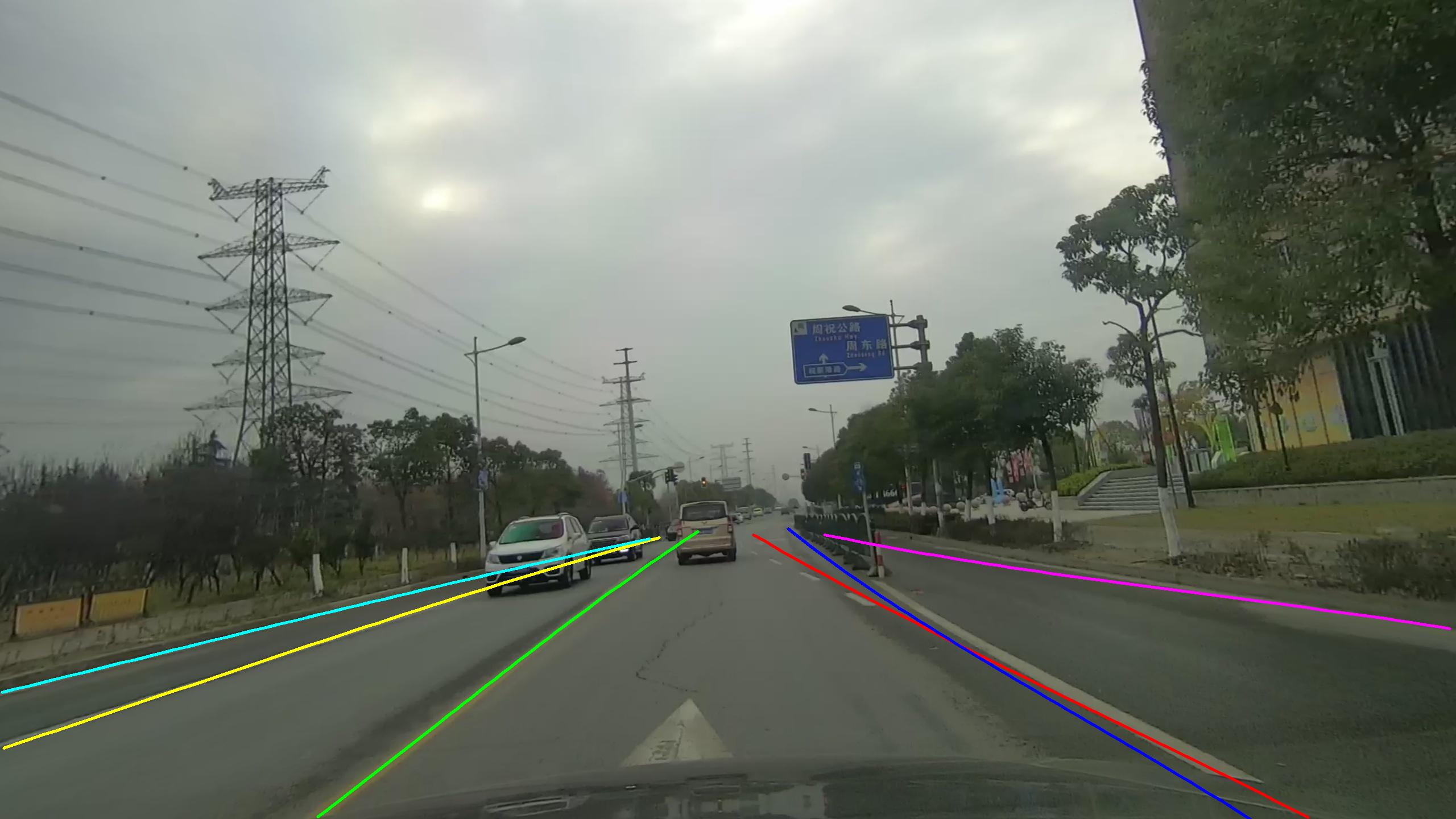}&
        \includegraphics[width=2.0in,totalheight=0.9in]{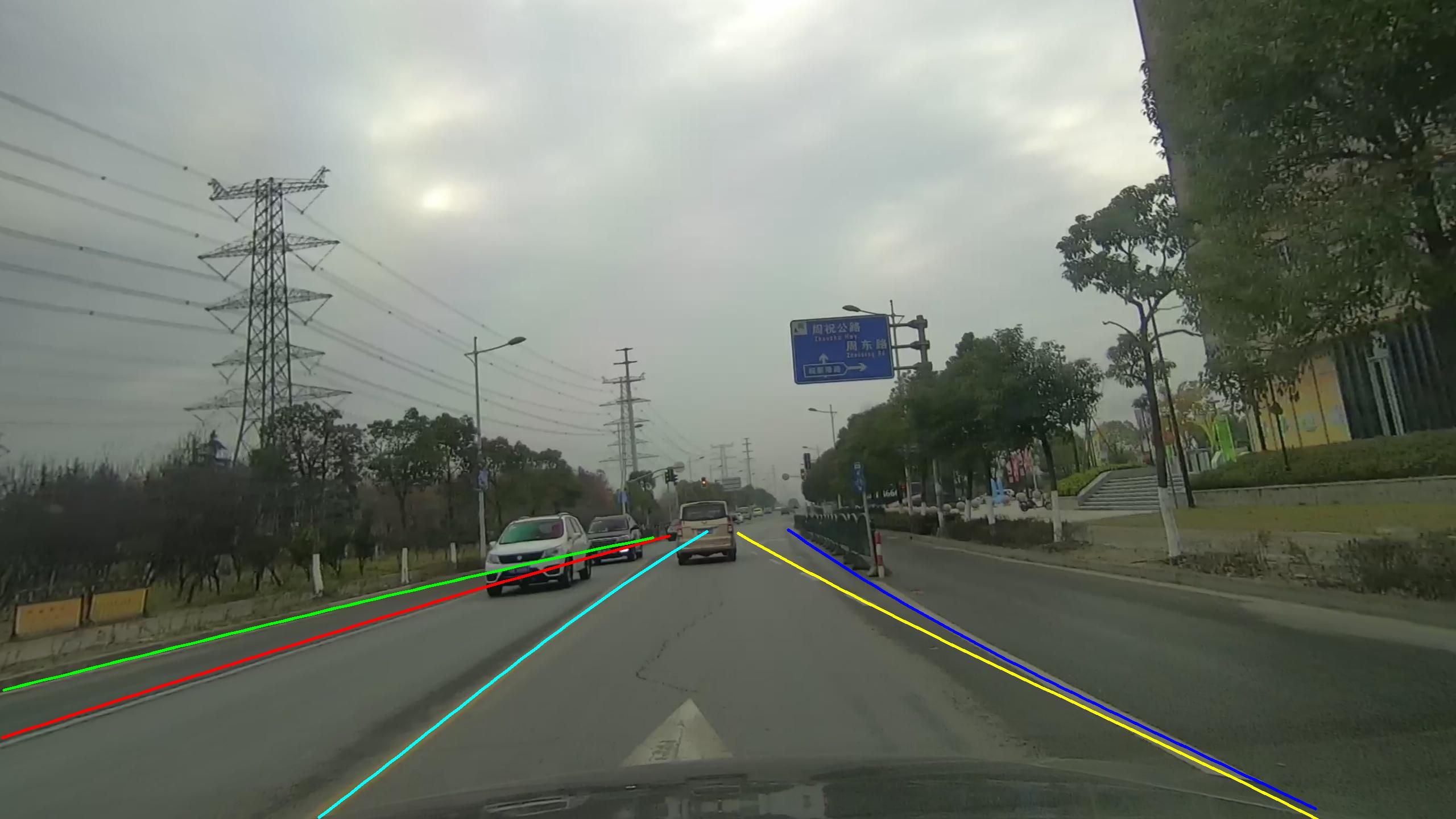}\\
        Ground truth & BézierLaneNet~\cite{bezier} & Ours
    \end{tabular}
\caption{Qualitative results on the CurveLanes~\cite{CurveLane} dataset.}
\label{fig:curvelane}
\vspace{-3mm}
\end{figure*}

We conduct experiments on the CurveLanes~\cite{CurveLane} dataset to better prove the effectiveness of our method on curves. The results are shown in Table~\ref{tab:CurveLanes-result}. Since the CurveLanes dataset~\cite{CurveLane} does not release the labels of the test set, we only report the evaluation results on the validation set. The CurveLanes~\cite{CurveLane} dataset has more than 90\% curve samples, so the improvement of our method on it can well prove that our method has a positive effect on curve modeling. The visual comparisons between DHPM and others are shown in Fig.~\ref{fig:curvelane}.

\subsection{Cross-Dataset Generalization Results}
To verify the generalization of our method, we conduct cross-dataset experiments between every two datasets. We use the model trained on one dataset as a starting point to train and test on another dataset. 
When re-training on the target datasets, we use the same loss function as BézierLaneNet~\cite{bezier}, which is illustrated in Equation~\eqref{eq:losssparse}.
As comparative experiments, we also conduct cross-dataset re-training experiments on the baseline model, utilizing the same training settings. Consistent improvements in Table~\ref{tab:Cross-Dataset} show that the starting checkpoint provided by our method is generally better, which demonstrates that our densely modulated proposals have stronger generalization ability and can be used as a good feature initialization.

\begin{table}[tb]
  \begin{center}
    \caption{Quantitative Results on the LLAMAS~\cite{llamas} Dataset}
    \renewcommand{\arraystretch}{1.4}
        \begin{tabular}{lccc} 
            \toprule
            \textbf{Method}                       & \textbf{F1} & \textbf{Presicion} & \textbf{Recall}  \\ 
            \hline
            \textbf{Anchor-based Method}            &             &             &        \\
            \cline{1-1}
            PolyLaneNet~\cite{PolyLaneNet}         & 88.40       & 88.87       & 87.93  \\
            CLRNet (ResNet-18)~\cite{CLRNet}       & 96.00       & -           & - \\
            LaneATT(ResNet-18) ~\cite{LaneATT}     & 93.46       & 96.92       & 90.24  \\
            UFLDv2 (ResNet-18)~\cite{UFLDv2}       & 94.58       & 95.29       & 93.88  \\
            UFLDv2 (ResNet-34)~\cite{UFLDv2}       & 94.95       & 95.75       & 94.17  \\
            \hline
            \textbf{Curve-based Method}             &             &             &        \\
            \cline{1-1}
            PolyLaneNet~\cite{PolyLaneNet}         & 88.40       & 88.87       & 87.93  \\
            BézierLaneNet (ResNet-18)~\cite{bezier} & 94.91       & 95.71       & 94.13  \\
            BézierLaneNet (ResNet-34)~\cite{bezier} & 95.17       & 95.89       & 94.46  \\ 
            \textbf{ours-BézierLaneNet (ResNet-18)} & 95.15       & 96.05       & 94.26  \\
            \textbf{ours-BézierLaneNet (ResNet-34)} & 95.30       & 96.16       & 94.46  \\
            \bottomrule
        \end{tabular}
        \label{tab:llamas-result}
    \end{center}
    \vspace{-3mm}
\end{table}

\begin{table}[t]
    \begin{center}
    \caption{Quantitative Results on the validation set of CurveLanes~\cite{CurveLane}}
    \renewcommand{\arraystretch}{1.4}
        \begin{tabular}{lccc} 
            \toprule
            \textbf{Method}  & \textbf{F1} & \textbf{Precision} & \textbf{Recall} \\ 
            \hline
            \textbf{Anchor-based Method}    &        &        &        \\
            \cline{1-1}
            {SCNN (LargeFOV)~\cite{culane}}   & {65.02}  & {76.13}  & {56.74}  \\
            {SAD (ENet)~\cite{ENet-SAD}}      & {50.31}  & {63.60}  & {41.60}  \\
            {PointLaneNet (MobileNet-v2)~\cite{PointLaneNet}} &{78.47} &{86.33} &{72.91} \\
            {CurveLanes-NAS (ResNet-18)~\cite{CurveLane}} & {81.12} & {93.58} & {71.59} \\
            {CondLaneNet (ResNet-18)~\cite{CondLaneNet}} & {85.09} & {87.75} & {82.58} \\
            \hline
            \textbf{Curve-based Method}     &        &        &        \\
            \cline{1-1}
            BézierLaneNet (ResNet-18)~\cite{bezier}        & 74.56  & 83.27  & 67.50  \\
            ours-BézierLaneNet (ResNet-18)                 & 75.03  & 82.75  & 68.62  \\
            \bottomrule
        \end{tabular}
    \label{tab:CurveLanes-result}
    \end{center}
    \vspace{-3mm}
\end{table}

\begin{table}[t]
  \centering
    \caption{Quantitative Comparison of Cross-Dataset Generalization}
    \renewcommand{\arraystretch}{1.4}
    \setlength{\tabcolsep}{1mm}
    \begin{tabular}{|c|p{0.85cm}p{0.85cm}|p{0.85cm}p{0.85cm}|p{0.85cm}p{0.85cm}|}\hline
      \multirow{2}{*}{\diagbox{Target}{Source}} &
      \multicolumn{2}{|c|}{TuSimple~\cite{tusimple}}   & \multicolumn{2}{|c|}{CULane~\cite{culane}}  & \multicolumn{2}{|c|}{LLAMAS~\cite{llamas}} \\ \cline{2-7} & baseline & ours & baseline & ours & baseline & ours\\ \hline
      TuSimple~\cite{tusimple} & 95.01  & \textbf{95.44}  & 94.98  & \textbf{95.13} & 95.51 & \textbf{95.70} \\ \hline
      CULane~\cite{culane}     & 73.48  & \textbf{73.91}  & 73.36  & \textbf{74.33} & 72.53 & \textbf{73.56} \\ \hline
      LLAMAS~\cite{llamas}     & 95.36  & \textbf{95.78}  & 95.52  & \textbf{95.66} & 95.42 & \textbf{95.75} \\ \hline
    \end{tabular}
    \label{tab:Cross-Dataset}
    \vspace{-3mm}
\end{table}

{
\begin{table}[t]
\centering
    \caption{Inference speed and model size. The inference speed was tested on an NVIDIA GTX 3090 GPU.}
    \renewcommand{\arraystretch}{1.4}
        \begin{tabular}{lcc} 
            \toprule
            \textbf{Method}                        & \textbf{FPS (image/s)} & \textbf{Params (M)}  \\ 
            \hline
            LaneATT (ResNet-18)~\cite{LaneATT}       & 205                   & 12.02              \\
            LaneATT (ResNet-34)~\cite{LaneATT}      & 183                   & 22.13               \\
            CLRNet (ResNet-18)~\cite{CLRNet}        & 130                   & 11.77               \\
            CLRNet (ResNet-34)~\cite{CLRNet}        & 106                   & 21.88               \\
            CondLaneNet (ResNet-18)~\cite{CondLaneNet}& 201                 & 11.93                \\
            CondLaneNet (ResNet-34)~\cite{CondLaneNet}& 149                 & 22.04                \\
            UFLDv2 (ResNet-18)~\cite{UFLDv2}        & 320                   & 206.30               \\
            UFLDv2 (ResNet-34)~\cite{UFLDv2}        & 162                   & 216.41               \\
            BézierLaneNet (ResNet-18)~\cite{bezier} & 212                   & 4.10                \\
            BézierLaneNet (ResNet-34)~\cite{bezier} & 179                   & 9.48                \\ 
            \textbf{ours-BézierLaneNet (ResNet-18)}               & 206                   & 4.10                \\
            \textbf{ours-BézierLaneNet (ResNet-34)}               & 183                   & 9.48                \\
            \bottomrule
        \end{tabular}
    \label{tab:FPS}
    \vspace{-3mm}
\end{table}
}

\vspace{-4mm}
\subsection{Speed and Parameters}
Inference speed and model size are important for autonomous driving algorithms. We report our inference speed and parameters in Table~\ref{tab:FPS}, where all FPS results are tested with $590 \times 1640$ random inputs on the same device, i.e., a single NVIDIA GTX 3090 GPU. Since our method only re-design the training method and does not modify the architecture, the testing speed and parameters are the same as BézierLaneNet. To sum up, our approach improves model performance and generalization without increasing new parameters and reducing inference speed.

\subsection{Ablation Study}
\begin{figure}[t]
    \centering
    \subfigure[Proposals under the supervision of $J_{ava}$ alone]{
    	\includegraphics[width=1.7in]{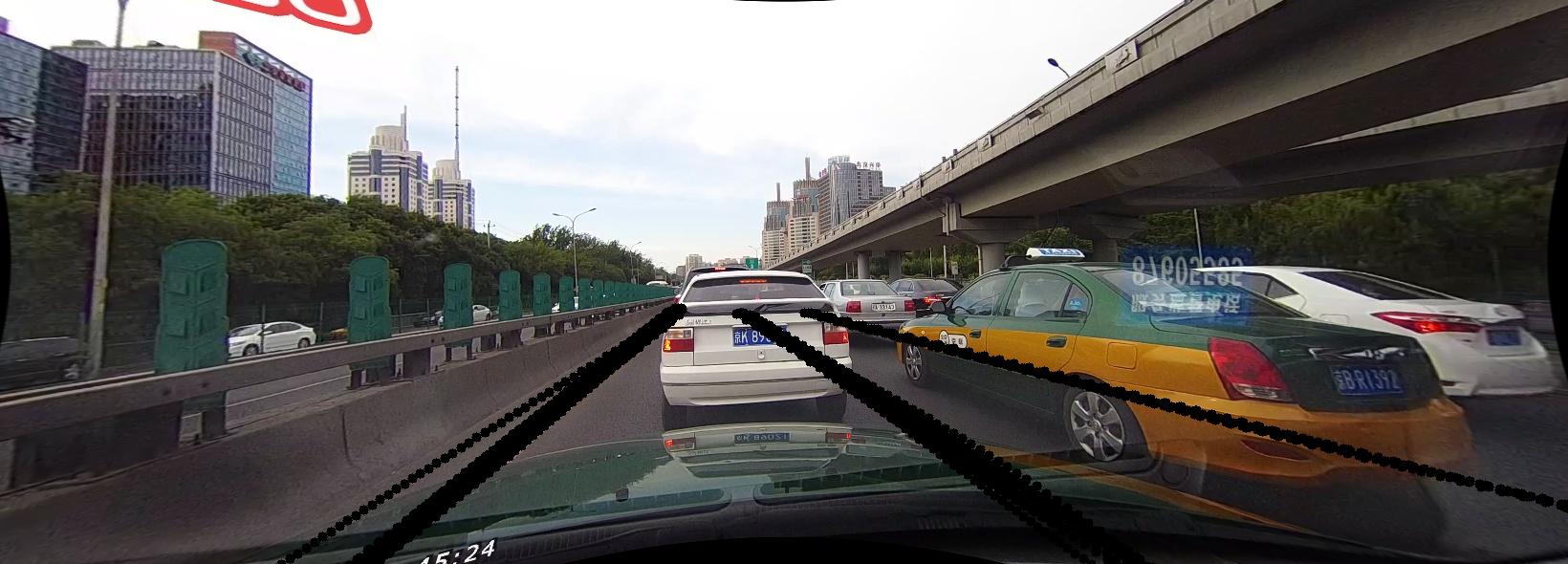}
    	\includegraphics[width=1.7in]{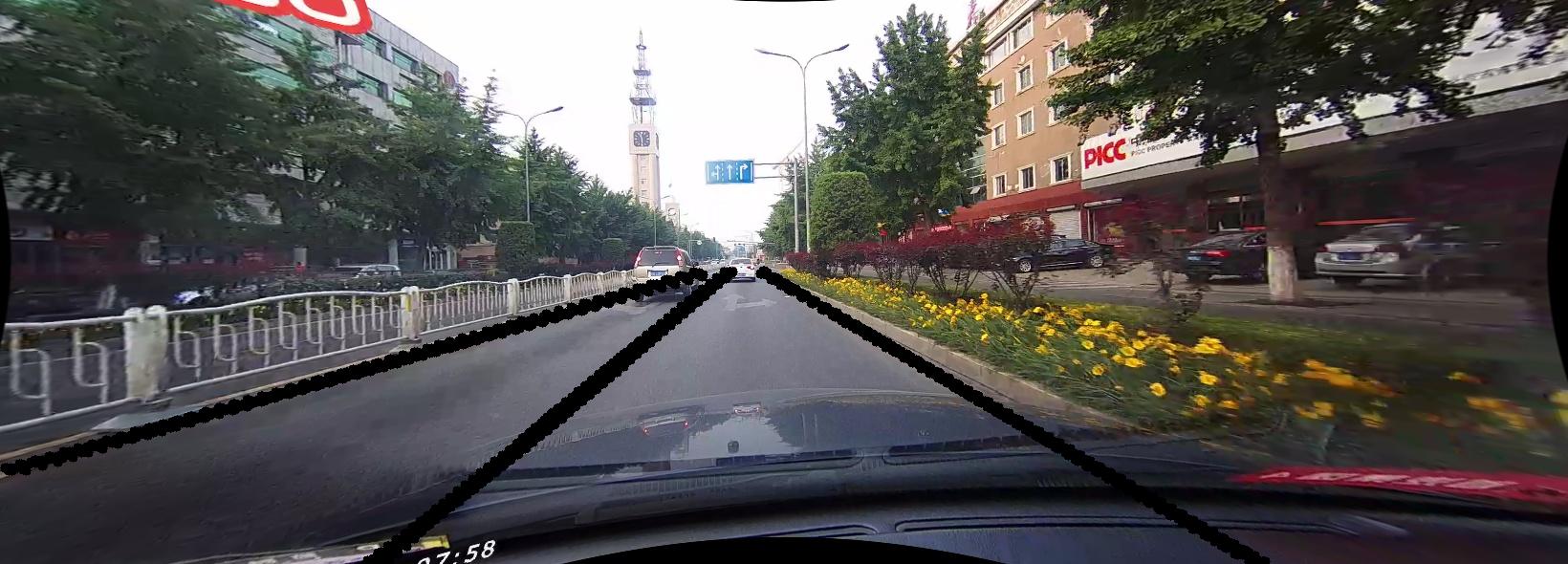}
    }
    \subfigure[Proposals under the supervision of $J_{div}$ alone]{
    	\includegraphics[width=1.7in]{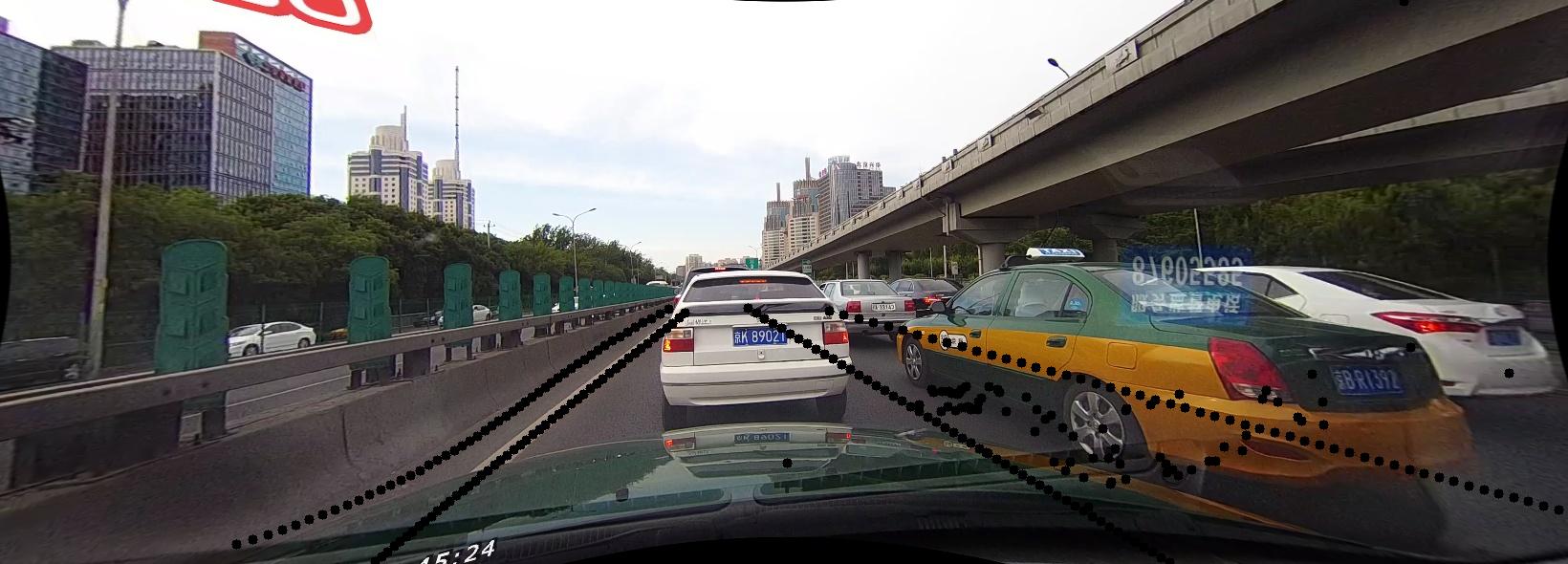}
    	\includegraphics[width=1.7in]{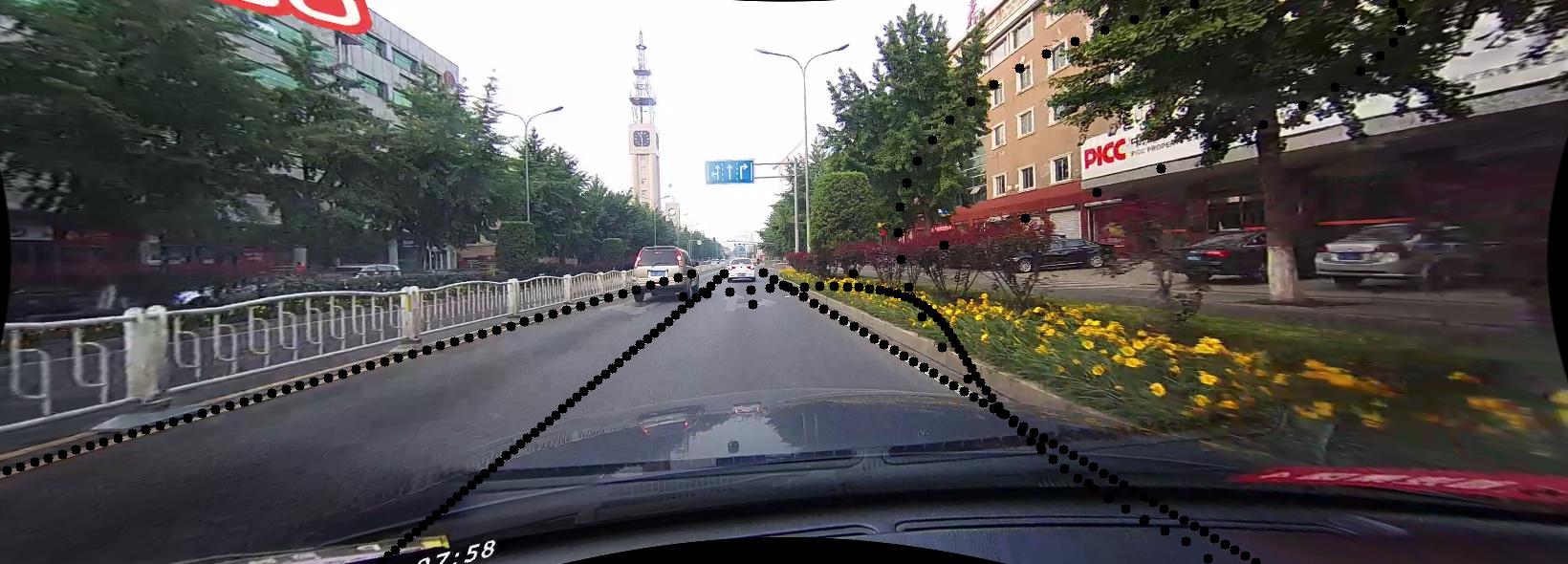}
    }
    \subfigure[Proposals under the supervision of $J_{ava}$ and $J_{div}$]{
    	\includegraphics[width=1.7in]{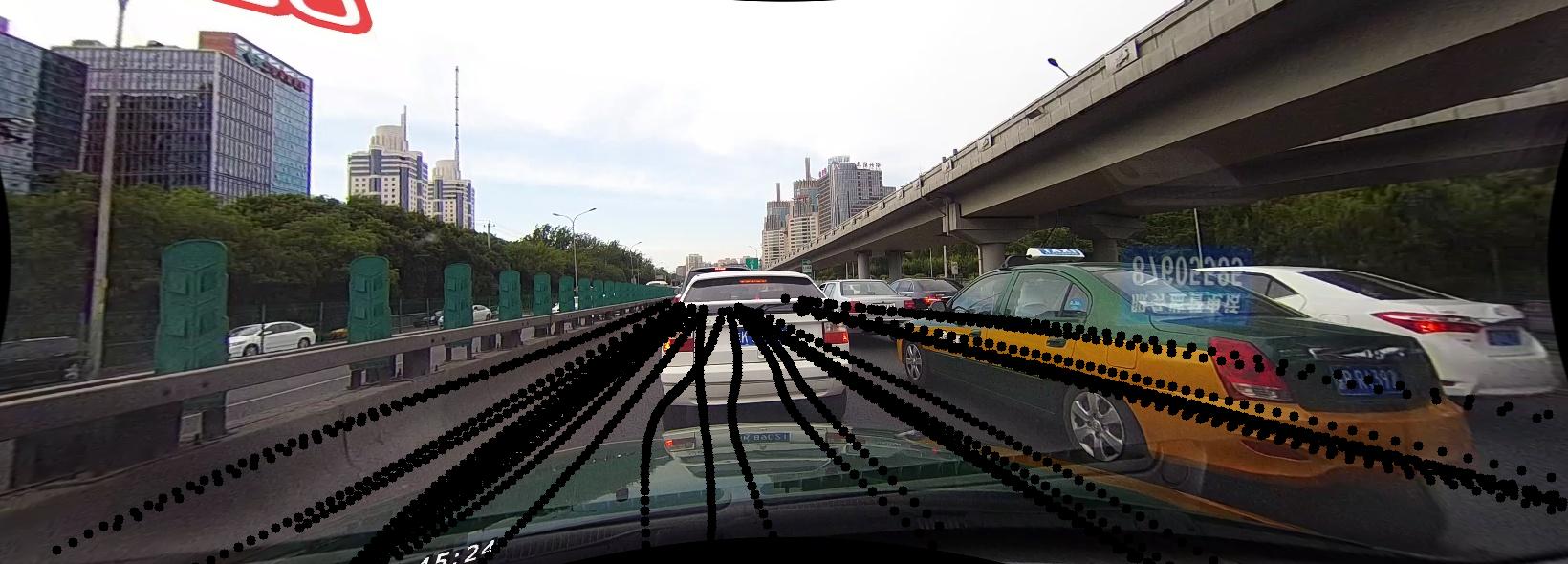}
    	\includegraphics[width=1.7in]{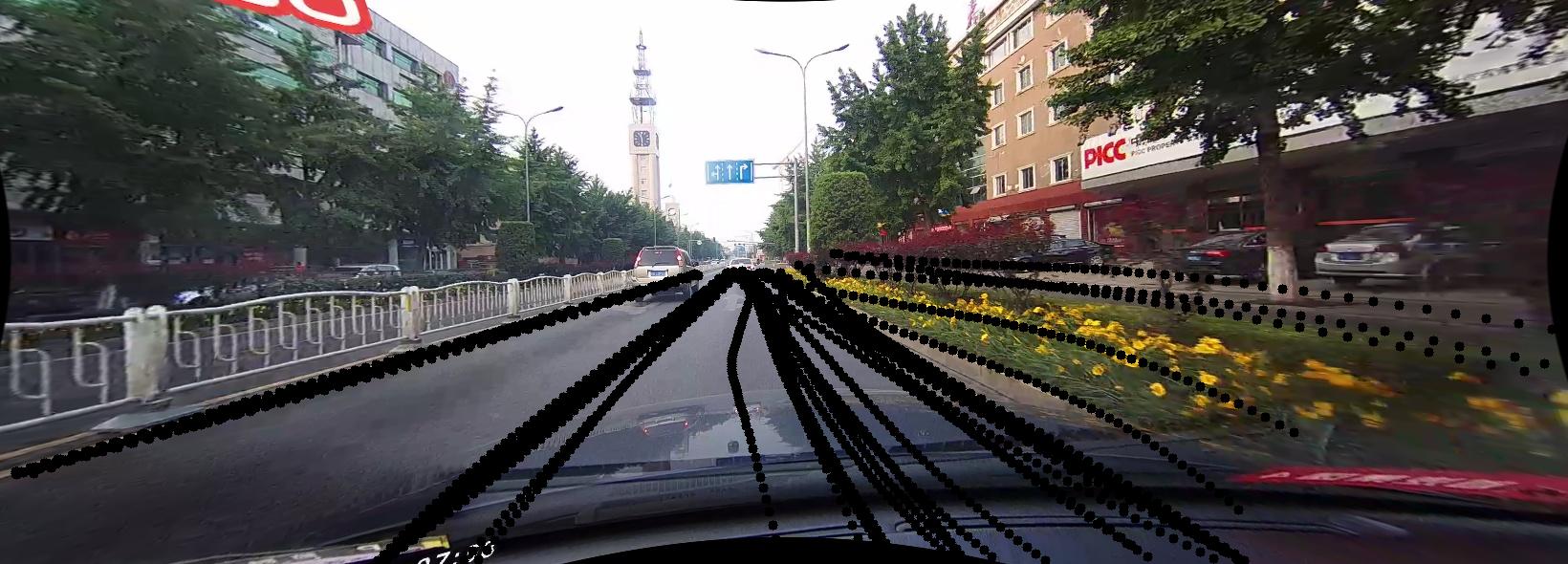}
    }
    \caption{The visualization of ablation experiments.
    The proposals sparsely supervised by the availability constraint (a) would cause the overlapping problem. Only applying the diversity constraint (b) would result in low-quality predictions, where most proposals are no longer located in the image. When we add the diversity constraint based on the availability constraint (c), proposals are more dispersed in the image space and of higher diversity.}
    \label{fig:ablation}
    \vspace{-3mm}
\end{figure}

We conduct ablation experiments on the CULane~\cite{culane} dataset to verify the effectiveness of each component. All the ablation experiments are performed based on the ResNet-18 version of our method. The results are shown in Table~\ref{tab:Ablation-study}, which are the average values of four independent experiments with the same training settings. The visualization of ablation experiments is shown in Fig.~\ref{fig:ablation}.     We emphasize that the availability constraint $J_{ava}$ aims to supervise the location and shape attributes of a \textbf{single} lane proposal, while the diversity constraint $J_{div}$ is proposed to restrain the relationship of \textbf{multiple} lane proposals.
    In other words, the availability constraint is a unitary constraint about $S_k$, while the diversity constraint is a dualistic constraint. They make their own contributions and assist each other in jointly optimizing the model. 

\begin{table}[t]
    \centering
    \caption{Ablation Study of the Improvement Strategies on CULane}
    \renewcommand{\arraystretch}{1.4}
    \setlength{\tabcolsep}{2mm}
        \begin{tabular}{cccccc} 
            \toprule
                Model & Basic constraints & $J_{ava}$ & $J_{div}$ & $J_{dis}$ & F1 Score  \\ 
                \hline
                (a) &\checkmark      &                         &                      &                           & 73.36     \\
                (b) &\checkmark      & \checkmark              &                      &                           & 73.14     \\
                (c) &\checkmark      &                         & \checkmark           &                           & 73.28     \\
                (d) &\checkmark      &                         &                      & \checkmark                & 73.92     \\
                (e) &\checkmark      & \checkmark              & \checkmark           &                           & 73.72     \\
                (f) &\checkmark      & \checkmark              &                      & \checkmark                & 73.66     \\
                (g) &\checkmark      &                         & \checkmark           & \checkmark                & 73.85     \\
                (h) &\checkmark      & \checkmark              & \checkmark           & \checkmark                & 74.33     \\
            \bottomrule
        \end{tabular}
        \label{tab:Ablation-study}
        \vspace{-4mm}
\end{table}

From Table~\ref{tab:Ablation-study} (a-c), we observe that availability constraint or diversity constraint do not positively impact training and even lead to performance degradation. From models (a-d), we observe that the models trained with only $J_{ava}$ or $J_{div}$ fail to bring a performance boost while $J_{dis}$ successfully improved the performance from the perspective of the confidence distribution of proposals. The reason for this phenomenon can be attributed to the fact that $J_{ava}$ and $J_{div}$ are designed to regulate the spatial distributions of proposals from different orientations. On the one hand, $J_{ava}$ is proposed to locate each proposal around the ground truth lanes with a reasonable shape. On the other hand, $J_{div}$ enhances the inter-proposal topological diversification to exploit the underlying relations between proposals. Therefore, the combination of two losses (i.e., model (e)) significantly improves the detection results of our method compared to the baseline. After adding the discrimination constraint, the average F1 score of our method can reach 74.33, which proves the feasibility of improving the quality of all proposals.

The goal of availability constraint is to improve the quality of all proposals. To achieve that, we need to constrain their shapes and locations to a reasonable range. Table~\ref{tab:avablity-constraint} presents the ablation experiment to evaluate each component (i.e., shape and location) in the availability constraint. We can observe that none of them can improve performance alone, because the attributes they constrain are orthogonal, so constraining the shape and location of the proposal at the same time will bring obvious performance improvement.

\begin{table}[t]
\centering
\caption{Ablation Study of Availability Constraint}
\renewcommand{\arraystretch}{1.4}
    \begin{tabular}{cccc} 
        \toprule
            Baseline & Location constraint & Shape constraint & F1 Score \\ 
            \hline
            \checkmark      &             &             & 73.85     \\
            \checkmark      & \checkmark  &             & 73.80     \\
            \checkmark      &             & \checkmark  & 74.06     \\
            \checkmark      & \checkmark  & \checkmark  & 74.33     \\
        \bottomrule
    \end{tabular}
    \label{tab:avablity-constraint}
\end{table}

As described above, the diversity constraint is divided into two parts: the constraint to increase shape differences and the upper limit of proposal-to-label matching. We conduct three groups of experiments to verify their effectiveness, and results are shown in Table~\ref{tab:diversity-constraint}.
By comparing the first two rows in Table~\ref{tab:diversity-constraint}, we can see that the proposed difference constraint has successfully improved the performance. Comparing the results of the last two rows, we can see the positive effect of an upper limit. Comparing these three groups of experiments proves that both parts of the diversity constraint are essential.

\begin{table}[t]
    \centering
    \caption{Ablation Study of Diversity Constraint}
    \renewcommand{\arraystretch}{1.4}
        \begin{tabular}{cccc} 
            \toprule
                Baseline        & Difference Constraint   & Upper Limit          & F1 Score  \\ 
                \hline
                \checkmark      &                         &                      & 73.66     \\
                \checkmark      & \checkmark              &                      & 74.01     \\
                \checkmark      & \checkmark              & \checkmark           & 74.33     \\
            \bottomrule
        \end{tabular}
        \label{tab:diversity-constraint}
        \vspace{-3mm}
\end{table}

Compared with the first two constraints, the discriminative constraints are relatively independent. The conclusion can be drawn by comparing (a-d) and (e-h) of Table~\ref{tab:Ablation-study}. We find that experiments with discriminative constraints have higher F1 scores than experiments without discriminative constraints, leading to the conclusion that discriminative constraints alone can have a positive effect.

\section{Conclusion}
\label{sec:conclusion}
In this paper, we have proposed dense hybrid proposal modulation to generate high-quality proposals for lane detection. The availability constraint makes all proposal lane lines have suitable locations and shapes while the diversity constraint avoids all the proposals under the supervision of availability constraint overlapping together and improves the diversity of proposals. The discriminative constraints improve the discriminative ability of detectors and filter out proposals that are not good enough. Benefitting from these constraints, we achieve very competitive results on four popular datasets.


{\small
    \bibliographystyle{IEEEtran}
    \bibliography{IEEE}
}

\end{document}